\documentclass{article}

\usepackage{microtype}
\usepackage{graphicx}
\usepackage{subcaption}
\usepackage{booktabs} % for professional tables
\usepackage{hyperref}
\usepackage{wrapfig}

\usepackage[nonatbib,preprint]{neurips_2026}
\usepackage[square,numbers]{natbib}

\usepackage{amsmath}
\usepackage{amssymb}
\usepackage{mathtools}
\usepackage{amsthm}
\usepackage{pifont}

\usepackage{siunitx}
\DeclareSIUnit{\cells}{cells}
\DeclareSIUnit\angstrom{\text{Å}}

\usepackage{textcomp}
\usepackage{minted}

\usepackage[most]{tcolorbox}
\tcbset{
  promptbox/.style={
    colback=gray!5,
    colframe=gray!60,
    boxrule=0.5pt,
    arc=4pt,
    left=6pt,
    right=6pt,
    top=6pt,
    bottom=6pt,
    fontupper=\ttfamily,
    breakable
  }
}
\usepackage{algorithm}
\usepackage{algorithmic}

\usepackage[capitalize,noabbrev]{cleveref}

\theoremstyle{plain}

\theoremstyle{definition}

\theoremstyle{remark}

\usepackage[utf8]{inputenc} % allow utf-8 input
\usepackage[T1]{fontenc}    % use 8-bit T1 fonts
\usepackage{url}            % simple URL typesetting
\usepackage{booktabs}       % professional-quality tables
\usepackage{amsfonts}       % blackboard math symbols
\usepackage{nicefrac}       % compact symbols for 1/2, etc.
\usepackage{microtype}      % microtypography
\usepackage{xcolor}         % colors

\usepackage[inline]{enumitem}

\newcommand{\ourmethod}{\texttt{PABLO}}
\newcommand{\ourmethodfullname}{Purely Agent-driven BLack-box Optimization (\texttt{PABLO})}

\title{Purely Agent-Driven Black-Box Optimization for Biological Design}

\author{%
  Natalie Maus\thanks{nmaus@csail.mit.edu} \\
  \And
  Yimeng Zeng \\
  \And
  Haydn Thomas Jones \\
  \And
  Yining Huang \\
  \And
  Gaurav Ng Goel \\
  \And
  Alden Rose \\
  \And
  Kyurae Kim \\
  \And
  Hyun-Su Lee \\
  \And
  Marcelo Der Torossian Torres \\
  \And
  Fangping Wan \\
  \And
  Cesar de la Fuente-Nunez \\
  \And
  Mark Yatskar \\
  \And
  Osbert Bastani \\
  \And
  Jacob R. Gardner \\
  \AND
  University of Pennsylvania, Philadelphia, PA 19104, USA \\
}

\begin{document}

\maketitle

\begin{abstract}
Many key challenges in biological design---such as small-molecule drug discovery, antimicrobial peptide development, and protein engineering---can be framed as black-box optimization over vast, complex structured spaces. Existing methods rely mainly on raw structural data and struggle to exploit the rich scientific literature. While large language models (LLMs) have been added to these pipelines, they have been confined to narrow roles within structure-centered optimizers. We instead cast biological black-box optimization as an agent-driven, language-based reasoning process. We introduce \ourmethodfullname{}, a hierarchical agentic system that uses scientific LLMs pretrained on chemistry and biology literature to generate and iteratively refine biological candidates. On both the standard GuacaMol molecular design and antimicrobial peptide optimization tasks, \ourmethod{} achieves state-of-the-art performance, substantially improving sample efficiency and final objective values over established baselines. Compared to prior optimization methods that incorporate LLMs, \ourmethod{} achieves competitive token usage per run despite relying on LLMs throughout the optimization loop. Beyond raw performance, the agentic formulation offers key advantages for realistic design: it naturally incorporates semantic task descriptions, retrieval-augmented domain knowledge, and complex constraints. In follow-up \textit{in vitro} validation, \ourmethod{}-optimized peptides showed strong activity against drug-resistant pathogens, underscoring the practical potential of \ourmethod{} for therapeutic discovery.
\end{abstract}

\section{Introduction}
\label{sec:intro}

Searching for novel biological entities, \textit{i.e.}, small molecules, antimicrobial peptides, and proteins, has often been approached as a high-dimensional black-box optimization problem~\citep{negoescu2011knowledge,gomez2018automatic,JTVAE,GEGL,Weighted_Retraining,ladder,Huawei,lolbo}.
The objective is to identify discrete sequences or graphs that maximize a costly, often non-differentiable function representing desired properties such as binding affinity, solubility, stability, or antimicrobial potency.
Until now, this problem has been tackled via genetic algorithms~\citep{SMILES_GA,STONED,Graph_GA_AND_MCTS,SynNet,GA_D,GuacaMol,MIMOSA,DoG-AE}, Bayesian optimization (BO)~\citep{GP_BO,ChemBO,BOSS,lolbo,r2-cite5-lsbo,r2-cite9-lsbo-nfbo,apexgo,robot}, reinforcement learning~\citep{REINVENT,MolDQN,GEGL}, and deep generative modeling~\citep{JTVAE,DoG-AE,lolbo,r2-cite5-lsbo,r2-cite9-lsbo-nfbo,PepDiffusion,HydrAMP}. 
These approaches typically train generative models such as diffusion models, VAEs, and flow matching models.
Because they operate only on domain-specific input modalities (\textit{e.g.}, SMILES/SELFIES strings, amino acid or nucleic acid sequences), they largely ignore human knowledge in natural language, \textit{i.e.}, mechanistic insights, structure--activity relationships, and domain heuristics, unless such information is manually engineered into the models or search operators. 
As a result, they often must relearn biological principles from scratch for each new task.

This limitation has inspired a new wave of LLM-enhanced optimization strategies aimed at incorporating higher-level knowledge.
In fact, a large body of recent work \citep{chemcrow, autonomous_chem_research, auditable_agent_mol_opt, MTMol, jones2025datasetdistillingknowledgepriors,saga} has demonstrated that LLMs are capable of multimodal reasoning, prediction, and question answering over molecules and language over time.
As such, a natural step would be to incorporate these into the molecular discovery pipeline.
Indeed, a growing body of work integrates LLMs into evolutionary loops as intelligent variation operators \citep{MOLLEO, AlphaEvolve}, or embeds them within BO or other surrogate-based optimization pipelines to provide structural priors, candidate generators, or surrogate models \citep{bo_w_llms, bo_w_llms_2, llambo, ChemBOMAS, lico}. 
Going even further, \citep{opro, towards_optimizing_llms, gptopt} treated the LLM \textit{itself} as an optimizer via iterative prompting and feedback, while \citep{llmbox} proposes a hybrid LLM-guided optimization system that incorporates language-model reasoning into a classical black-box optimization workflow. 

At this point, we ask: what if we removed manually designed optimization strategies entirely, and instead relied solely on language models to explore the design space and develop their own strategies? 
Such an approach would immediately achieve the following:
(a) LLMs can incorporate domain knowledge acquired during pretraining into the optimization,
(b) domain-specific tools can be used via agentic tool calling, and
(c) the optimization process becomes highly flexible and easily adaptable to new settings by modifying the LLMs’ context.

We propose \ourmethodfullname{}, a framework where LLMs pretrained on chemistry and biology literature drive the entire optimization loop.
Unlike methods that embed LLMs within BO or evolutionary scaffolds, \ourmethod{} coordinates multiple LLM agents with distinct roles to explore the design space.
Namely, the \textit{Planner Agent} synthesizes high-level search strategies from experimental history, the \textit{Worker Agents} execute these strategies, and the \textit{Explorer Agent} proposes diverse global hypotheses.
This division of labor is general: the system is agnostic \textit{a priori} about how to construct new candidates---no need for acquisition or fitness functions, and all construction strategies are proposed by an LLM.
Furthermore, this hierarchy allows for revising the search strategy online, a capability that is difficult to express in LLM-enhanced BO/EA pipelines.

Scientific knowledge is incorporated into \ourmethod{} in two complementary ways.
First, the underlying models are pretrained on chemistry and biology literature, yielding representations that encode relationships and heuristics beyond what structure-only optimization can provide.
Second, the agentic framework supports external tools such as retrieval-augmented literature search \citep{lewis2020retrieval}, enabling targeted access to domain knowledge on demand.
\ourmethod{} also seamlessly incorporates semantic task descriptions and complex constraints via natural language interactions with the LLM.

Our contributions are summarized as follows:
\vspace{-2ex}
\begin{enumerate}[itemsep=0ex,leftmargin=3ex,label=$\bullet$]

\item We introduce \ourmethod{}, a general-purpose black-box optimization algorithm for biological design problems that replaces traditional structure-centric search components with a coordinated set of LLM-driven agents. 
\ourmethod{} is the first hierarchical, purely agent-driven framework that formulates biological black-box optimization as an end-to-end language-driven reasoning process rather than as an LLM component embedded within a structure-centric optimizer.

\item We evaluate \ourmethod{} on antimicrobial peptide design and 10 challenging GuacaMol molecular design tasks, showing it consistently outperforms structure-based baselines (Graph GA~\citep{graphGA}, NF-BO~\citep{r2-cite9-lsbo-nfbo}, GEGL~\citep{GEGL}) and recent LLM-based approaches (AlphaEvolve~\citep{AlphaEvolve}, LLAMBO~\citep{llambo}), finding higher-quality solutions with far fewer black-box evaluations and setting a new state of the art on GuacaMol.

\item
Even though \ourmethod{} relies entirely on LLMs, it does not require substantially more total tokens than existing methods that use LLMs only as a sub-component—and in some cases is significantly more token-efficient.  

\item We show that expressing the optimization loop in natural language allows integrating auxiliary problem information without redesigning the algorithm. In particular, \ourmethod{} can incorporate (i) semantic task descriptions, (ii) external literature via retrieval-augmented tools, (iii) complex design constraints, and (iv) diverse portfolio optimization objectives from \citet{robot}.

\item We validate \ourmethod{}-designed peptides using \textit{in vitro} minimum inhibitory concentration (MIC) assays, which show significant antimicrobial activity against clinically relevant pathogens.

\end{enumerate}

\vspace{-1ex}
\section{Method}
\vspace{-1ex}
\label{sec:method}
\begin{figure*}[!ht]
  \vspace{-4ex}
  \centering
    \centerline{\includegraphics[width=\columnwidth]{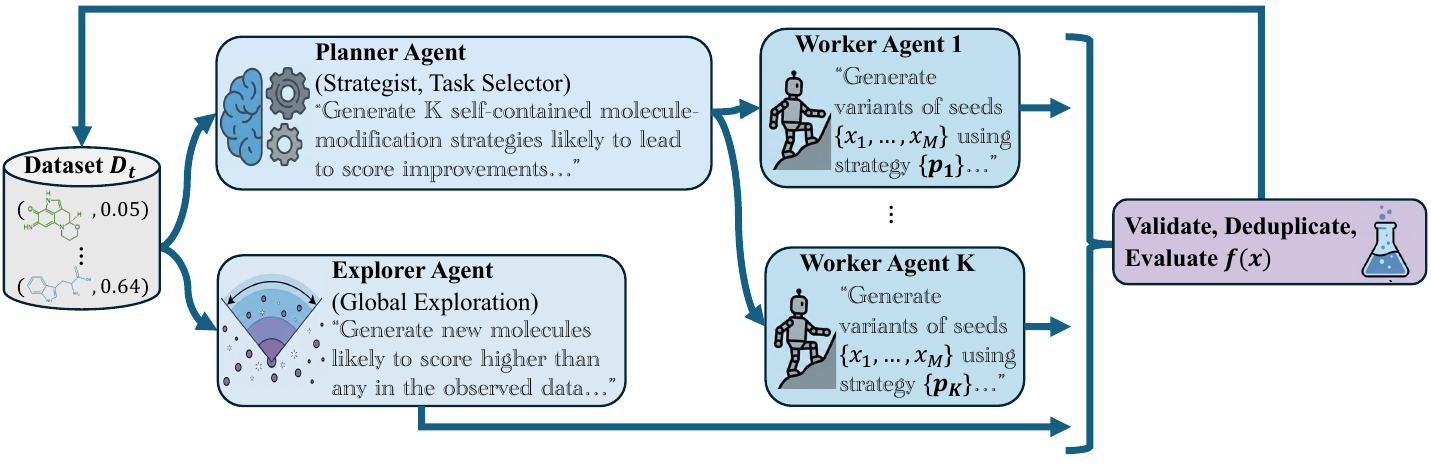}}
    \vspace{-2ex}
    \caption{
    Illustration of a single iteration of \ourmethod{}. Each iteration begins with (i) global candidate exploration, (ii) strategy generation via the Planner Agent, and (iii) local refinement of incumbents via planner-proposed strategies. 
    All candidate generations are filtered for validity, novelty, and feasibility before evaluation. \ourmethod{}-pseudocode is also provided in \cref{alg:agent_opt}.
    }
    \label{fig:diagram}
    \vspace{-3ex}
\end{figure*}

\ourmethodfullname{} is a hierarchical agentic framework that treats black-box optimization as an iterative loop of global strategy selection followed by local refinement. In each iteration, the local refinement phase improves the current best solutions using strategies chosen at the global level. In contrast, common methods like Bayesian optimization (BO) and evolutionary strategies (EA) fix their candidate selection approach in advance. For instance, BO typically commits to a static acquisition function strategy, with optimization feedback affecting only updates to the surrogate model posterior.

\ourmethod{} does not commit to any fixed strategy beyond factoring the optimization problem into the hierarchical agentic system defined below. Instead, global-level agents infer patterns from the observed optimization history and propose new hypotheses online for making progress.

We consider optimization over a structured discrete space $\mathcal{X}$ (\textit{e.g.}, protein/peptide sequences or molecular representations).
We describe \ourmethod{} in terms of maximization:
{%
\setlength{\belowdisplayskip}{0ex} \setlength{\belowdisplayshortskip}{0ex}
\setlength{\abovedisplayskip}{1ex} \setlength{\abovedisplayshortskip}{1ex}
\begin{equation}
    \operatorname*{maximize}_{x \in \mathcal{X}} f(x).
\end{equation}
}%

We assume a strict oracle evaluation budget $N_{\text{budget}}$.
After $t$ oracle calls, the optimization history is
$\mathcal{D}_t = \{(x_i, y_i)\}_{i=1}^t$.
A design goal of \ourmethod{} is \emph{domain generality}: the same optimization loop applies across biological modalities (\textit{e.g.}, peptides, small molecules, or other structured discrete objects). Switching domains amounts to changing the domain description in natural language (\textit{e.g.}, ``SMILES string'' $\leftrightarrow$ ``peptide sequence'') and lightweight validity filters. We provide further discussion of applying \ourmethod{} to new domains in \cref{sec:domains}.

\vspace{-1ex}
\subsection{Distilling data into context}
\vspace{-1ex}
\label{sec:coverage_sampling}

LLMs need a compact representation of the current optimization state that reflects both successful and attempted solutions. From the dataset $\mathcal{D}_{t}$, we construct a straightforward global context $C_{\text{global}}$ to meet these needs:
\begin{enumerate}[itemsep=.1ex,leftmargin=3ex]
\vspace{-1ex}
    \item[\ding{182}] Include the top-$k$ best candidates as positive exemplars.
    \item[\ding{183}] Sample additional candidates from the remaining history by sorting them according to their rank and selecting them uniformly with a random initial offset.
\vspace{-1ex}
\end{enumerate}
Step \ding{183} ensures the context spans the full performance spectrum.
This context includes successful and unsuccessful designs, allowing agents to infer implicit structure---activity relationships and propose edits that meaningfully change candidate scores. 
% ================================================================

\vspace{-1ex}
\subsection{Global search strategies}
\vspace{-1ex}

Black-box optimization typically requires
(1) \textbf{global exploration} to find new high-potential regions of $\mathcal{X}$ and
(2) \textbf{local exploitation} to iteratively refine candidates once a promising direction is found.  
We introduce one agent for each role, using chain-of-thought--finetuned models that can analyze the global optimization state and apply scientific knowledge captured during training. We use \texttt{Intern-S1}~\citep{InternS1} for both:
\begin{enumerate}[itemsep=.1ex,leftmargin=3ex,parsep=1ex]
    \vspace{-1ex}
    \item \textbf{An \texttt{Explorer} Agent} (Global Search). This directly proposes new candidates based on the global optimization state and inferred structure–activity trends. 
    It is also explicitly encouraged to explore new structures.

    \item A \textbf{\texttt{Planner} Agent} (Local Strategy Design). This agent maintains a memory of local search tactics and develops new ones from the global state and this memory. The model’s scientific understanding helps it propose and prioritize strategies that are more likely to yield functional improvements. Example tactics proposed by the Planner can be found in \Cref{app:planner_generated_tasks_ex}.
\end{enumerate}

\vspace{-1ex}
\subsubsection{Explorer}
\vspace{-1ex}

The Explorer Agent receives the global context $C_{\text{global}}$ and directly proposes a batch of candidates:
\begin{equation}
    B_{\text{global}} \leftarrow \textsc{ExplorerAgent}(C_{\text{global}}).
\end{equation}
See \cref{app:explorer_prompt_ex} for example Explorer Agent prompts.

In isolation, the \texttt{Explorer} most closely resembles previous LLM-based optimization methods: it reviews a summary of the current data and suggests new molecules. 
It suggests molecules iteratively and receives feedback until it has failed to improve the objective \texttt{MAX\_FAILS} times in a row.

\vspace{-1ex}
\subsubsection{Planner}
\vspace{-1ex}
\label{sec:planner_agent}

Global search can discover promising input regions, but it is inefficient for fine-grained refinement. To focus local edits, we use a second global agent, the \texttt{Planner}. The \texttt{Planner} agent designs and selects local search strategies expressed as natural language \emph{tasks}.

The \texttt{Planner} has access to a \textbf{Task Registry} $\mathcal{R}$: a bounded library of task prompts specifying reusable mutation operators or refinement heuristics (\textit{e.g.}, ``increase hydrophobicity while preserving motif'' for sequences, or ``replace unstable functional group'' for molecules). Each task in the registry tracks the following performance statistics:
\begin{itemize}[leftmargin=*,noitemsep,nolistsep]
    \item \texttt{attempts}: number of times the task has been executed;
    \item \texttt{successes}: number of executions that produced at least one improving candidate;
    \item \texttt{success rate}: successes / attempts.
\end{itemize}
The registry is initialized with a small set of domain-specific ``default" tasks that represent reasonably simple modification strategies for the given domain and provide examples that help the \texttt{Planner} learn the structure/style of effective task prompts. 
See \cref{app:task_registry_init_ex} for example ``default" tasks. 

Given $C_{\text{global}}$ and a performance summary, the \texttt{Planner} returns a set $\mathcal{P}_{\text{work}}$ of task prompts to execute:
\begin{equation}
    \mathcal{P}_{\text{work}} \leftarrow \textsc{PlannerAgent}(C_{\text{global}}, \mathcal{R}).
\end{equation}
At each round, the \texttt{Planner} is prompted to output ``8--10 tasks total,'' including ``2--3 exploitation tasks (targeted at patterns you observed)", ``2--3 exploration tasks (creative, untried modification types)", and ``2--4 reliable existing tasks that have ($>0$\%) success rates".
See \cref{app:planner_prompt_ex} for examples of full \texttt{Planner} prompts combining $C_{\text{global}}$ and $\mathcal{R}$.

The registry has a fixed maximum size. When adding a new task would exceed capacity, \ourmethod{} prunes the worst-performing non-default task (by success rate with tie-breaking on attempts), ensuring the library remains compact while preserving effective tactics.

% ================================================================
\vspace{-1ex}
\subsection{Executing local search strategies (Worker)}
\vspace{-1ex}
\label{sec:local_search}

Each task prompt $p \in \mathcal{P}_{\mathrm{work}}$ generated by the \texttt{Planner} is fed to independent \texttt{Worker} agents.
The goal of each \texttt{Worker} agent is to use the strategy they are provided to locally improve a single ``seed input'' from the current history $\mathcal{D}_t$---$K$ tasks and $M$ seed inputs, thus resulting in $K\times M$ worker agent runs per optimization round. 

\vspace{-1ex}
\paragraph{Seed input selection.}
A set of $M$ seed candidates $S_{\mathrm{seed}} = \{x^{(1)}, \ldots, x^{(M)}\}$ is chosen from $\mathcal{D}_t$ via greedy selection under a diversity threshold, using a domain-specific distance function $\operatorname{dist}(\,\cdot\, , \,\cdot\,)$ (\textit{e.g.}, normalized edit distance for sequences or fingerprint distance for molecules).
This ensures local search is initiated from distinct regions rather than near-duplicates.

\vspace{-1ex}
\paragraph{Worker hill climbing.} Each worker aims to optimize its seed using the given strategy until progress plateaus. A \texttt{Worker} starts with its seed molecule as $x_{\mathrm{curr}}$ and is repeatedly prompted to improve this molecule with the planner strategy $p$, producing a batch:
\begin{equation}
    B \leftarrow \textsc{WorkerAgent}(p, x_{\mathrm{curr}}),
\end{equation}
See \cref{app:worker_prompt_ex} for example \texttt{Worker} prompts that combine $p$ and $x_\mathrm{curr}$.
The batch is validated, deduplicated, and used to update $x_{\mathrm{curr}}$ if any member improves performance. A \texttt{Worker} terminates after \texttt{MAX\_FAILS} consecutive failures to improve $x_{\mathrm{curr}}$, and the outer loop in \autoref{fig:diagram} ends when all workers have terminated. Outcomes are recorded in the task registry $\mathcal{R}$, allowing the \texttt{Planner} to allocate more budget to productive local strategies.

% ================================================================
\vspace{-1ex}
\subsection{Validation, deduplication, and execution}
\vspace{-1ex}
\label{sec:constraints}

LLMs may generate candidates that are invalid, duplicates, or violate domain-specific constraints. \ourmethod{} post-processes all agent outputs before evaluation.
This post-processing step parses candidates in free-form text, enforces domain validity (\textit{e.g.}, alphabet constraints or canonicalization), scores duplicates by memoization, and applies hard feasibility filters.

Biological design problems often require constraints such as synthesizability or similarity to a known template.
\ourmethod{} supports both soft and hard constraints:
\begin{itemize}[leftmargin=*,noitemsep,nolistsep]
\item
Soft constraints are encoded into the prompts of the \texttt{Explorer} and \texttt{Worker} to nudge generation towards feasible regions;
\item
Hard constraints are encoded by an indicator function, while infeasible candidates are rejected before evaluation.
\end{itemize}

\vspace{-1ex}
\subsection{Diverse portfolio optimization}
\vspace{-1ex}
\label{sec:diverse_portfolio}

For biological design, it is often beneficial to obtain multiple solutions, since a single solution may later prove infeasible due to failure risks (\textit{e.g.}, toxicity, poor manufacturability). Instead, practitioners often require a diverse \emph{portfolio} of many strong candidates~\citep{robot, apexgo}. \ourmethod{} supports this with a straightforward adaptation to optimize an aggregate score over a diverse set:
{%
\setlength{\belowdisplayskip}{0ex} \setlength{\belowdisplayshortskip}{0ex}
\setlength{\abovedisplayskip}{1ex} \setlength{\abovedisplayshortskip}{1ex}
\begin{equation}
\begin{split}
    \operatorname*{\arg\max}_{S\subset\mathcal{X}, |S|=M} \operatorname*{Agg}\big(\{f(x): x \in S\}\big) 
    \qquad
    \text{subject to}
    \qquad 
    \operatorname{dist}(x_i, x_j) \ge \beta \quad \forall i\neq j,
\end{split}
\end{equation}
}%
where $\operatorname{Agg}$ is an aggregation function (\textit{e.g.}, the average), and $\operatorname{dist}$ is any domain-specific measure of diversity, \textit{e.g.}, sequence edit distance and molecular fingerprint distance. 

\vspace{-1ex}
\subsection{Extensibility via external tools}
\vspace{-1ex}
External tools can be incorporated into \ourmethod{} without modifying the core optimization loop (Algorithm~\ref{alg:agent_opt}). 
Because each agent operates through natural language prompts, we can add capabilities by granting agents access to callable tools (\textit{e.g.}, retrieval systems, property predictors, simulation interfaces) \cite{toolformer} that augment reasoning with external knowledge. 
Tools can be assigned to any agent as needed; for instance, global exploration often benefits from giving the \texttt{Explorer} access to knowledge tools. 
In \cref{sec:exp_ablation}, we show that a literature retrieval tool improves optimization performance on molecular design tasks, illustrating how domain knowledge can be incorporated.

\vspace{-1ex}
\section{Experiments}
\vspace{-1ex}
\label{sec:experiments}
We evaluate \ourmethod{} on two biological design domains: (i) small-molecule optimization on GuacaMol benchmark tasks~\citep{GuacaMol} and (ii) antimicrobial peptide (AMP) design using a black-box minimum inhibitory concentration (MIC) oracle~\citep{apex1}. Our experiments are designed to answer:
(1) How does the base version of \ourmethod{}, without task descriptions or other tools, perform relative to state-of-the-art molecular optimization baselines? 
(2) How do the different components of \ourmethod{} contribute to its final performance?
(3) Does \ourmethod{} benefit from agentic capabilities such as task descriptions for grey-box optimization and tools like literature search?
(4) Can \ourmethod{} handle extensions without substantial system modification: constraints and diverse portfolio optimization?

The code and data for reproducing experiments are available at \url{https://anonymous.4open.science/r/agentic-biological-design-753B/}.

\vspace{-1ex}
\subsection{Experimental setup}
\label{sec:exp_setup}

\vspace{-1ex}
\paragraph{Small-molecule benchmark tasks.}
For molecule optimization, we evaluate on 10 GuacaMol tasks~\citep{GuacaMol}, selected from among the ``challenging'' tasks that prior work does not already achieve perfect scores on (1.0). These tasks involve optimizing multi-property objectives over the discrete space of valid SMILES strings.

\vspace{-1ex}
\paragraph{AMP design tasks.}
For peptides, we optimize antimicrobial peptides (AMPs) to \emph{minimize} the mean predicted MIC (lower is better) against the panel of 11 pathogenic bacteria listed in \cref{tab:bacteria}. We use the APEX 1.1 model~\citep{apex1}, an in-silico predictor of MIC, as our black-box objective function. We report the best mean predicted MIC found so far over the evaluation budget.

\vspace{-1ex}
\paragraph{Baselines.}
We compare \ourmethod{} to strong baselines appropriate for each domain and setting. In all plots, \ourmethod{} and all baseline methods show the mean and standard error over $10$ runs unless otherwise noted. 

\begin{itemize}[itemsep=0ex,leftmargin=3ex]
    \vspace{-1ex}
    \item \textbf{GuacaMol baselines.} We compare to recent state-of-the-art methods from Bayesian optimization (NF-BO~\citep{r2-cite9-lsbo-nfbo}), reinforcement learning (GEGL~\citep{GEGL}), and LLM-enhanced optimization (AlphaEvolve~\citep{AlphaEvolve}, LLAMBO~\citep{llambo}, MOLLEO~\citep{MOLLEO}, BOPRO~\citep{bo_w_llms}, and LICO~\citep{lico}).
    Since LICO reports AUC Top-10 under the PMO protocol~\citep{pmo}, we compute the same metric from \ourmethod{}'s trajectories and report the comparison separately in \cref{tab:lico_comparison}, where \ourmethod{} outperforms LICO on all 10 GuacaMol tasks at both 1K and 10K evaluations. 
    For BOPRO, BOPRO-L and BOPRO-S signify runs with the Mistral-Large and Mistral-Small LLMs respectively~\citep{mistral_large_instruct_2411, mistral_small_24b}. Since BOPRO-L is expensive ($\sim600$ USD per task), we only run BOPRO-L on 3/10 tasks. We also compare to ``Random": random sampling from the SELFIES-VAE~\citep{lolbo}.
    Additionally, in \cref{tab:top1_results_1}, we compare against 25 additional baseline methods curated by the recent GuacaMol benchmarking study of \citet{eff_matters_benchmark}, and refer readers there for full citations and discussion of each method. 
    
    \item \textbf{AMP baselines.} We compare to APEX-GO~\citep{apexgo} (latent-space Bayesian optimization), HydrAMP~\citep{HydrAMP} (AMP generation with a conditional variational autoencoder (VAE)), and PepDiffusion~\citep{PepDiffusion} (AMP generation with latent diffusion). (See Figure~\ref{fig:amp_all}.)
\end{itemize}

\paragraph{Implementation details.}
Across all experiments, \ourmethod{} optimizes under a strict evaluation budget of $20{,}000$ oracle calls. Each evaluation corresponds to one valid candidate design (SMILES string or peptide sequence) after filtering for validity and novelty. Further implementation details are provided in \cref{sec:app_imp_details}. 

LLM outputs undergo deterministic post-processing before oracle calls: candidates are parsed from raw text generations, filtered for domain validity (e.g., amino-acid alphabet constraints, SMILES canonicalization). To ensure the oracle budget is not spent on infeasible or duplicate designs, duplicates are removed (both within-batch and against history), and hard constraints are enforced via rejection. When comparing to prior work in~\autoref{fig:guacamol_results}, \ourmethod{} \textbf{does not} access task descriptions or literature search tools; models see only molecules and their numeric scores. 

\subsection{Molecule optimization on GuacaMol}
\label{sec:exp_molecules}

\begin{figure*}[!t]
    \vspace{-4ex}
    \centering
    \centerline{\includegraphics[width=\columnwidth]{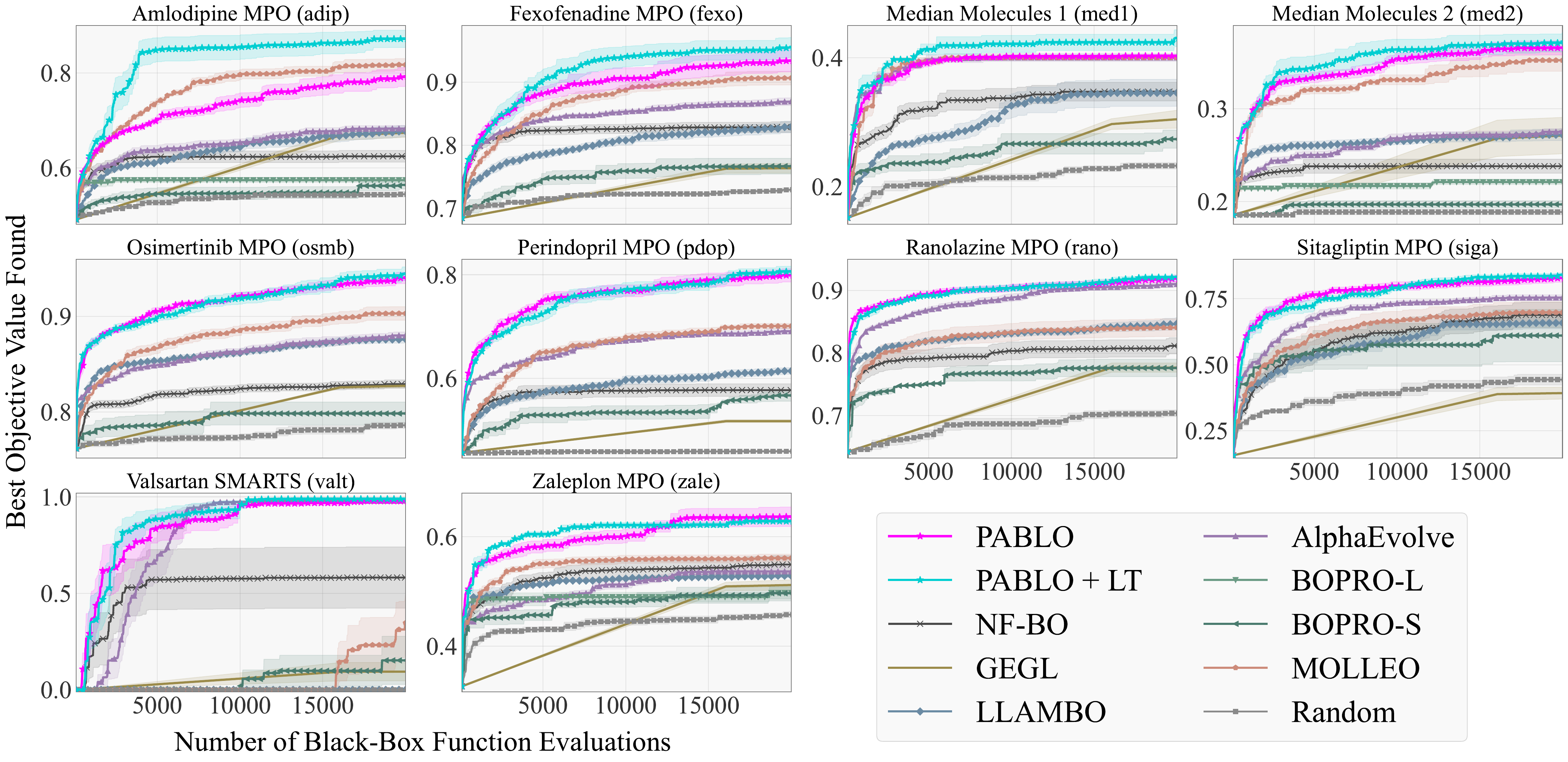}}
    \vspace{-1ex}
    \caption{
    GuacaMol optimization results on 10 tasks. 
    \ourmethod{} achieves state-of-the-art performance.
    } 
    \label{fig:guacamol_results}
    \vspace{-3ex}
\end{figure*}

\begin{wrapfigure}{r}{0.53\textwidth}
  \vspace{-8ex}
  \begin{center}
        \includegraphics[width=0.55\columnwidth]{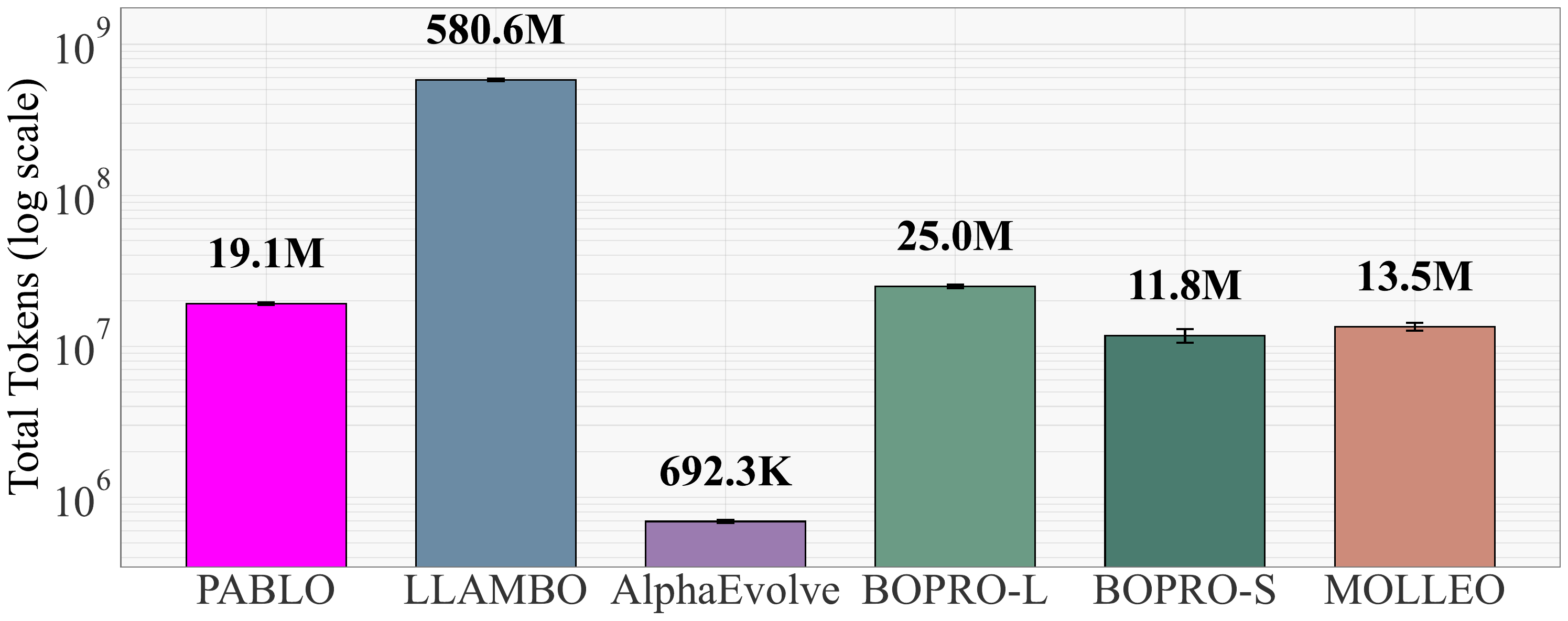}
  \end{center}
    \vspace{-1ex}
    \caption{
    Average number of LLM tokens used per run by \ourmethod{} and other LLM-based baselines. 
    }
    \label{fig:token_use_bar}
    \vspace{-3ex}
\end{wrapfigure}
Figure~\ref{fig:guacamol_results} shows optimization trajectories on 10 GuacaMol tasks. \ourmethod{} improves faster and more consistently reaches higher final objective values than all baselines, and is clearly outperformed on only one task (\texttt{adip}). \Cref{fig:token_use_bar} compares the average LLM tokens used per run for all LLM-based methods. AlphaEvolve uses very few LLM tokens overall, calling an LLM only occasionally to tweak its candidate-generating program. Among methods that use LLMs in the loop, \ourmethod{}'s token usage is roughly average, despite relying on LLMs exclusively.
We provide a detailed discussion of wall-clock runtime, hosted inference cost, and practical scalability trade-offs in \cref{app:limits,tab:llm_runtime_cost}.

\paragraph{Aggregate performance and literature comparison.}
To contextualize these results against a broader benchmark suite, Table~\ref{tab:top1_results_1} reports Top-1 performance at 10K evaluations and compares \ourmethod{} against 26 additional literature baselines reported by \citet{eff_matters_benchmark}. Compared against these baselines, \ourmethod{} ranks first in every task, often by substantial margins.

\begin{wrapfigure}{r}{0.53\textwidth}
\vspace{-6ex}
\centering
\caption{
Benchmark on 10 GuacaMol tasks.
We report the mean and standard deviation of top-1 molecules from 5 independent runs at a 10K evaluation budget following Table 20 in \citet{eff_matters_benchmark}. We compare 26 methods in total and show the top 3 below, with the full table in Appendix~\ref{sec:appendix-results}. 
}
\label{tab:top1_results_1}
\resizebox{0.5\columnwidth}{!}{
\begin{tabular}{l|ccc}
\toprule
\textbf{Task} & \textbf{\ourmethod{}} & \textbf{Graph GA} & \textbf{REINVENT} \\
\midrule
med1 & 0.398\scriptsize{$\pm$0.034} & 0.350\scriptsize{$\pm$0.050} & \textbf{0.399\scriptsize{$\pm$0.058}} \\
med2 & \textbf{0.339\scriptsize{$\pm$0.035}} & 0.324\scriptsize{$\pm$0.040} & 0.332\scriptsize{$\pm$0.045} \\
pdop & \textbf{0.772\scriptsize{$\pm$0.041}} & 0.625\scriptsize{$\pm$0.054} & 0.644\scriptsize{$\pm$0.071} \\
osmb & \textbf{0.919\scriptsize{$\pm$0.017}} & 0.880\scriptsize{$\pm$0.029} & 0.909\scriptsize{$\pm$0.040} \\
adip & \textbf{0.853\scriptsize{$\pm$0.080}} & 0.783\scriptsize{$\pm$0.078} & 0.735\scriptsize{$\pm$0.086} \\
siga & \textbf{0.773\scriptsize{$\pm$0.043}} & 0.689\scriptsize{$\pm$0.214} & 0.080\scriptsize{$\pm$0.034} \\
zale & \textbf{0.606\scriptsize{$\pm$0.011}} & 0.421\scriptsize{$\pm$0.086} & 0.478\scriptsize{$\pm$0.150} \\
valt & \textbf{0.951\scriptsize{$\pm$0.079}} & 0.000\scriptsize{$\pm$0.000} & 0.197\scriptsize{$\pm$0.382} \\
rano & \textbf{0.893\scriptsize{$\pm$0.015}} & 0.810\scriptsize{$\pm$0.072} & 0.865\scriptsize{$\pm$0.068} \\
fexo & \textbf{0.934\scriptsize{$\pm$0.045}} & 0.845\scriptsize{$\pm$0.053} & 0.910\scriptsize{$\pm$0.073} \\
\midrule
\textbf{Sum} & \textbf{7.439} & 5.727 & 5.549 \\
\textbf{Rank} & 1 & 2 & 3 \\
\bottomrule
\end{tabular}
}
\vspace{-4ex}
\end{wrapfigure}

\vspace{-1ex}
\subsection{\ourmethod{} extensions}
\vspace{-1ex}

In \cref{fig:guacamol_ablation} and \cref{tab:ablation}, we study two optional extensions to PABLO: a literature retrieval tool (LT) and task awareness (TA).

\vspace{-1ex}
\paragraph{Extension 1: literature tool (LT).}
We equip the \texttt{Explorer} with a literature retrieval tool (LT) that implements a retrieval-augmented generation (RAG) pipeline: given a single molecule (a SMILES string chosen by the agent), it returns a structured dictionary of context from papers that either explicitly mention the queried molecule or highly similar molecules found via fingerprint similarity search. Examples of queries and retrieved context are in \cref{fig:mol-rag}. We do not modify the architecture of \ourmethod{}; the tool is simply provided as an additional resource. As shown in \cref{fig:guacamol_results} (\ourmethod{} + LT curves) and \cref{tab:ablation}, enabling LT yields clear improvements on several tasks.

\vspace{-1ex}
\paragraph{Extension 2: task awareness (TA).}
Task awareness (TA) gives agents an explicit natural-language description of the optimization objective, which we append to the prompts of both the \texttt{Explorer} and \texttt{Planner}. For example, the Amlodipine MPO task is: \emph{“Maximize similarity to amlodipine while having exactly 3 total rings.”} As shown in \cref{fig:guacamol_ablation} and \cref{tab:ablation}, TA makes many GuacaMol optimization tasks nearly trivial: Intern S1 can directly propose structures aligned with the scoring logic, often reaching near-optimal solutions within a few hundred function evaluations. Because of this, all primary PABLO results in \cref{fig:guacamol_results}, \cref{tab:top1_results_1}, etc. are reported \emph{without} task descriptions, matching the standard black-box benchmark assumption. \texttt{+ TA} is only used where explicitly indicated.

\vspace{-1ex}
\subsection{\ourmethod{} ablations}
\vspace{-1ex}
\label{sec:exp_ablation}

\begin{wrapfigure}{r}{0.5\textwidth}
    \vspace{-8ex}
    \begin{center}
        \includegraphics[width=0.5\columnwidth]{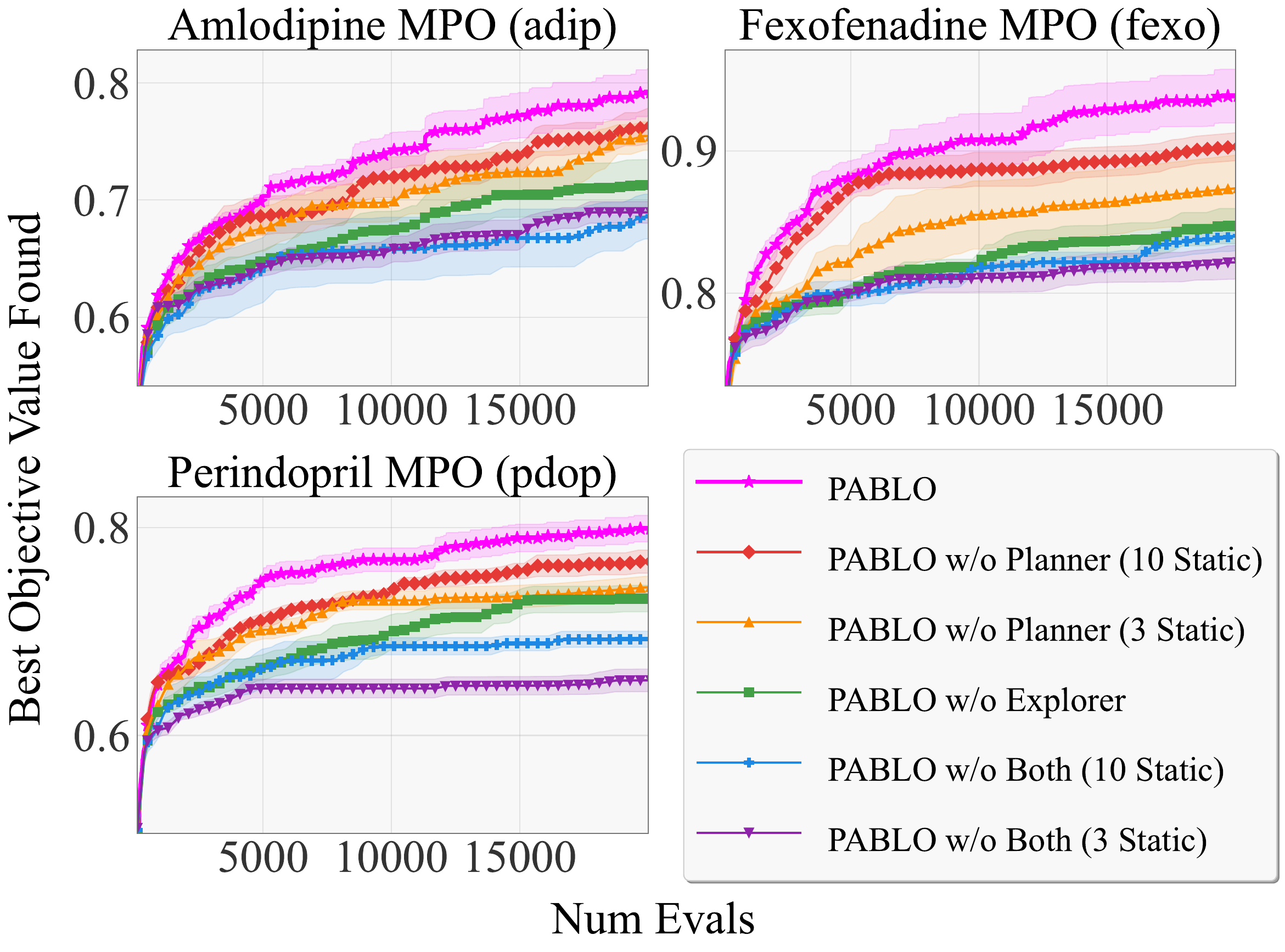}
    \end{center}
    \vspace{-3ex}
    \caption{
    Ablations on representative GuacaMol tasks showing the contribution of the Planner and Explorer Agents. 
    }
    \label{fig:agents_ablation}
    \vspace{-4ex}
\end{wrapfigure}
We ablate the major components of \ourmethod{} here--see \cref{sec:appendix-ablations} for additional studies of \ourmethod{}-hyperparameters. 

In \cref{fig:agents_ablation}, we ablate major components of \ourmethod{} on representative GuacaMol MPO tasks. Removing the \texttt{Explorer} (\textit{\ourmethod{} w/o Explorer}) consistently degrades performance, showing that global, context-driven exploration is crucial for rapidly locating promising regions.
We also ablate the \texttt{Planner} by replacing it with fixed \texttt{Worker} prompts: (i) \textit{3 Static Worker Prompts} from the default Task Registry and (ii) \textit{10 Static Worker Prompts}, which add seven hand-designed molecule-editing prompts. These static-prompt baselines are competitive, but \ourmethod{} with the Planner Agent consistently achieves higher final performance and better sample efficiency. The Planner Agent even outperforms the 10-prompt baseline, showing that \texttt{Planner} prompts are more effective than a fixed set. The static prompts are listed in \cref{sec:app_static_worker_prompts}. Removing \emph{both} the \texttt{Explorer} and \texttt{Planner} (\textit{\ourmethod{} w/o Both}) gives the worst performance, underscoring their complementarity.
\vspace{-1ex}
\subsection{Peptide optimization (AMP design)}
\vspace{-1ex}
\label{sec:exp_peptides}

\begin{wrapfigure}{r}{0.5\textwidth}
  \vspace{-10ex}
  \centering
  \begin{center}
      \includegraphics[width=0.5\columnwidth]{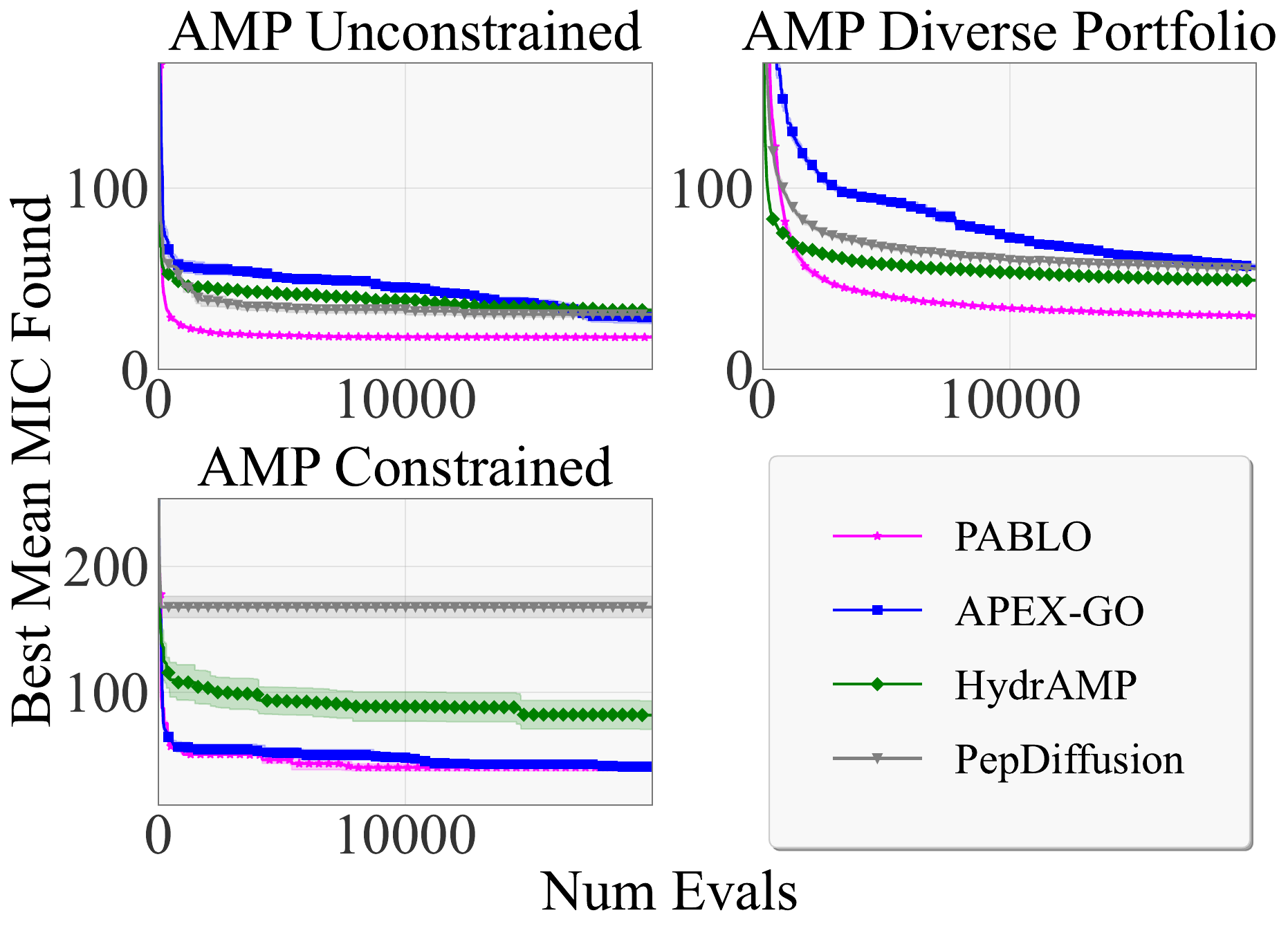}
  \end{center}
  \vspace{-2ex}
  \caption{
Antimicrobial peptide (AMP) optimization. We plot predicted MIC versus black-box evaluations (lower MIC is better). (\textbf{Upper Left}): The predicted MIC of the best peptide found so far. (\textbf{Lower Left}): Template-constrained optimization; we show the predicted MIC of the best ``feasible" peptide found so far. (\textbf{Upper Right}): Template-free optimization of a diverse portfolio of 20 AMPs; we show the mean predicted MIC of the best sufficiently-diverse portfolio so far.
  }
  \label{fig:amp_all}
  \vspace{-6ex}
\end{wrapfigure}
Figure~\ref{fig:amp_all} Upper Left Panel shows AMP optimization. \ourmethod{} improves rapidly and consistently identifies peptides with substantially lower predicted MIC than all baselines under the same evaluation budget.

\vspace{-1ex}
\subsection{Constraints and diverse portfolio optimization}
\vspace{-1ex}
\label{sec:exp_extensions}

\paragraph{Template-constrained AMP optimization.}
We evaluate \ourmethod{} under a realistic hard constraint: candidates must be at least 75\% similar to one of 10 ``trusted'' template peptides, simulating lead optimization rather than de novo design. These 10 templates, mined from extinct organisms and selected by \citet{apex1}, were chosen because extinct-like peptides may better evade antibiotic resistance in modern bacteria. As shown in Figure~\ref{fig:amp_all} (Lower Left), \ourmethod{} maintains its advantage under these template constraints, indicating that prompt modifications and rejection are moderately sufficient to handle constraints.
\vspace{-1ex}
\paragraph{Diverse portfolio optimization.}
We consider portfolio optimization—targeting simultaneous optimization of 20 peptides—to reduce downstream failure risk when optimizing against an \textit{in silico} predictor. We impose a pairwise diversity constraint requiring all pairs in the 20-AMP portfolio to have at least 0.75 dissimilarity by edit distance. Figure~\ref{fig:amp_all} (upper right) shows that \ourmethod{} substantially improves portfolio quality (mean MIC over the best diverse set) compared to baselines under the same budget.
\begin{wrapfigure}{r}{0.5\textwidth}
  \begin{center}
      \includegraphics[width=0.5\columnwidth]{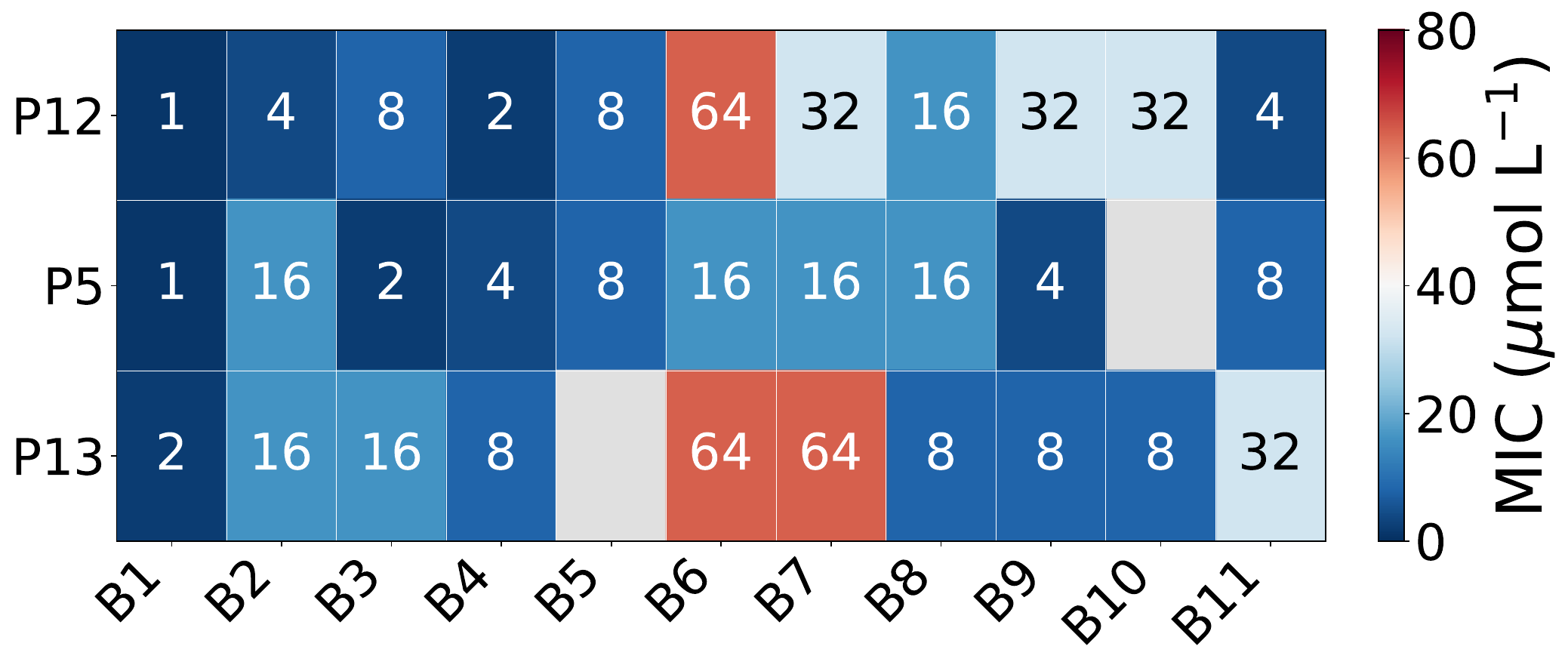}
  \end{center}
  \vspace{-2ex}
  \caption{
\textit{In vitro} MIC results against the 11 target bacteria (B1–B11; see \cref{tab:bacteria}) achieved by the three best-performing peptides from the $M=20$-peptide portfolio produced by one run of \ourmethod{} on the AMP design task. Peptide sequences are listed in \cref{tab:optimized_peptides}. For complete \textit{in vitro} results see \cref{fig:heatmap_20_targets}.
}
\label{fig:heatmap_condensed}
  \vspace{-6ex}
\end{wrapfigure}
\vspace{-1ex}
\paragraph{\textit{In vitro} performance.} To validate that a diverse portfolio might lead to strong AMPs, we performed \textit{in vitro} experiments on a portfolio of peptides produced by one run of \ourmethod{}. We refer to this portfolio as P1–P20 (sequences and APEX 1.1–predicted MICs are listed in \cref{tab:optimized_peptides}). \textit{In vitro} experimental procedures are detailed in \cref{sec:in-vitro}.

\cref{fig:heatmap_condensed} presents measured \textit{in vitro} MIC values against the 11 target bacteria (B1–B11; see \cref{tab:bacteria}) for the three best-performing peptides in the portfolio, ranked by lowest average \textit{in vitro} MIC across B1–B11. Full \textit{in vitro} results for all 20 peptides (P1–P20) are provided in \cref{fig:heatmap_20_targets}.

Although all 20 peptides had APEX 1.1–predicted MICs below $30\, \si{\micro\mole\per\liter}$, several showed weak or no detectable experimental activity (gray cells in \cref{fig:heatmap_20_targets}), illustrating the known gap between prediction and biological performance. Nonetheless, for every target species, at least one peptide achieved strong inhibition (MIC $\leq 16$ $\si{\micro\mole\per\liter}$).
\paragraph{Held-out bacteria.} The full heatmap in \cref{fig:heatmap_20_targets} also shows in vitro results for nine additional bacteria (B12–B20; see \cref{tab:extra_bacteria}) that were not part of the optimization objective. Several peptides still showed strong activity against these held-out targets, indicating broad-spectrum activity.

\section{Related work}
\label{sec:related}

\paragraph{Biological design as black-box optimization.}
The discovery of novel biological entities—such as small molecules, peptides, or proteins—can be formalized as a maximization problem of some objective function over a discrete structured space $\mathcal{X}$ \citep{coley2020autonomous}. We seek $x^* = \text{argmax}_{x \in \mathcal{X}} f(x)$, where $f$ might measure binding affinity, solubility, or drug-likeness. The machine learning literature offers a diverse set of strategies to solve these biological optimization problems:

\textbf{Virtual screening:} High-throughput approaches evaluate fixed libraries of compounds. Recent methods like MolPAL \citep{MolPAL} and active learning strategies \citep{Graff2021-wr} iterate between screening and model updating to identify top candidates from pool-based libraries.
    
\textbf{Genetic algorithms (GAs):} GAs explore $\mathcal{X}$ via stochastic crossover and mutation. These include string-based methods like SMILES-GA \citep{SMILES_GA} and STONED \citep{STONED}, as well as graph-based approaches like Graph GA \citep{Graph_GA_AND_MCTS}. Recent work has also explored augmenting GAs with neural networks \citep{GA_D} or combining them with synthesis trees \citep{SynNet}.
    
\textbf{Bayesian optimization (BO):} BO builds a probabilistic surrogate model to guide acquisition \citep{frazier2018tutorial, shahriari2015taking, garnett2023bayesian}. To handle discrete biological data, recent methods operate either directly over string spaces (e.g., BOSS \citep{BOSS}) or over continuous latent spaces learned by generative models (Latent Space BO) \citep{Weighted_Retraining,ladder,lambo-no-llm,r2-cite10-lsbo-cobo,r2-cite5-lsbo,r2-cite6-lsbo,r2-cite8-lsbo,r2-cite7-lsbo-gvae,lolbo,robot,r2-cite9-lsbo-nfbo}.
    
\textbf{Deep generative models (DGMs):} These models learn a distribution over valid structures. Variational Autoencoders (VAEs) map discrete structures to continuous latent spaces for optimization, including the seminal SMILES-VAE \citep{gomez2018automatic}, Junction Tree VAE \citep{JTVAE}, and SELFIES-based models \citep{lolbo}. Recent advances also include Normalizing Flows \citep{r2-cite9-lsbo-nfbo}, Diffusion Models \citep{ho2021classifierfree}, and score-based modeling like MARS \citep{MARS} and GFlowNets \citep{GFlowNet}.
    
\textbf{Reinforcement learning (RL):} RL agents learn policies to construct molecules step-by-step, as seen in methods like REINVENT, MolDQN, and GEGL ~\citep{REINVENT, Olivecrona2017-es,MolDQN,GEGL}. 

\paragraph{LLM-enhanced optimization.}
A growing body of work integrates Large Language Models (LLMs) into optimization pipelines. Evolutionary methods such as AlphaEvolve \citep{AlphaEvolve} and related approaches \citep{MOLLEO} use LLMs as intelligent variation operators.
Recent BO and other surrogate-guided methods embed LLMs into optimization pipelines to provide structural priors, propose candidates for acquisition-driven search, or serve as surrogate models \citep{bo_w_llms, bo_w_llms_2, llambo, ChemBOMAS, lico}.

Many works study \emph{LLMs as optimizers}, using iterative prompting and textual feedback loops to optimize black-box objectives \citep{opro, towards_optimizing_llms, gptopt}. They are usually tested on prompt optimization, math problems, or generic black-box functions, not biological design tasks with structured chemical or sequence spaces. 
Related work couples LLM reasoning with standard optimizer infrastructure for LLM-guided black-box optimization \citep{llmbox}, but typically embeds the LLM in a fixed scaffold (e.g., replacing a random mutation operator) or simple iterative proposal loops without explicit hierarchical strategy roles, rather than giving it autonomy to plan and revise the search strategy. 
Recent chemistry-focused agent frameworks show the promise of tool-using, stepwise-planning LLM agents for multi-step planning and iterative refinement \citep{chemcrow, autonomous_chem_research, auditable_agent_mol_opt,ElAgenteMatter2025,RAMOS20252514,saga}. We extend this paradigm to \emph{black-box optimization for design}.

\section{Discussion}
\label{sec:discussion}
Our results demonstrate that a relatively simple agentic factorization of black-box optimization can achieve state-of-the-art performance on standard molecular design benchmarks, outperforming even fairly complex state-of-the-art baselines.
The success of \ourmethod{} suggests that LLMs are becoming sufficiently knowledgeable and capable of step-by-step reasoning, making them competitive with manually-defined optimization policies. Moreover, this approach is compelling because it easily enables capabilities that can be difficult to incorporate into more purpose-built methods:
\begin{enumerate}[noitemsep,nolistsep,leftmargin=*,label=$\bullet$]
    \item Semantic Task Awareness: Accelerate search using natural language descriptions of the objective.
    \item External Knowledge Integration: Seamlessly incorporating retrieval-augmented generation (RAG) tools to ground optimization in scientific literature.
    \item Flexible Constraint Handling: Addressing complex, non-differentiable constraints (like peptide template similarity) through simple prompt engineering. 
\end{enumerate}

\bibliographystyle{unsrtnat}
\bibliography{refs}

%%%%%%%%%%%%%%%%%%%%%%%%%%%%%%%%%%%%%%%%%%%%%%%%%%%%%%%%%%%

\newpage
\appendix
\counterwithin{figure}{section} % reset figure numbering for each appendix section
\renewcommand{\thefigure}{\thesection.\arabic{figure}}
\counterwithin{table}{section}
\renewcommand{\thetable}{\thesection.\arabic{table}}
\section{\ourmethod{} pseudocode algorithm}
\label{sec:pablo_algo}
In \cref{alg:agent_opt}, we provide a pseudocode algorithm for \ourmethod{}.

\begin{algorithm}[!ht]
\caption{\ourmethodfullname{}: hierarchical agentic optimization with success persistence and task memory. Each outer iteration alternates global exploration, strategy selection, and local multi-trajectory refinement. All LLM-generated candidates are validated and deduplicated prior to oracle evaluation to avoid wasting budget. In particular, the ``Filter" step filters out all candidates that are already in $\mathcal{D}_t$ (duplicates) or violate domain constraints (invalid).}
\label{alg:agent_opt}
\begin{algorithmic}[1]
\STATE \textbf{Input:} oracle $f$, budget $N_{\text{budget}}$, history $\mathcal{D}_0$, Task Registry $\mathcal{R}$
\WHILE{$|\mathcal{D}_t| < N_{\text{budget}}$}

    \STATE \color{blue} \textit{// Build global context} \color{black}
    \STATE $C_{\text{global}} \leftarrow \textsc{HistoryCoverageSample}(\mathcal{D}_t)$

    \STATE \color{blue} \textit{// Phase I: Global search with persistence} \color{black}
    \STATE $\mathrm{fails} \leftarrow 0$
    \WHILE{$fails < \texttt{MAX\_FAILS}$}
        \STATE $B_{\text{global}} \leftarrow \textsc{ExplorerAgent}(C_{\text{global}})$
        \STATE $\tilde{B}_{\text{global}} \leftarrow \textsc{Filter}(B_{\text{global}}, \mathcal{D}_t)$
        \STATE Evaluate $f(x)$ for $x \in \tilde{B}_{\text{global}}$ and add to $\mathcal{D}_t$
        \IF{$\textsc{ImprovedStatistic}(\mathcal{D}_t)$}
            \STATE $\mathrm{fails} \leftarrow 0$
        \ELSE
            \STATE $\mathrm{fails} \leftarrow \mathrm{fails} + 1$
        \ENDIF
        \STATE $C_{\text{global}} \leftarrow \textsc{HistoryCoverageSample}(\mathcal{D}_t)$
    \ENDWHILE

    \STATE \color{blue} \textit{// Phase II: Planner Agent selects tasks} \color{black}
    \STATE $\mathcal{P}_{\text{work}} \leftarrow \textsc{PlannerAgent}(C_{\text{global}}, \mathcal{R})$

    \STATE \color{blue} \textit{// Phase III: Local multi-trajectory refinement} \color{black}
    \FOR{each task prompt $p \in \mathcal{P}_{\text{work}}$}
        \STATE $S_{\text{seed}} \leftarrow \textsc{SelectDiverseSeeds}(\mathcal{D}_t)$
        \FOR{each seed $x^{(j)} \in S_{\text{seed}}$}
            \STATE $x_{\text{curr}} \leftarrow x^{(j)}$; $\mathrm{fails} \leftarrow 0$
            \WHILE{$\mathrm{fails} < \texttt{MAX\_FAILS}$}
                \STATE $B_{\text{loc}} \leftarrow \textsc{WorkerAgent}(p, x_{\text{curr}})$
                \STATE $\tilde{B}_{\text{loc}} \leftarrow \textsc{Filter}(B_{\text{loc}}, \mathcal{D}_t)$
                \STATE Evaluate $f(x)$ for $x \in \tilde{B}_{\text{loc}}$ and add to $\mathcal{D}_t$
                \IF{$\textsc{ImprovesTrajWithoutCollapse}(\tilde{B}_{\text{loc}})$}
                    \STATE $x_{\text{curr}} \leftarrow \arg\max_{x \in \tilde{B}_{\text{loc}}} f(x)$
                    \STATE $\mathrm{fails} \leftarrow 0$; record success in $\mathcal{R}$
                \ELSE
                    \STATE $\mathrm{fails} \leftarrow \mathrm{fails} + 1$; record failure in $\mathcal{R}$
                \ENDIF
            \ENDWHILE
        \ENDFOR
    \ENDFOR

\ENDWHILE
\end{algorithmic}
\end{algorithm}
\newpage 
\section{Additional experimental results}
\label{sec:appendix-results}
In \cref{tab:top1_results_2}-\cref{tab:top1_results_7}, we provide the remaining columns of \cref{tab:top1_results_1} from the main text. 
\begin{table}[htbp]
\centering
\caption{\cref{tab:top1_results_1} Continued}
\label{tab:top1_results_2}
\resizebox{\textwidth}{!}{
\begin{tabular}{l|cccc}
\toprule
\textbf{Task} & \textbf{REINVENT SELFIES} & \textbf{LSTM HC} & \textbf{STONED} & \textbf{GP BO} \\
\midrule
med1 & 0.399$\pm$0.063 & 0.388$\pm$0.064 & 0.295$\pm$0.036 & 0.345$\pm$0.044 \\
med2 & 0.313$\pm$0.040 & 0.339$\pm$0.049 & 0.265$\pm$0.038 & 0.337$\pm$0.033 \\
pdop & 0.610$\pm$0.070 & 0.568$\pm$0.037 & 0.522$\pm$0.027 & 0.562$\pm$0.036 \\
osmb & 0.878$\pm$0.028 & 0.859$\pm$0.023 & 0.848$\pm$0.024 & 0.837$\pm$0.020 \\
adip & 0.706$\pm$0.068 & 0.739$\pm$0.063 & 0.638$\pm$0.054 & 0.681$\pm$0.067 \\
siga & 0.409$\pm$0.170 & 0.262$\pm$0.079 & 0.526$\pm$0.169 & 0.318$\pm$0.117 \\
zale & 0.441$\pm$0.109 & 0.413$\pm$0.126 & 0.373$\pm$0.088 & 0.269$\pm$0.084 \\
valt & 0.000$\pm$0.000 & 0.000$\pm$0.000 & 0.000$\pm$0.000 & 0.000$\pm$0.000 \\
rano & 0.851$\pm$0.095 & 0.824$\pm$0.073 & 0.862$\pm$0.113 & 0.817$\pm$0.080 \\
fexo & 0.842$\pm$0.044 & 0.818$\pm$0.047 & 0.851$\pm$0.058 & 0.805$\pm$0.053 \\
\midrule
\textbf{Sum} & 5.449 & 5.210 & 5.180 & 4.971 \\
\textbf{Rank} & 4 & 5 & 6 & 7 \\
\bottomrule
\end{tabular}
}
\end{table}
\begin{table}[htbp]
\centering
\caption{(Continued)}
\label{tab:top1_results_3}
\resizebox{\textwidth}{!}{%
\begin{tabular}{l|cccc}
\toprule
\textbf{Task} & \textbf{DoG-Gen} & \textbf{LSTM HC SELFIES} & \textbf{SMILES GA} & \textbf{SynNet} \\
\midrule
med1 & 0.322$\pm$0.053 & 0.362$\pm$0.058 & 0.207$\pm$0.014 & 0.244$\pm$0.019 \\
med2 & 0.297$\pm$0.040 & 0.274$\pm$0.031 & 0.210$\pm$0.009 & 0.259$\pm$0.016 \\
pdop & 0.587$\pm$0.044 & 0.521$\pm$0.028 & 0.459$\pm$0.014 & 0.610$\pm$0.039 \\
osmb & 0.850$\pm$0.028 & 0.832$\pm$0.018 & 0.835$\pm$0.019 & 0.821$\pm$0.016 \\
adip & 0.621$\pm$0.034 & 0.600$\pm$0.012 & 0.570$\pm$0.006 & 0.596$\pm$0.020 \\
siga & 0.252$\pm$0.099 & 0.349$\pm$0.089 & 0.504$\pm$0.145 & 0.067$\pm$0.040 \\
zale & 0.343$\pm$0.111 & 0.360$\pm$0.093 & 0.396$\pm$0.097 & 0.402$\pm$0.059 \\
valt & 0.000$\pm$0.000 & 0.000$\pm$0.000 & 0.000$\pm$0.000 & 0.000$\pm$0.000 \\
rano & 0.823$\pm$0.057 & 0.795$\pm$0.099 & 0.780$\pm$0.082 & 0.783$\pm$0.038 \\
fexo & 0.808$\pm$0.036 & 0.769$\pm$0.039 & 0.771$\pm$0.041 & 0.797$\pm$0.031 \\
\midrule
\textbf{Sum} & 4.903 & 4.862 & 4.732 & 4.579 \\
\textbf{Rank} & 8 & 9 & 10 & 11 \\
\bottomrule
\end{tabular}%
}
\end{table}

\begin{table}[htbp]
\centering
\caption{(Continued)}
\label{tab:top1_results_4}
\resizebox{\textwidth}{!}{
\begin{tabular}{l|cccc}
\toprule
\textbf{Task} & \textbf{MIMOSA} & \textbf{DST} & \textbf{GA+D} & \textbf{VAE BO SELFIES} \\
\midrule
med1 & 0.296$\pm$0.039 & 0.281$\pm$0.036 & 0.219$\pm$0.037 & 0.231$\pm$0.017 \\
med2 & 0.238$\pm$0.016 & 0.201$\pm$0.024 & 0.161$\pm$0.028 & 0.206$\pm$0.006 \\
pdop & 0.557$\pm$0.047 & 0.502$\pm$0.026 & 0.337$\pm$0.147 & 0.482$\pm$0.024 \\
osmb & 0.817$\pm$0.022 & 0.827$\pm$0.018 & 0.784$\pm$0.129 & 0.802$\pm$0.010 \\
adip & 0.594$\pm$0.009 & 0.582$\pm$0.054 & 0.527$\pm$0.124 & 0.593$\pm$0.022 \\
siga & 0.209$\pm$0.085 & 0.205$\pm$0.106 & 0.482$\pm$0.175 & 0.244$\pm$0.083 \\
zale & 0.287$\pm$0.103 & 0.344$\pm$0.119 & 0.359$\pm$0.119 & 0.379$\pm$0.091 \\
valt & 0.000$\pm$0.000 & 0.000$\pm$0.000 & 0.000$\pm$0.000 & 0.064$\pm$0.072 \\
rano & 0.773$\pm$0.139 & 0.752$\pm$0.163 & 0.775$\pm$0.244 & 0.564$\pm$0.065 \\
fexo & 0.743$\pm$0.030 & 0.778$\pm$0.041 & 0.737$\pm$0.174 & 0.707$\pm$0.011 \\
\midrule
\textbf{Sum} & 4.514 & 4.472 & 4.381 & 4.272 \\
\textbf{Rank} & 12 & 13 & 14 & 15 \\
\bottomrule
\end{tabular}
}
\end{table}

\begin{table}[htbp]
\centering
\caption{(Continued)}
\label{tab:top1_results_5}
\resizebox{\textwidth}{!}{
\begin{tabular}{l|cccc}
\toprule
\textbf{Task} & \textbf{MolPAL} & \textbf{MARS} & \textbf{Screening} & \textbf{VAE BO SMILES} \\
\midrule
med1 & 0.309$\pm$0.028 & 0.233$\pm$0.017 & 0.271$\pm$0.029 & 0.267$\pm$0.043 \\
med2 & 0.273$\pm$0.021 & 0.203$\pm$0.015 & 0.244$\pm$0.021 & 0.222$\pm$0.011 \\
pdop & 0.504$\pm$0.020 & 0.488$\pm$0.016 & 0.500$\pm$0.028 & 0.484$\pm$0.028 \\
osmb & 0.816$\pm$0.020 & 0.809$\pm$0.021 & 0.801$\pm$0.016 & 0.801$\pm$0.010 \\
adip & 0.651$\pm$0.043 & 0.546$\pm$0.034 & 0.613$\pm$0.039 & 0.611$\pm$0.036 \\
siga & 0.117$\pm$0.030 & 0.083$\pm$0.037 & 0.142$\pm$0.060 & 0.114$\pm$0.068 \\
zale & 0.286$\pm$0.064 & 0.296$\pm$0.023 & 0.280$\pm$0.101 & 0.139$\pm$0.046 \\
valt & 0.000$\pm$0.000 & 0.000$\pm$0.000 & 0.000$\pm$0.000 & 0.064$\pm$0.077 \\
rano & 0.556$\pm$0.064 & 0.776$\pm$0.050 & 0.532$\pm$0.059 & 0.598$\pm$0.076 \\
fexo & 0.709$\pm$0.006 & 0.755$\pm$0.034 & 0.706$\pm$0.021 & 0.719$\pm$0.016 \\
\midrule
\textbf{Sum} & 4.221 & 4.189 & 4.089 & 4.019 \\
\textbf{Rank} & 16 & 17 & 18 & 19 \\
\bottomrule
\end{tabular}
}
\end{table}

\begin{table}[htbp]
\centering
\caption{(Continued)}
\label{tab:top1_results_6}
\resizebox{\textwidth}{!}{
\begin{tabular}{l|cccc}
\toprule
\textbf{Task} & \textbf{JT-VAE BO} & \textbf{Pasithea} & \textbf{DoG-AE} & \textbf{GFlowNet} \\
\midrule
med1 & 0.212$\pm$0.019 & 0.216$\pm$0.021 & 0.203$\pm$0.014 & 0.237$\pm$0.019 \\
med2 & 0.192$\pm$0.003 & 0.194$\pm$0.006 & 0.201$\pm$0.010 & 0.198$\pm$0.009 \\
pdop & 0.463$\pm$0.019 & 0.447$\pm$0.016 & 0.464$\pm$0.026 & 0.478$\pm$0.021 \\
osmb & 0.800$\pm$0.011 & 0.792$\pm$0.009 & 0.793$\pm$0.026 & 0.817$\pm$0.016 \\
adip & 0.585$\pm$0.000 & 0.585$\pm$0.000 & 0.539$\pm$0.017 & 0.482$\pm$0.016 \\
siga & 0.169$\pm$0.096 & 0.230$\pm$0.085 & 0.039$\pm$0.033 & 0.045$\pm$0.020 \\
zale & 0.302$\pm$0.089 & 0.243$\pm$0.084 & 0.156$\pm$0.093 & 0.118$\pm$0.061 \\
valt & 0.000$\pm$0.000 & 0.064$\pm$0.126 & 0.000$\pm$0.000 & 0.000$\pm$0.000 \\
rano & 0.587$\pm$0.041 & 0.443$\pm$0.054 & 0.744$\pm$0.025 & 0.701$\pm$0.030 \\
fexo & 0.702$\pm$0.016 & 0.707$\pm$0.041 & 0.723$\pm$0.045 & 0.727$\pm$0.017 \\
\midrule
\textbf{Sum} & 4.012 & 3.921 & 3.862 & 3.803 \\
\textbf{Rank} & 20 & 21 & 22 & 23 \\
\bottomrule
\end{tabular}
}
\end{table}

\begin{table}[htbp]
\centering
\caption{(Continued)}
\label{tab:top1_results_7}
\resizebox{0.75\textwidth}{!}{%
\begin{tabular}{l|ccc}
\toprule
\textbf{Task} & \textbf{GFlowNet-AL} & \textbf{Graph MCTS} & \textbf{MolDQN} \\
\midrule
med1 & 0.229$\pm$0.012 & 0.242$\pm$0.023 & 0.188$\pm$0.028 \\
med2 & 0.191$\pm$0.009 & 0.148$\pm$0.010 & 0.108$\pm$0.009 \\
pdop & 0.464$\pm$0.020 & 0.334$\pm$0.038 & 0.282$\pm$0.062 \\
osmb & 0.812$\pm$0.010 & 0.738$\pm$0.018 & 0.699$\pm$0.018 \\
adip & 0.466$\pm$0.016 & 0.483$\pm$0.024 & 0.383$\pm$0.033 \\
siga & 0.028$\pm$0.017 & 0.210$\pm$0.088 & 0.015$\pm$0.009 \\
zale & 0.048$\pm$0.020 & 0.166$\pm$0.065 & 0.042$\pm$0.024 \\
valt & 0.000$\pm$0.000 & 0.000$\pm$0.000 & 0.000$\pm$0.000 \\
rano & 0.705$\pm$0.034 & 0.369$\pm$0.096 & 0.171$\pm$0.077 \\
fexo & 0.732$\pm$0.015 & 0.611$\pm$0.024 & 0.532$\pm$0.039 \\
\midrule
\textbf{Sum} & 3.675 & 3.301 & 2.420 \\
\textbf{Rank} & 24 & 25 & 26 \\
\bottomrule
\end{tabular}%
}
\end{table}

\newpage 
\textcolor{white}{x}
\newpage 
\textcolor{white}{x}
\subsection{LICO comparison}
In \cref{tab:lico_comparison}, we provide a direct empirical comparison to LICO~\citep{lico} on all 10 GuacaMol molecular optimization tasks considered in \cref{sec:experiments}. 
Unlike the other GuacaMol baselines in \cref{sec:experiments}, LICO reports AUC Top-10, an alternative metric used by the PMO benchmark~\citep{pmo}. 
To enable a fair comparison, we compute \ourmethod{}'s AUC Top-10 from the same raw molecule optimization trajectories used to generate \cref{fig:guacamol_results}. 
In \cref{tab:lico_comparison}, we compare against LICO's reported results from Tables 1 and 2 of the LICO paper, corresponding to 1K and 10K evaluation budgets, respectively. 
\ourmethod{} outperforms LICO on all 10 GuacaMol tasks at both budgets.   
\begin{table}[t]
\centering
\caption{Comparison between \ourmethod{} and LICO~\citep{lico} on the PMO~\citep{pmo} AUC Top-10 metric at 1K and 10K evaluations. Higher is better.}
\label{tab:lico_comparison}
\begin{tabular}{lccc}
\toprule
\textbf{Task} & \textbf{\ourmethod{}} & \textbf{LICO} & \textbf{Diff} \\
\midrule
\multicolumn{4}{l}{\textbf{AUC Top-10 @ 1K}} \\
\midrule
adip & $0.578 \pm 0.014$ & $0.541 \pm 0.026$ & $+0.037$ \\
fexo & $0.764 \pm 0.022$ & $0.700 \pm 0.023$ & $+0.064$ \\
med1 & $0.266 \pm 0.012$ & $0.217 \pm 0.019$ & $+0.049$ \\
med2 & $0.259 \pm 0.014$ & $0.193 \pm 0.009$ & $+0.066$ \\
osmb & $0.845 \pm 0.015$ & $0.759 \pm 0.008$ & $+0.086$ \\
pdop & $0.598 \pm 0.014$ & $0.473 \pm 0.009$ & $+0.125$ \\
rano & $0.837 \pm 0.018$ & $0.687 \pm 0.029$ & $+0.150$ \\
siga & $0.457 \pm 0.048$ & $0.315 \pm 0.097$ & $+0.142$ \\
valt & $0.114 \pm 0.161$ & $0.000 \pm 0.000$ & $+0.114$ \\
zale & $0.476 \pm 0.011$ & $0.404 \pm 0.022$ & $+0.072$ \\
\midrule
\textbf{Sum} & \textbf{5.194} & \textbf{4.289} & \textbf{+0.905} \\
\midrule
\multicolumn{4}{l}{\textbf{AUC Top-10 @ 10K}} \\
\midrule
adip & $0.686 \pm 0.020$ & $0.679 \pm 0.027$ & $+0.007$ \\
fexo & $0.863 \pm 0.033$ & $0.772 \pm 0.023$ & $+0.091$ \\
med1 & $0.370 \pm 0.011$ & $0.291 \pm 0.016$ & $+0.079$ \\
med2 & $0.320 \pm 0.015$ & $0.280 \pm 0.019$ & $+0.040$ \\
osmb & $0.896 \pm 0.008$ & $0.820 \pm 0.012$ & $+0.076$ \\
pdop & $0.724 \pm 0.028$ & $0.557 \pm 0.028$ & $+0.167$ \\
rano & $0.886 \pm 0.014$ & $0.774 \pm 0.008$ & $+0.112$ \\
siga & $0.718 \pm 0.041$ & $0.567 \pm 0.034$ & $+0.151$ \\
valt & $0.709 \pm 0.217$ & $0.000 \pm 0.000$ & $+0.709$ \\
zale & $0.567 \pm 0.026$ & $0.515 \pm 0.017$ & $+0.052$ \\
\midrule
\textbf{Sum} & \textbf{6.739} & \textbf{5.255} & \textbf{+1.484} \\
\bottomrule
\end{tabular}
\end{table}
%

% ------------------------------

\subsection{Additional \textit{in vitro} results}
\label{sec:appendix-in-vitro-results}

This section provides complete \textit{in vitro} validation results for the diverse portfolio of $M=20$ antimicrobial peptides produced by one run of \ourmethod{}. Experimental procedures for measuring \textit{in vitro} MIC values are detailed in \cref{sec:in-vitro}.

\cref{tab:optimized_peptides} lists the amino acid sequences of all 20 optimized peptides (P1–P20) along with their APEX 1.1–predicted MIC values. These peptides were optimized \textit{in silico} to minimize average predicted MIC across 11 target bacteria (B1–B11; species names provided in \cref{tab:bacteria}). A key motivation for optimizing a diverse portfolio rather than a single candidate is to mitigate downstream failure risk: \textit{in silico} predictions are imperfect, and peptides may fail to synthesize or exhibit unexpected behavior in solution. By generating 20 diverse candidates that all exhibit strong predicted inhibitory activity, we increase the probability that at least one will demonstrate true efficacy \textit{in vitro} against each target bacterium.

\begin{figure}[!ht]
  \vskip 0.2in
  \begin{center}
    \centerline{\includegraphics[width=\columnwidth]{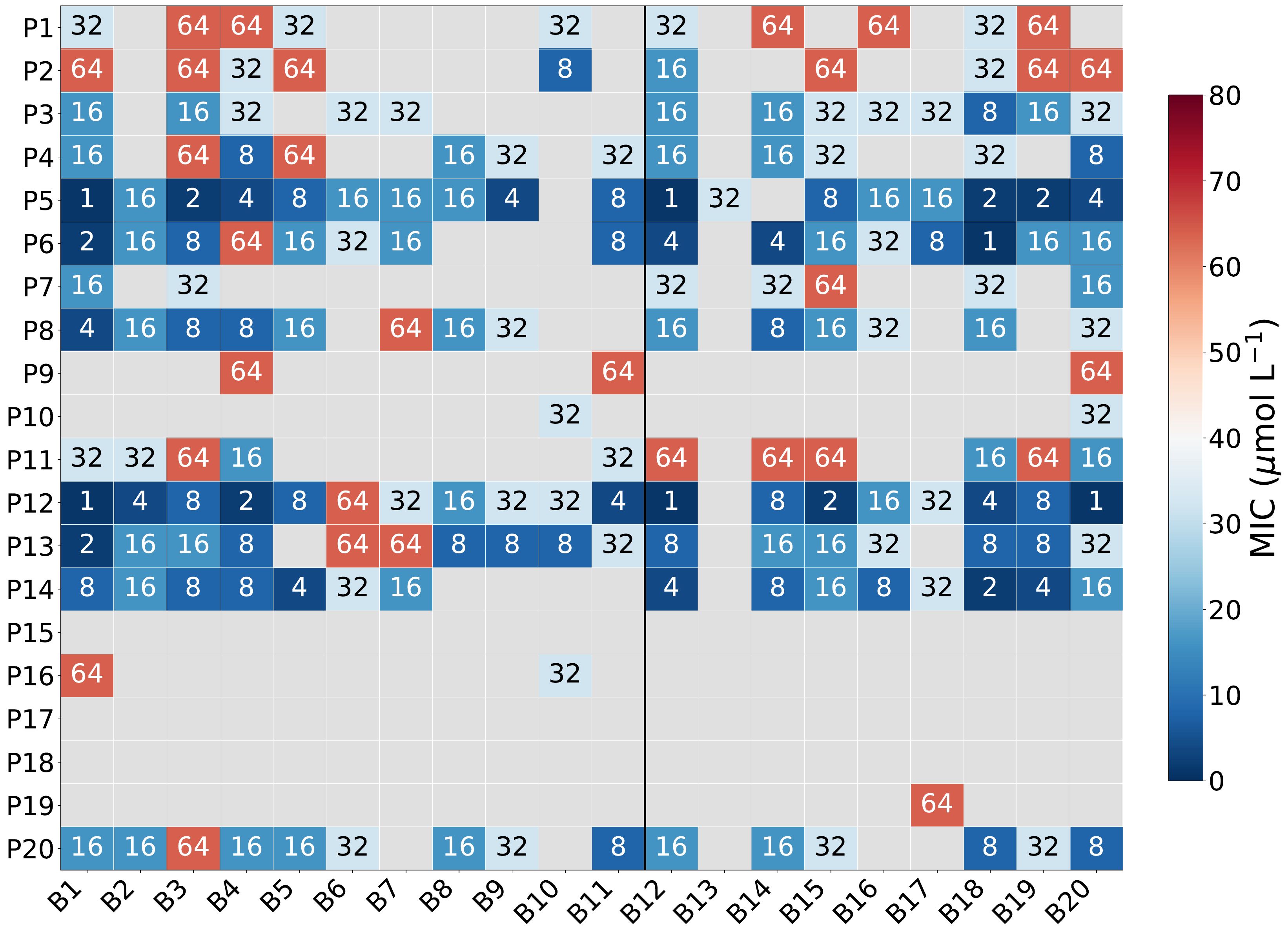}}
    \caption{
\textit{In vitro} MIC results for all 20 peptides (P1–P20) in the optimized diverse portfolio produced by one run of \ourmethod{}. Results are shown against the 11 optimization target bacteria (B1–B11; see \cref{tab:bacteria}) as well as 9 additional bacteria (B12–B20; see \cref{tab:extra_bacteria}) that were not part of the optimization objective. Peptide sequences for P1–P20 are listed in \cref{tab:optimized_peptides}. Gray cells indicate no detectable activity at the tested concentrations.
}
    \label{fig:heatmap_20_targets}
  \end{center}
\end{figure}

\cref{fig:heatmap_20_targets} presents measured \textit{in vitro} MIC values for all 20 peptides against 20 bacterial targets (B1–B20). These include the 11 bacteria used as optimization targets during \textit{in silico} design (B1–B11; see \cref{tab:bacteria}) as well as 9 additional bacteria (B12–B20; see \cref{tab:extra_bacteria}) that were not part of the optimization objective. The additional bacteria provide an evaluation of broader-spectrum antimicrobial activity beyond the original targets. Gray cells indicate no detectable inhibitory activity at the concentrations tested.

For the original 11 optimization targets (B1–B11), at least one peptide in the portfolio achieved very strong \textit{in vitro} activity (MIC $\leq 16$ $\si{\micro\mole\per\liter}$) against each bacterium, validating the diverse portfolio optimization strategy. Notably, several peptides also exhibited strong inhibitory activity against the held-out bacteria B12–B20, despite these targets not being included in the optimization objective. This demonstrates that optimizing for efficacy against a diverse panel of bacteria can yield peptides with broad-spectrum antimicrobial properties that generalize to previously unseen targets.

\begin{table}[!ht]
\centering
\caption{Example portfolio $M=20$ diverse peptides optimized by a single random run of \ourmethod{}. Specifically, this portfolio was the one selected for \textit{in vitro} validation. Peptide IDs P1-P20 used to identify the $M=20$ unique peptides in the portfolio in all \textit{in vitro} results provided in this paper (e.g., in \cref{fig:heatmap_20_targets} and \cref{fig:heatmap_condensed}). The ``Pred. MIC" column provides the APEX 1.1 Model's predicted MICs averaged across the 11 target bacteria listed in \cref{tab:bacteria} (the black-box objective value used during optimization with \ourmethod{}).
}
    \label{tab:optimized_peptides}
\resizebox{0.9\columnwidth}{!}{
    \begin{tabular}{lcc}
        \toprule
        Peptide ID & Sequence & Pred. MIC \\
        \midrule
        P1  & \texttt{FLRWKLRFRIRLIL} & 17.5 \\
        P2  & \texttt{KIRWRIRILFRLLLKKF} & 22.0 \\
        P3  & \texttt{WKRLFKRIKIVLRWF} & 23.7 \\
        P4  & \texttt{WRLIILRAARWLLK} & 24.7 \\
        P5  & \texttt{IALRRWIFKLAKALKW} & 25.2 \\
        P6  & \texttt{RIKFWKLRIIKFF} & 25.6 \\
        P7  & \texttt{LKMKIALLKLVAGKKL} & 26.2 \\
        P8  & \texttt{KIILKIRWRWLLNIAKLAAFK} & 26.6 \\
        P9  & \texttt{FKLWRRWWWLVLR} & 26.8 \\
        P10 & \texttt{WRWLAKIAIRAFWKLKIKW} & 26.9 \\
        P11 & \texttt{LIRFRFRLKWRLF} & 27.0 \\
        P12 & \texttt{KWIKLVRWFKWIKF} & 27.1 \\
        P13 & \texttt{IFRWLKRWVFRW} & 27.5 \\
        P14 & \texttt{KWKKKIFLKVRFW} & 27.7 \\
        P15 & \texttt{FWWKFIRWLRRILLRRFFRW} & 27.8 \\
        P16 & \texttt{LKKIWLRIIIKRFLRWKFRLLL} & 27.8 \\
        P17 & \texttt{LFWKKFRIRWRWWWLRIFLRWRWLLRWWWILLRFRFF} & 28.1 \\
        P18 & \texttt{IRWFKRLRFRLWWWFRFLRRVF} & 28.1 \\
        P19 & \texttt{ILKIRFRWKIRLFFKLLRKWLWWF} & 28.3 \\
        P20 & \texttt{KIFWRILILGRLLIKRFLKKKLLVKW} & 28.5 \\
        \bottomrule
    \end{tabular}
}
\end{table}

\newpage 
\textcolor{white}{x}
\newpage 
\textcolor{white}{x}

\subsection{Additional ablations}
\label{sec:appendix-ablations}
\begin{figure}[!ht]
  \vskip 0.2in
  \begin{center}
    \centerline{\includegraphics[width=0.8\columnwidth]{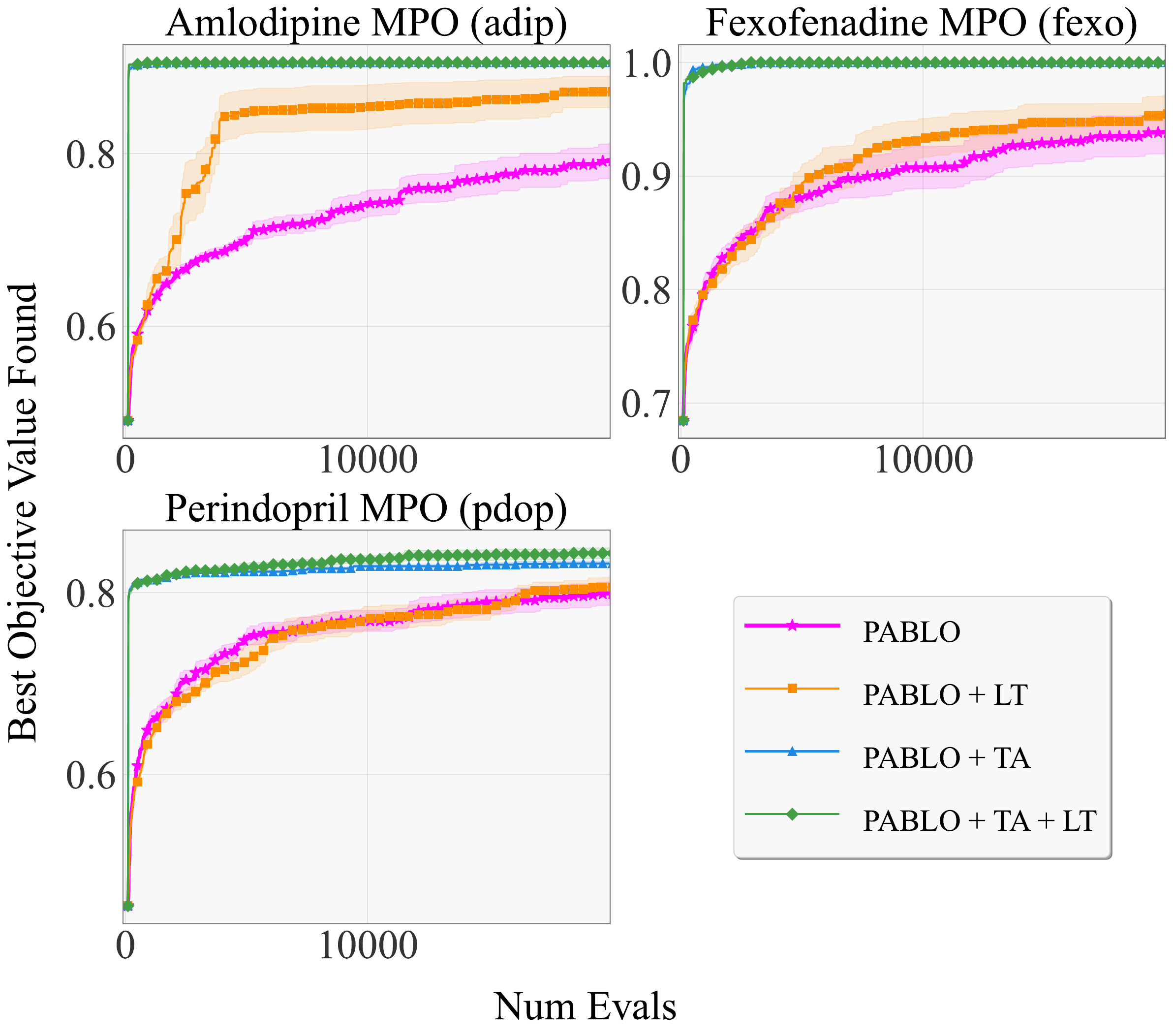}}
    \caption{
    Ablations on representative GuacaMol tasks showing the contribution of Task Awareness (TA) and the Literature RAG Tool (LT). See \cref{tab:ablation} for the same comparison across all selected 10 GuacaMol tasks. 
    }
    \label{fig:guacamol_ablation}
  \end{center}
\end{figure}

\begin{figure}[!ht]
  \vskip 0.2in
  \begin{center}
    \centerline{\includegraphics[width=0.8\columnwidth]{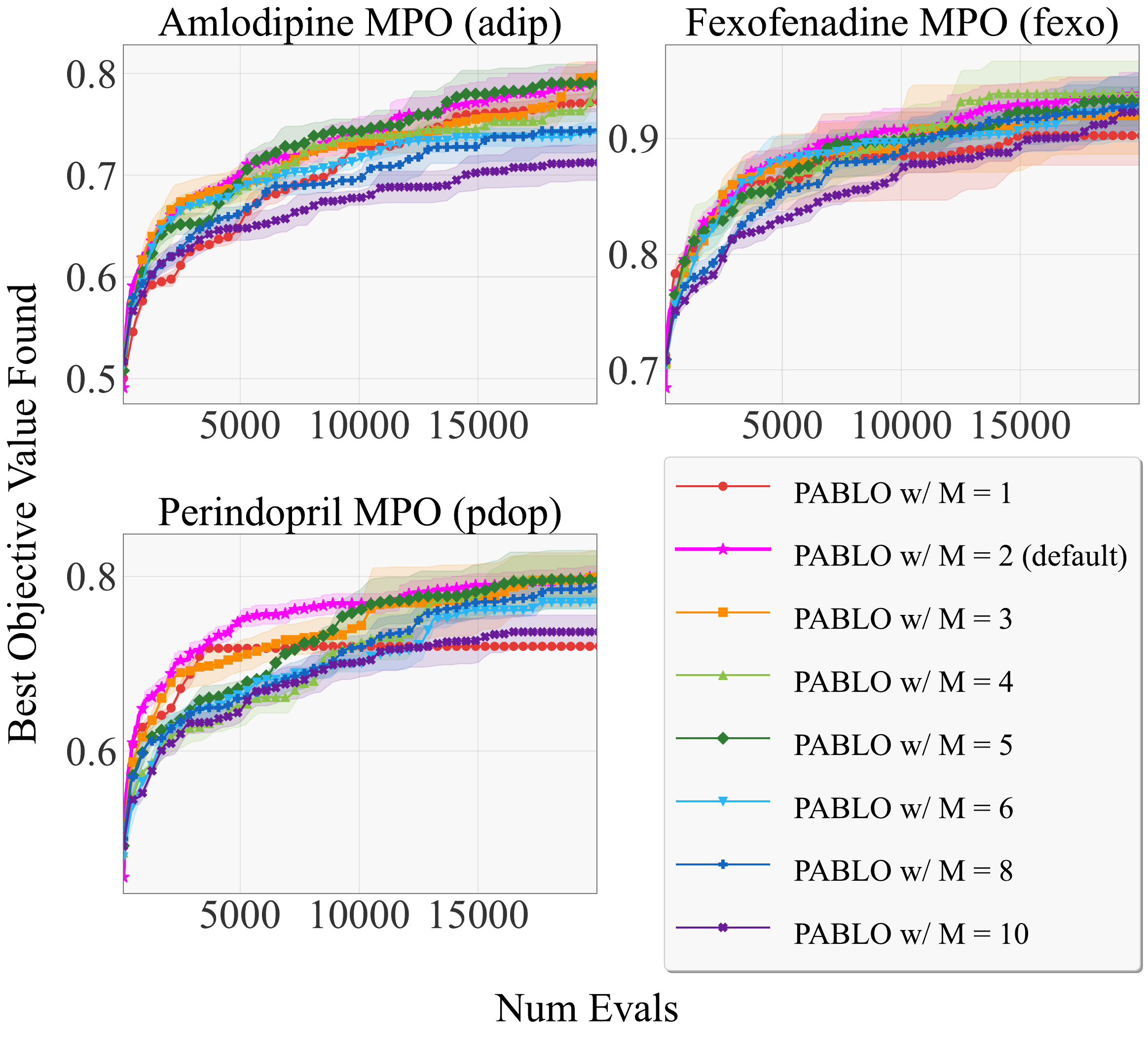}}
    \caption{
    Ablation of the number of diverse seeds ($M$) used for multi-trajectory local optimization in \ourmethod{} on three representative GuacaMol MPO tasks. We plot the objective value of the best molecule found so far versus the number of black-box evaluations.
    }
    \label{fig:n_div_seeds_ab}
  \end{center}
\end{figure}

\begin{figure}[!ht]
  \vskip 0.2in
  \begin{center}
  \centerline{\includegraphics[width=0.8\columnwidth]{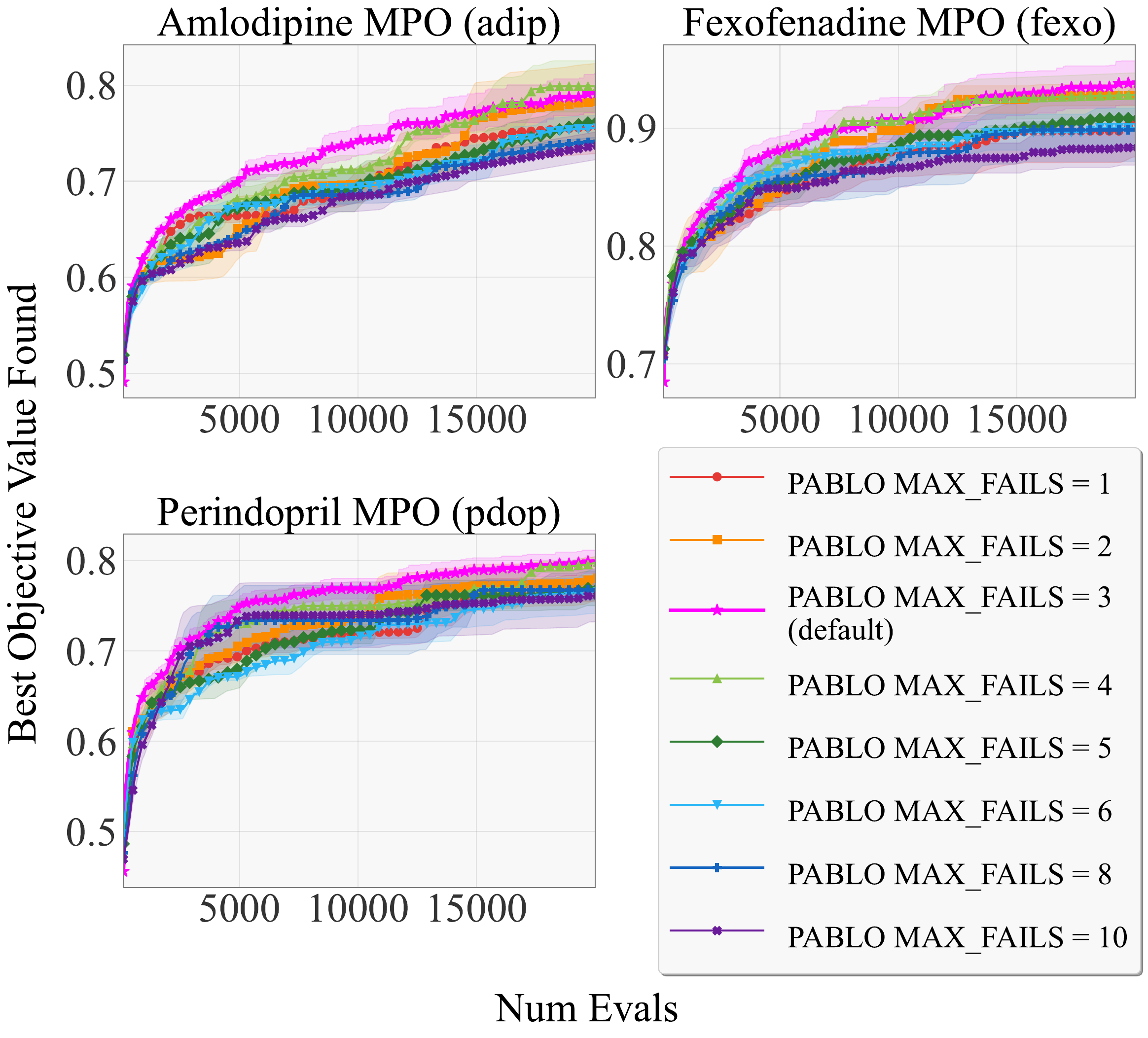}}
    \caption{
Ablation of the \texttt{MAX\_FAILS} hyperparameter in \ourmethod{} on three representative GuacaMol MPO tasks. We plot the objective value of the best molecule found so far versus the number of black-box evaluations.
}
    \label{fig:n_fails_ab}
  \end{center}
\end{figure}

In \cref{tab:ablation} and \cref{fig:guacamol_ablation}, we provide additional analysis of the effect of adding two optional enhancements 1) the Literature RAG Tool (LT) and 2) Task Awareness (TA) to \ourmethod{}, as described in \cref{sec:experiments}. \cref{tab:ablation} provides results on all 10 selected GuacaMol tasks while \cref{fig:guacamol_ablation} provides results on three representative GuacaMol tasks. As we saw in \cref{fig:guacamol_results}, adding the LT leads to significant performance improvements on some tasks. In \cref{fig:guacamol_ablation}, notice that adding TA leads yields the largest consistent gains across tasks in both convergence speed and final objective value. 
\begin{table}[!ht]
\centering
\caption{Ablation study comparing the effect of adding the Literature RAG Tool (LT) and Task Aware (TA) components to \ourmethod{} on all 10 selected GuacaMol tasks. As in \cref{tab:top1_results_1}, we report the mean and standard deviation of Top-1 molecules from 5 independent runs at 10K evaluations on our selected 10 tasks.}
\label{tab:ablation}
\resizebox{\textwidth}{!}{
\begin{tabular}{l|cccc}
\toprule
\textbf{Task} & \textbf{\ourmethod{}} & \textbf{\ourmethod{} + LT} & \textbf{\ourmethod{} + TA} & \textbf{\ourmethod{} + LT + TA} \\
\midrule
med1 & 0.402$\pm$0.007 & 0.421$\pm$0.029 & 0.446$\pm$0.031 & \textbf{0.447$\pm$0.036} \\
med2 & 0.343$\pm$0.010 & 0.364$\pm$0.035 & 0.416$\pm$0.008 & \textbf{0.456$\pm$0.025} \\
pdop & 0.769$\pm$0.036 & 0.772$\pm$0.041 & 0.829$\pm$0.006 & \textbf{0.836$\pm$0.014} \\
osmb & 0.922$\pm$0.012 & 0.919$\pm$0.018 & \textbf{1.000$\pm$0.000} & 0.999$\pm$0.004 \\
adip & 0.742$\pm$0.048 & 0.853$\pm$0.080 & \textbf{0.906$\pm$0.000} & \textbf{0.906$\pm$0.000} \\
siga & 0.798$\pm$0.051 & 0.791$\pm$0.029 & \textbf{0.804$\pm$0.051} & 0.786$\pm$0.056 \\
zale & 0.600$\pm$0.040 & 0.621$\pm$0.010 & 0.716$\pm$0.002 & \textbf{0.734$\pm$0.030} \\
valt & 0.957$\pm$0.037 & 0.960$\pm$0.068 & \textbf{0.989$\pm$0.006} & 0.988$\pm$0.006 \\
rano & 0.904$\pm$0.016 & 0.904$\pm$0.012 & 0.957$\pm$0.014 & \textbf{0.961$\pm$0.021} \\
fexo & 0.910$\pm$0.059 & 0.941$\pm$0.051 & \textbf{1.000$\pm$0.000} & \textbf{1.000$\pm$0.000} \\
\midrule
\textbf{Sum} & 7.347 & 7.546 & 8.063 & \textbf{8.113} \\
\bottomrule
\end{tabular}
}
\end{table}
In \cref{fig:n_div_seeds_ab}, we ablate the number of diverse seeds $M$ used to initialize the multi-trajectory local refinement stage of \ourmethod{}. We compare $M \in \{1,2,3,4,5,6,8,10\}$ on three representative GuacaMol MPO tasks. The setting $M=1$ corresponds to a simplified single-trajectory variant that always refines only the current best molecule. While this strategy can be competitive on some tasks (e.g., adip), it performs substantially worse on others (e.g., pdop and fexo), indicating that a single trajectory can become trapped in suboptimal local regions of chemical space.
Using multiple diverse seeds ($M>1$) mitigates this failure mode by enabling parallel refinement from distinct starting points, increasing the likelihood that at least one trajectory escapes poor local optima. Since it is not known \emph{a priori} which tasks will exhibit such local traps, we adopt $M=2$ as a robust default: it provides consistently strong performance across tasks while requiring minimal additional budget to maintain multiple trajectories. Performance is relatively stable for moderate values ($M=2$--$6$), suggesting \ourmethod{} is not overly sensitive to the precise choice of $M$. In contrast, very large values ($M=10$) can reduce performance by spreading the evaluation budget too thin across too many competing trajectories, reducing the depth of refinement per trajectory.
In \cref{fig:n_fails_ab}, we ablate the \texttt{MAX\_FAILS} hyperparameter controlling the persistence mechanism in \ourmethod{}. Recall that \texttt{MAX\_FAILS} specifies the number of consecutive unsuccessful attempts allowed before an agent stops persisting along its current strategy (global hypothesis or local refinement prompt) and moves on. We compare \texttt{MAX\_FAILS} $\in \{1,2,3,4,5,6,8,10\}$ on three representative GuacaMol MPO tasks.
Overall, we find that \ourmethod{} is robust to the choice of \texttt{MAX\_FAILS} within a reasonable range. Values in the range 2--4 achieve very similar performance across all tasks, with \texttt{MAX\_FAILS}=3 performing best overall, justifying our default choice. In contrast, setting \texttt{MAX\_FAILS}=1 tends to underperform, consistent with the intuition that allowing only a single attempt is overly reactive and can prematurely abandon promising strategies before they yield improvements. At the other extreme, very large values (e.g., \texttt{MAX\_FAILS}=10) degrade performance by over-committing evaluations to unproductive strategies, delaying necessary exploration or prompt adaptation. These results support the use of a small but non-trivial persistence window, where agents are given multiple opportunities to exploit a promising direction while still avoiding inefficient stagnation.

\newpage 
\textcolor{white}{x}
\newpage 
\textcolor{white}{x}
\newpage 
\section{Broader impacts}
\label{sec:impacts}
This work presents methods for computationally guided biological design, with a specific application to antimicrobial peptide discovery. The beneficial potential is significant: antimicrobial resistance is a growing global health crisis, and accelerating the discovery of novel antimicrobial agents could save lives. However, generative biological design tools also carry dual-use risks: the same methods that can design beneficial therapeutics could in principle be misused to engineer harmful biological agents. We believe responsible deployment of such systems requires institutional oversight and strict adherence to biosafety protocols.

\newpage 
\section{Limitations of \ourmethod{}}
\label{app:limits}

\ourmethod{} leverages LLMs as optimization agents, which introduces practical limitations related to (i) inference cost and (ii) agent latency/runtime.
Overall, \ourmethod{} trades increased inference overhead for substantially improved sample efficiency and solution quality, which is often the dominant priority in biological design settings where oracle evaluations often correspond to costly experiments.

\begin{figure}[t]
  \vskip 0.2in
  \begin{center}
    \centerline{\includegraphics[width=0.7\columnwidth]{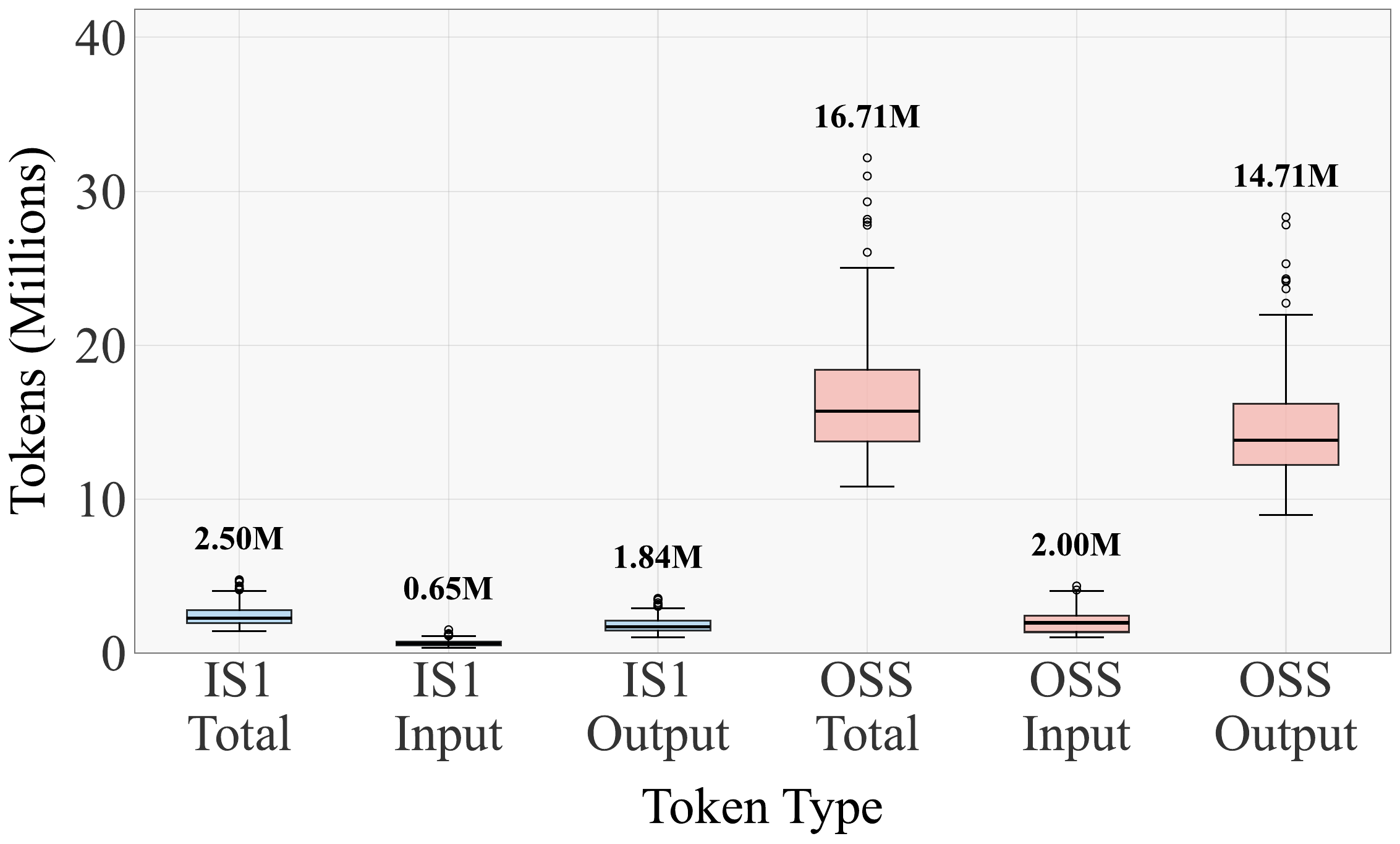}}
    \caption{
    Token usage distribution across optimization runs for the two LLMs used in \ourmethod{}: Intern S1 (IS1) and GPT-OSS-120B (OSS). We report input tokens, output tokens, and total tokens per run, where all runs use a budget of 20K total black-box function evaluations.
    }
    \label{fig:token_usage}
  \end{center}
\end{figure}

\paragraph{LLM inference cost.}
A primary limitation of \ourmethod{} is that it requires LLM queries throughout optimization. To quantify this overhead, we report token usage statistics in \cref{fig:token_usage} over $n=100$ full optimization runs of \ourmethod{}. As shown in the figure, each run consumes on average 2.50M tokens on Intern S1 (0.65M input; 1.84M output) and 16.71M tokens on GPT-OSS-120B (2.00M input; 14.71M output), for a total of 19.21M tokens per optimization run. \cref{fig:token_usage} also shows that most tokens are generated by the Worker Agent model (GPT-OSS-120B output tokens), reflecting \ourmethod{}'s deliberate design choice to allocate the bulk of generation to a fast, relatively inexpensive model.

To provide a concrete estimate of what this would cost under a hosted API deployment, we compute the \emph{equivalent inference cost per run} using public token pricing for GPT-OSS-120B (\$0.09/M input tokens and \$0.45/M output tokens). Under these rates, the mean GPT-OSS-120B cost per run is:
\[
\text{Cost}_{\text{OSS}}
= (2.00\times 0.09) + (14.71\times 0.45)
\approx \$6.80.
\]
Intern S1 pricing depends on the deployment/provider. Under a representative hosted inference rate of \$0.15/M input tokens and \$0.60/M output tokens, the mean Intern S1 cost is:
\[
\text{Cost}_{\text{IS1}}
= (0.65\times 0.15) + (1.84\times 0.60)
\approx \$1.20.
\]
This yields an estimated total hosted inference cost of:
\[
\text{Cost}_{\text{Total}} \approx \$8.00 \;\;\text{per optimization run}.
\]

Importantly, in our experiments we did \emph{not} incur per-token API costs, since we self-hosted both LLMs on our institutional GPU compute cluster. Intern S1 was served using \texttt{vLLM} on 4$\times$B200 GPUs, and GPT-OSS-120B was served using \texttt{SGLang} on 2$\times$B200 GPUs \citep{vllm, sglang}. We report the hosted-cost estimate above to make the computational overhead transparent and comparable for researchers deploying \ourmethod{} through public APIs.

While non-trivial, this inference cost is typically small relative to downstream experimental validation. In wet-lab design pipelines, synthesizing and assaying even a small number of candidates can cost orders of magnitude more than LLM inference. Moreover, \ourmethod{} is cost-aware by design: the majority of candidate generation is performed by the cheaper Worker model (GPT-OSS-120B), while the higher-capacity model (Intern S1) is used strategically for global reasoning (Explorer Agent) and prompt/task synthesis (Planner Agent), thereby limiting expensive calls.

\paragraph{Runtime and agent latency.}
A second limitation is wall-clock runtime, since optimization proceeds through sequential agent calls and oracle evaluations. In our experiments, runs of \ourmethod{} required an average of 99.7 hours to reach 20K total black-box evaluations. Although agent latency can be a bottleneck, this runtime is on par with (and often faster than) strong baselines that involve expensive training, model refitting, or repeated sampling, such as NF-BO, LLAMBO, and related deep generative optimization methods. In practice, \ourmethod{} benefits from the high throughput of GPT-OSS-120B, which accounts for most generation tokens, enabling competitive end-to-end runtime despite the hierarchical agentic loop.

Overall, \ourmethod{} trades increased inference overhead for substantially improved sample efficiency and solution quality, which is often the dominant priority in biological design settings where oracle evaluations correspond to costly experiments.

\paragraph{Runtime/cost trade-offs relative to baselines.}
To make this practical trade-off more explicit, \cref{tab:llm_runtime_cost} reports approximate hosted inference cost and wall-clock runtime per independent GuacaMol optimization run for \ourmethod{} and the main LLM-enhanced baselines. 
The two quantities capture different aspects of practical scalability. 
The estimated dollar cost reflects the marginal cost one would pay when using hosted LLM APIs, computed from token counts, each method's LLM, and public API pricing. 
By contrast, wall-clock runtime reflects the end-to-end time required by our experimental implementations, including sequential LLM calls, candidate generation, validation, and oracle evaluation. 
These estimates should therefore be interpreted as approximate and implementation-dependent rather than as fixed hardware-independent constants.

\begin{table}[t]
\centering
\caption{
Approximate per-run hosted inference cost and wall-clock runtime for LLM-enhanced GuacaMol optimization methods. 
Cost estimates use token counts, each method's LLM, and public API pricing. 
Wall-clock times are measured from our experimental runs and should be interpreted as approximate, since hardware, batching, model-serving infrastructure, and oracle implementation can substantially affect runtime.
}
\label{tab:llm_runtime_cost}
\begin{tabular}{lcc}
\toprule
\textbf{Method} & \textbf{\$/run} & \textbf{Hours/run} \\
\midrule
\ourmethod{} & \$8 & 100 \\
LLAMBO & \$157 & 100 \\
AlphaEvolve & \$0.19 & 1 \\
BOPRO-L & \$71 & 200 \\
BOPRO-S & \$2.36 & 50 \\
MOLLEO & \$20 & 50 \\
\bottomrule
\end{tabular}
\end{table}

Several conclusions follow from \cref{tab:llm_runtime_cost}. 
First, \ourmethod{} is not the lowest-latency method: a full 20K-evaluation run takes approximately 100 hours in our implementation, which is a real practical limitation. 
Second, the monetary cost of \ourmethod{} is moderate relative to other LLM-enhanced optimizers: its estimated hosted inference cost is substantially lower than LLAMBO and BOPRO-L, comparable to MOLLEO up to constant factors, and higher than AlphaEvolve and BOPRO-S. 
Third, AlphaEvolve is especially efficient in both cost and wall-clock time because it modifies a candidate-generating program rather than using LLM agents for repeated decision making. 
This is an important advantage of AlphaEvolve in settings where latency or very large numbers of repeated optimization runs are the primary concern, even though \ourmethod{} achieves stronger optimization performance on the biological design tasks considered here.

The 100-hour wall-clock runtime should also be interpreted in the context of the 20K-evaluation benchmark budget. 
In many GuacaMol tasks, \ourmethod{} reaches strong objective values much earlier, often within the first 1K--5K evaluations. 
Under our implementation, this corresponds roughly to 5--25 hours of wall-clock time. 
Thus, for practical deployments where the goal is to obtain a high-quality candidate rather than exhaust a fixed benchmark budget, early stopping can substantially reduce runtime. 
Moreover, in our experiments, the LLMs were self-hosted using \texttt{vLLM} or \texttt{SGLang} and shared across simultaneous optimization runs. 
Therefore, a single optimization run's wall-clock duration should not be interpreted as requiring a dedicated GPU allocation for the entire run; in a scaled deployment, LLM serving can be amortized across many concurrent runs.

Overall, the practical value of \ourmethod{} depends on the relative importance of sample efficiency, final solution quality, latency, and marginal inference cost. 
When oracle evaluations correspond to wet-lab synthesis, biological assays, expensive simulations, docking, or structure prediction, sample efficiency and final candidate quality may dominate wall-clock latency. 
When the objective is extremely cheap, many independent optimization runs are required, or rapid turnaround is the primary goal, lower-latency methods such as AlphaEvolve or lightweight classical optimizers may be preferable. 
Reducing \ourmethod{}'s latency through smaller specialized agent models, better batching, asynchronous Worker execution, adaptive early stopping, and improved LLM serving is an important direction for future work.

Additionally, we note that classical (non-LLM) ML methods for biological design often rely on computationally expensive deep generative models to generate candidate molecules or proteins. As a result, these approaches often incur substantial wall-clock runtimes comparable to LLM-based methods. For example, NF-BO~\citep{r2-cite9-lsbo-nfbo} requires approximately 50 hours on average on the same GuacaMol tasks.

\paragraph{Data contamination.}
Additionally, potential data contamination as an inherent caveat for any LLM-based optimization approach. However, we want to emphasize several points that argue against simple LLM memorization as the explanation for \ourmethod{}'s strong empirical performance. 
\ourmethod{} operates as a black-box optimizer that receives only SMILES strings and numeric scores. \ourmethod{} is not given target molecules, oracle definitions, or task names. In our primary GuacaMol results (\cref{fig:guacamol_results}, \cref{tab:top1_results_1}), we do not use task descriptions or task awareness and the \ourmethod{} agents see only structure-score pairs. Additionally, optimization trajectories show gradual improvement over thousands of evaluations rather than immediate recovery of near-optimal molecules. This is inconsistent with simple retrieval of memorized optima.

\subsection{Optimization perspective and theoretical limitations}
\label{sec:theory_discussion}

\ourmethod{} is primarily an empirical systems contribution. 
In this section, we provide optimization-level intuition for \ourmethod{} and discuss why formal convergence or sample-complexity analysis remains challenging for this class of structured biological design problems.

\paragraph{Challenges for formal analysis.}
A believable analysis is very hard for these kinds of structured optimization problems regardless of whether one is using agents, Bayesian optimization (BO), or any other framework. 
BO convergence results, for example, typically (1) assume the objective is a sample path from the chosen prior, or make roughly equivalent frequentist assumptions about bounded reproducing-kernel Hilbert space (RKHS) norm, and (2) result in extremely pessimistic regret bounds in the dimensionality of the objective function. 
The first assumption is highly suspicious in the structured domain in particular, because it would require believing, e.g., that the antimicrobial activity of a peptide is a draw from a Gaussian process whose prior covariance function is an RBF kernel on top of a massive normalizing-flow encoder. 
The second issue has certainly become more interesting in recent years as, e.g., local and smoothness-assuming approaches to BO achieve sharply more optimistic results than the bounds imply.

For studying LLMs formally as optimizers specifically, there are serious roadblocks. 
For example, it is known that simply swapping the order of instructions to the LLM changes the output token probabilities, even if the text is identical. 
This non-exchangeability of conditioning on information makes LLMs fundamentally incompatible with traditional analyses of posterior inference one might like to do, even before coming to the problem that conditioning the LLM densely on the input space---a fundamental requirement for eventual convergence in other optimization schemes, where uncertainty must eventually collapse everywhere---seems challenging with in-context learning alone due to context sizes. 
For these reasons, we do not claim formal convergence or sample-complexity guarantees for \ourmethod{}.

\paragraph{Optimization-level intuition.}
At a high level, \ourmethod{} can be understood as an online policy over a combinatorial space of natural-language search strategies, where the black-box objective provides a reward signal that drives strategy selection. 
The Planner maintains a task registry with empirical success rates, effectively implementing a multi-armed bandit over local search operators: strategies that consistently improve candidates are selected more often, while unproductive ones are pruned. 
The Explorer provides a complementary global diversification mechanism, analogous to the restart/exploration component in algorithms like random-restart hill climbing or CMA-ES population resampling. 
The key algorithmic insight is that expressing both local strategies and global hypotheses in natural language---rather than as fixed mutation operators or acquisition functions---allows the system to adapt its search policy online based on observed structure--activity patterns, which fixed-strategy optimizers cannot do.

That said, formalizing this intuition into convergence or sample-complexity guarantees remains an open problem. 
This is non-trivial for any optimizer over structured discrete spaces. 
Methods like BO have storied histories of intuition about when and how they work that agents lack. 
However, in many cases the ability of an agent to (1) condition on prior knowledge baked into the model parameters or a searchable database, and (2) condition on feedback written in natural language---as evidenced by the relatively consistent lift obtained by the use of a literature search tool---is a significant advantage even though formalizing this advantage theoretically remains an open problem.

\newpage 
\section{Additional implementation details}
\label{sec:app_imp_details}
In this section, we provide additional implementation details for \ourmethod{}. 

\subsection{\ourmethod{} hyperparameters}

All hyperparameters were fixed for all runs of \ourmethod{} across tasks unless otherwise noted.

\paragraph{Global context construction.}
The global context $C_{\text{global}}$ provided to the Explorer Agent has a fixed size of $20$ candidates in all experiments. We include the top-$k$=8 best-performing candidates as positive exemplars. The remaining $12$ candidates are sampled from the rest of the optimization history at approximately uniform rank intervals (with a random offset), to ensure that the context spans the observed performance spectrum.

\paragraph{Success persistence.}
The persistence loop in both global and local phases is controlled by a patience counter \texttt{MAX\_FAILS}. In all experiments in \cref{sec:experiments}, we use \texttt{MAX\_FAILS} $= 3$. An ablation study justifying this default is shown in \cref{fig:n_fails_ab}.

\paragraph{Number of diverse seeds.}
For multi-trajectory local optimization, we use $M = 2$ diverse seeds in all experiments. An ablation over $M$ is provided in \cref{fig:n_div_seeds_ab}, which shows that $M=2$ provides a strong trade-off between robustness to local optima and efficient allocation of evaluation budget.

\paragraph{Distance function and diversity threshold for $S_{\mathrm{seed}}$.}
As discussed in \cref{sec:method}, selection of $S_{\mathrm{seed}} = \{x^{(1)}, \ldots, x^{(M)}\}$ on each round of \ourmethod{} involves greedy selection under a diversity threshold, using a domain-specific distance function $\operatorname{dist}(\,\cdot\, , \,\cdot\,)$. For all molecule design tasks in this paper, we used fingerprint similarity (FPS) to measure distance between molecules (computed using RDKit) and a threshold of $0.5$. Thus, we require that ever pair of molecules in $S_{\mathrm{seed}}$ has FPS no higher than 0.5. 
For all peptide design tasks, we used normalized Levenshtein edit distance to measure distance between peptide amino acid sequences. In particular, normalized edit distance between two peptide sequences means we compute Levenshtein edit distance between the two sequences, and then divide by the length of the shorter sequence. For all peptide design runs we use a diversity threshold of 0.75. Thus we require that ever pair of peptides in $S_{\mathrm{seed}}$ has normalized edit distance of at least 0.75. 

\paragraph{Hyperparameters for diverse portfolio optimization.}
For runs of our method to optimize a diverse set of $M=20$ peptides (rather than just one peptide), we similarity require a means of measuring distance between peptides and a diversity threshold (the minimum distance we want to require between peptide sequences in our diverse portfolio). For this, we use the same distance function and diversity threshold as we do for constructing the diverse set of peptides $S_{\mathrm{seed}}$. In particular, we measure distance between two peptide sequences using normalized edit distance: (Levenshtein edit distance) / (the length of the shorter sequence), and we use the same diversity threshold of threshold 0.75. Thus, we require all pairs of peptides in the optimized diverse portfolio to have a normalized Levenshtein edit distance of at least 0.75.

\paragraph{Task Registry initialization and size.}
The Task Registry is initialized with $3$ domain-specific default tasks. These default tasks are hand-designed to illustrate simple, standard modification strategies for the given domain. The exact default tasks used for each domain (molecules and peptides) are listed in \cref{app:task_registry_init_ex}. 

The Task Registry has a fixed maximum capacity of \texttt{MAX\_TASKS=20} tasks. When adding a new task would exceed this limit, the worst-performing non-default task (based on empirical success rate) is pruned. We chose a capacity of $20$ to maintain a diverse library of local search strategies while avoiding excessive context length that could degrade Planner Agent reasoning performance.

\paragraph{Planner Agent task selection ($|\mathcal{P}_{\text{work}}|$).}
At each Planner Agent invocation, the prompt instructs the agent to return approximately $8$--$10$ task prompts, consisting of a mixture of existing and newly synthesized tasks. The exact number is determined by the agent at generation time.

\paragraph{Explorer Agent batch size ($|B_{\text{global}}|$).}
At each global search step, the Explorer Agent is prompted to generate $10$--$20$ new candidates. The precise number of outputs is determined by the agent.

\paragraph{Worker batch size ($|B_{\text{loc}}|$).}
During local refinement, each Worker Agent is prompted to generate $5$--$10$ candidate modifications per iteration. The exact number is determined by the agent at generation time.

\subsection{Initialization data}
In all experiments, \ourmethod{} was initialized with $100$ data points. 
For AMP design tasks, we initialize with the $10$ template sequences in \cref{tab:templates} plus $90$ additional sequences generated via random mutations to those templates. 
For molecular design tasks, following \citet{r2-cite9-lsbo-nfbo}, we initialize with the first $100$ SMILES from the GuacaMol dataset~\citep{GuacaMol}. 

All initialization points are counted toward the total evaluation budget of $20{,}000$ oracle calls and are included on the x-axis in all plots in \cref{sec:experiments}. For baseline methods, following \citet{eff_matters_benchmark}, we adopt the initialization strategy specified in each baseline’s original implementation when available; otherwise, we use the same $100$ initialization data points as for \ourmethod{} to ensure fair comparison.

\paragraph{Initialization under zero-signal regimes.}
All GuacaMol tasks have objective functions bounded in $[0,1]$. This can sometimes produce large regions of search space with identical minimum score (exactly $0.0$). In such cases, an initial dataset may contain no score variation, providing no signal about how candidate structure influences the objective. To avoid this degenerate regime, we ensure that the initialization set contains at least one molecule with non-zero score before beginning optimization. 

Concretely, if all $100$ initial molecules have score $0.0$, we continue sampling additional SMILES uniformly at random from the GuacaMol dataset and evaluating them until at least one molecule with score $>0$ is found. These additional initialization evaluations are still counted toward the total budget and are included on the x-axis of all plots.

Importantly, when there is even minimal score differentiation (e.g., scores of $0.0$ vs. $10^{-100}$), PABLO can leverage this signal effectively. The random sampling fallback is needed only for the rare edge-case when the initialization dataset contains no score variation (truly zero-variance initialization). In such settings, no optimizer has useful information to exploit, and random initialization is in fact standard practice across the black-box optimization and Bayesian optimization (BO) literature. For example, most standard BO methods begin with a random or space-filling initialization phase (e.g., LHS, Sobol) before switching to acquisition-guided search (e.g.,~\citep{garnett2023bayesian, balandat2020botorch}). \ourmethod{}'s fallback to random sampling in this regime is consistent with this standard practice.

In our experiments, this edge-case arose for only one objective: Valsartan SMARTS. In our runs, between approximately $200$ and $1400$ additional molecules were evaluated before observing a non-zero score. The highest score observed during this random initialization phase was extremely small ($4.87 \times 10^{-36}$), but even this minimal deviation from zero was sufficient to provide a usable optimization signal. With a single non-zero example, \ourmethod{} was then able to reliably discover molecules with near-optimal scores $>0.9$. In contrast, when initialized with only zero-scoring molecules, \ourmethod{} made no progress and remained at $0.0$ for the entire optimization run.

We therefore recommend this strategy for benchmarks with hard lower bounds and large zero-signal regions, where initial samples may all share the same minimum score. In such cases, allocating additional budget to random exploration until at least some score variation is observed can dramatically improve optimization performance.

This issue does not arise in our peptide experiments or in many realistic experimental or predictive oracles, where scores are continuous and do not exhibit large flat regions at an absolute minimum. In those settings, even small variations in initialization data provide sufficient signal for \ourmethod{} to begin effective optimization without any special initialization procedure.

\begin{figure}[t]
    \centering
    \begin{subfigure}[b]{0.45\textwidth}
        \centering
        \includegraphics[width=0.8\textwidth]{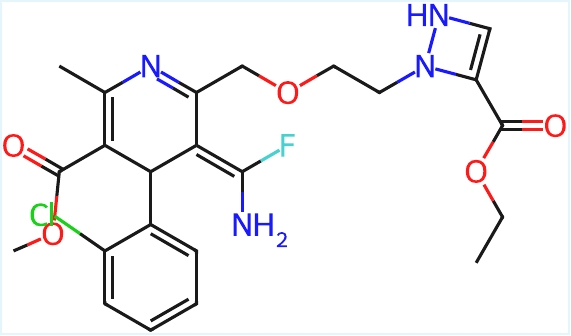}
        \caption{Query molecule}
        \label{fig:mol-rag-query}
    \end{subfigure}
    \hfill
    \begin{subfigure}[b]{0.45\textwidth}
        \centering
        \includegraphics[width=0.8\textwidth]{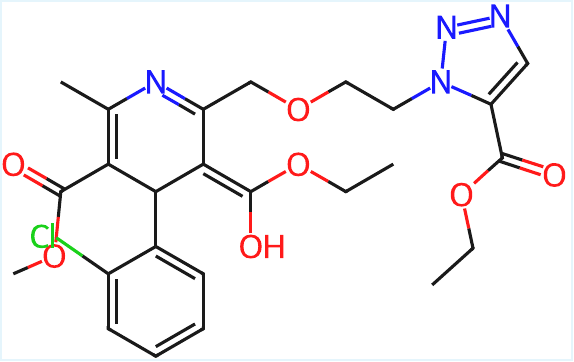}
        \caption{Retrieved: \textbf{9a} (PMID: 2342074)}
        \label{fig:mol-rag-retrieved}
    \end{subfigure}
    
    \begin{quote}
    \small\textit{``Thus, the 1,2,3-triazole derivatives with a 5-carbethoxy (9a) and 5-carboxamido (11a) substituent are some 5-fold less active as calcium antagonists than their corresponding 4-isomers (8a and 10a, respectively).''}
    \end{quote}
    
    \caption{Molecule RAG retrieval example. Given a query structure, we retrieve papers discussing structurally similar molecules from PubMed.}
    \label{fig:mol-rag}
\end{figure}
\newpage 
\section{Applying \ourmethod{} to new domains}
\label{sec:domains}

Experiments in \cref{sec:experiments} cover two distinct biological modalities: small molecules and antimicrobial peptides (AMPs). 
These two modalities differ substantially in representation, search space, and objective structure. 
Yet, transitioning between molecules and AMPs required changing only the validity filter and three default task hints; the core optimization loop was unchanged. 

Importantly, \ourmethod{} is designed to be general-purpose, such that applying \ourmethod{} to a new domain requires minimal effort. 
The core optimization loop---planner/explorer/worker decomposition, task registry dynamics, history compression, score-driven adaptation, etc.---remains unchanged across domains. 
Furthermore, core \ourmethod{} agents and prompts are purposefully defined in a domain-agnostic way so that adding new domains is as straightforward as possible. 

Applying \ourmethod{} to a new domain (e.g., DNA, RNA, larger proteins) requires updating two domain-specific components:
\begin{itemize}
    \item \textbf{Validity filter:} An inexpensive function checking that a proposed candidate is a valid object in the relevant search space. 
    For molecules, this is a check that a SMILES string represents a valid structure (via RDKit). 
    For peptides, it is a check that the sequence contains only allowed amino acid characters. 
    Applying this to DNA is as simple as checking that the sequence contains only A, C, G, and T.

    \item \textbf{Default task hints:} \ourmethod{} bootstraps the task registry with three default tasks that serve as examples to ground the Planner in what a reasonable ``modification'' looks like. 
    Importantly, these were themselves generated by prompting an LLM (e.g., ``suggest 3 simple ways to modify a molecule'')---no domain expertise or manual prompt engineering was required. 
    Adapting to a new domain is as simple as prompting an LLM with the new domain name (e.g., ``suggest 3 simple ways to modify a DNA sequence'').
\end{itemize}

Outside of these, all other prompts stay the same except for literal domain names (e.g., find-and-replace ``SMILES string'' with ``DNA sequence''). 
Please also see the provided \ourmethod{} GitHub codebase, which includes a step-by-step guide in the README for running \ourmethod{} on any new domain.

\newpage 
\section{Compute resources}
\label{sec:compute}
We report the compute resources used to produce the empirical results in this paper. 
Experiments were run on a combination of our internal Locust compute cluster and the University of Pennsylvania PARCC compute cluster. 
The Locust cluster consists of 20 nodes, each with 2 Intel Xeon Silver 4310 CPU sockets, 250 GiB RAM, and one NVIDIA RTX A5000 GPU with 24 GiB GPU memory. 
On PARCC, experiments were run on the Betty high-performance computing and artificial intelligence cluster. 
The Betty AI system includes DGX B200 nodes with 8 NVIDIA Blackwell B200 GPUs per node and 180 GB HBM3e memory per GPU. 
Betty also includes CPU nodes with AMD EPYC 9374F processors, including standard-memory nodes with 64 CPU cores and 384 GB RAM per node, and large-memory CPU nodes with 64 CPU cores and 1152 GB RAM per node.

\paragraph{Compute specifications.}
For \ourmethod{}, GPU workers were used to host the LLMs used by the Explorer, Planner, and Worker agents, while CPU workers were used for experiment orchestration, validation, deduplication, objective evaluation, and logging. 
The LLM servers were shared across experiments: a single hosted Intern-S1 or GPT-OSS-120B server could serve requests from multiple simultaneous \ourmethod{} runs. 
Thus, the wall-clock time of an individual \ourmethod{} run should not be interpreted as requiring a dedicated GPU allocation for the full duration of the run. 
For baselines with learned generative models, surrogate models, or neural acquisition components, we used GPU workers when required by the original method implementation; otherwise, experiments were run on CPU workers. 
GPU memory usage for \ourmethod{} was dominated by LLM serving, while CPU memory usage was primarily used for storing optimization histories, candidate pools, parsed model outputs, and oracle-evaluation results.

\paragraph{Execution time and per-run cost.}
Each independent \ourmethod{} run on a GuacaMol task required approximately 100 hours of wall-clock time. 
Runs of \ourmethod{} with the Literature Tool (PABLO + LT) required similar wall-clock time, approximately 105 hours per run. 
AMP optimization runs used the same \ourmethod{} implementation and required approximately 100 hours per independent run. 
Because LLM serving was shared across concurrent runs, the marginal compute cost of a \ourmethod{} run is better reflected by token usage than by assigning a full GPU worker to each run. 
As reported in \cref{fig:token_use_bar}, \ourmethod{} uses approximately 19.1M LLM tokens per GuacaMol run on average. 
Using public API pricing for the corresponding LLMs, this corresponds to approximately \$8 per \ourmethod{} run, although our experiments used internally hosted models rather than paid API calls.

For LLM-enhanced GuacaMol baselines, approximate wall-clock time and estimated public-API token cost per independent run were: 100 hours and \$157 for LLAMBO, 1 hour and \$0.19 for AlphaEvolve, 200 hours and \$71 for BOPRO-L, 50 hours and \$2.36 for BOPRO-S, and 50 hours and \$20 for MOLLEO. 
CPU-only or lightweight baselines, including random search and some classical optimization baselines, were substantially less expensive, while GPU-based Bayesian-optimization, reinforcement-learning, and generative-model baselines varied by implementation but typically required tens of hours per independent run.

\paragraph{Total compute.}
We estimate that the reported experiments required approximately 30{,}000--50{,}000 GPU-worker hours and 20{,}000 CPU-worker hours in total. 
This estimate includes LLM serving for the main GuacaMol optimization experiments, PABLO + LT extension experiments, LLM-enhanced baseline runs, AMP optimization experiments, ablations, and appendix studies. 
The GPU-worker-hour estimate counts time spent keeping LLM servers and GPU-based baseline jobs active; because LLM servers were shared across multiple concurrent optimization runs, it is lower than the sum obtained by multiplying each \ourmethod{} run's wall-clock duration by one dedicated GPU. 
For example, the main GuacaMol experiments include 10 tasks with 10 independent runs per method for most methods, with BOPRO-L run on only 3 of the 10 tasks due to its higher cost.

\paragraph{Compute resources for preliminary experiments.}
The full research project required additional compute beyond the final reported experiments for method development, hyperparameter selection, debugging, failed runs, and preliminary ablations that are not included in the paper. 
These preliminary experiments were run on the same Locust and PARCC clusters described above. 
We estimate that preliminary experiments required approximately 10{,}000--20{,}000 additional GPU-worker hours and 5{,}000 additional CPU-worker hours.
\newpage
\section{Antimicrobial peptide (AMP) optimization task details}
\label{sec:amp-task-more-detials}
\begin{table}[!ht]
\centering
\caption{Names of the $11$ target bacteria used for the AMP design task.}
    \label{tab:bacteria}
\resizebox{\columnwidth/2}{!}{
    \begin{tabular}{lc}
        \toprule
        Objective ID & Target Pathogenic Bacteria \\
        \midrule
        B1  & \texttt{A. baumannii ATCC 19606} \\
        B2 & \texttt{E. coli ATCC 11775} \\
        B3 & \texttt{E. coli AIC221} \\
        B4 & \texttt{E. coli AIC222-CRE} \\
        B5 & \texttt{K. pneumoniae ATCC 13883} \\
        B6 & \texttt{P. aeruginosa PAO1} \\
        B7 & \texttt{P. aeruginosa PA14} \\
        B8 & \texttt{S. aureus ATCC 12600} \\
        B9 & \texttt{S. aureus ATCC BAA-1556-MRSA} \\
        B10 & \texttt{E. faecalis ATCC 700802-VRE} \\
        B11 & \texttt{E. faecium ATCC 700221-VRE} \\
        \bottomrule
    \end{tabular}
}
\end{table}
\begin{table}[!ht]
\centering
\caption{Names of the $9$ extra (non-target) bacteria use for \textit{in vitro} experiments testing broad spectrum activity of optimized peptides and included in \cref{fig:heatmap_20_targets}.}
    \label{tab:extra_bacteria}
\resizebox{\columnwidth/2}{!}{
    \begin{tabular}{lc}
        \toprule
        Objective ID & Pathogenic Bacteria \\
        \midrule
        B12 & \texttt{A. baumannii ATCC BAA-1605-CGTPACCIMRA} \\
        B13 & \texttt{E. cloacae ATCC 13047} \\
        B14 & \texttt{E. coli ATCC BAA-3170-CRE} \\
        B15 & \texttt{E. coli K-12 BW25113} \\
        B16 & \texttt{K. pneumoniae ATCC BAA-2342-EIRK} \\
        B17 & \texttt{P. aeruginosa ATCC BAA-3197-FBCRP} \\
        B18 & \texttt{S. enterica ATCC 9150} \\
        B19 & \texttt{S. enterica Typhimurtium ATCC 700720} \\
        B20 & \texttt{B. subtilis ATCC 23857} \\
        \bottomrule
    \end{tabular}
}
\end{table}
\begin{table}[!ht]
\centering
\caption{Template amino acid sequences used for the ``template constrained" AMP design task.}
\label{tab:templates}
\resizebox{\columnwidth/3}{!}{
    \begin{tabular}{c} 
        \toprule
        Template Amino Acid Sequences \\
        \midrule
        \texttt{RACLHARSIARLHKRWRPVHQGLGLK} \\
        \texttt{KTLKIIRLLF} \\
        \texttt{KRKRGLKLATALSLNNKF} \\
        \texttt{KIYKKLSTPPFTLNIRTLPKVKFPK} \\
        \texttt{RMARNLVRYVQGLKKKKVI} \\
        \texttt{RNLVRYVQGLKKKKVIVIPVGIGPHANIK} \\
        \texttt{CVLLFSQLPAVKARGTKHRIKWNRK} \\
        \texttt{GHLLIHLIGKATLAL} \\
        \texttt{RQKNHGIHFRVLAKALR} \\
        \texttt{HWITINTIKLSISLKI} \\
        \bottomrule
    \end{tabular}
}
\end{table}
\cref{tab:bacteria} specifies the $11$ target bacteria used for the AMP design task from \cref{sec:experiments}. Specifically, we designed peptides to optimize the mean predicted MIC across the $11$ target bacteria in \cref{tab:bacteria}, where in-silico MICs were predicted using the Apex 1.1 model as an oracle \citep{apex1}. The first seven bacteria are Gram negative bacteria (Objective IDs B1-B7) and the last four (Objective IDs B8-B11) are Gram positive.

\cref{tab:templates} gives the $10$ template amino acid sequences used for the ``template constrained" variation of the AMP design task from \cref{sec:experiments}.
\newpage 
\section{\textit{In vitro} minimal inhibitory concentration (MIC) experimental methods}
\label{sec:in-vitro}

\paragraph{Peptide synthesis and characterization.}
Peptides were synthesized on an automated peptide synthesizer (Symphony X, Gyros Protein Technologies) by standard 9-fluorenylmethyloxycarbonyl (Fmoc)-based solid-phase peptide synthesis (SPPS) on Fmoc-protected amino acid-Wang resins (100–200 mesh). In addition to preloaded resins, standard Fmoc-protected amino acids were employed for chain elongation. N,N-Dimethylformamide (DMF) was used as the primary solvent throughout synthesis. Stock solutions included: 500 $\si{\milli\mole\per\liter}$ Fmoc-protected amino acids in DMF, a coupling mixture of HBTU (450 $\si{\milli\mole\per\liter}$) and N-methylmorpholine (NMM, 900 $\si{\milli\mole\per\liter}$) in DMF, and 20\% (v/v) piperidine in DMF for Fmoc deprotection. After synthesis, peptides were deprotected and cleaved from the resin using a cleavage cocktail of trifluoroacetic (TFA)/triisopropylsilane (TIS)/dithiothreitol (DTT)/water (92.8\% v/v, 1.1\% v/v, 0.9\% w/v, 4.8\%, w/w) for 2.5 hours with stirring at room temperature. 
The resin was removed by vacuum filtration, and the peptide-containing solution was collected. Crude peptides were precipitated with cold diethyl ether and incubated for 20 min at $\SI{-20}{\degreeCelsius}$, pelleted by centrifugation, and washed once more with cold diethyl ether. The resulting pellets were dissolved in 0.1\% (v/v) aqueous formic acid and incubated overnight at $\SI{-20}{\degreeCelsius}$, followed by lyophilization to obtain dried peptides. For characterization, peptides were dried, reconstituted in 0.1\% formic acid, and quantified spectrophotometrically. Peptide separations were performed on a Waters XBridge $C_{18}$ column ($4.6 \times \SI{50}{\milli\meter}$, $\SI{3.5}{\micro\meter}$, $\SI{120}{\angstrom}$) at room temperature using a conventional high-performance liquid chromatography (HPLC) system. Mobile phases were water with 0.1\% formic acid (solvent A) and acetonitrile with 0.1\% formic acid (solvent B). A linear gradient of 1–95\% B over 7 min was applied at $1.5 \si{\milli\liter\per\minute}$. UV detection was monitored at $220 \si{\nano\meter}$. Eluates were analyzed on Waters SQ Detector 2 with electrospray ionization in positive mode. Full scan spectra were collected over m/z 100–2,000. Selected Ion Recording (SIR) was used for targeted peptides. Source conditions were capillary voltage $3.0 \si{\kilo\volt}$, cone voltage 25-40 $\si{\volt}$, source temperature $\SI{120}{\degreeCelsius}$, and desolvation temperature $\SI{350}{\degreeCelsius}$. Mass spectra were processed with MassLynx software. Observed peptide masses were compared with theoretical values, and quantitative analysis was based on integrated SIR peak areas.

\paragraph{Bacterial strains and growth conditions.}
The bacterial panel utilized in this study consisted of the following pathogenic strains: 
\textit{Acinetobacter baumannii} ATCC 19606; 
\textit{A. baumannii} ATCC BAA-1605 (resistant to ceftazidime, gentamicin, ticarcillin, piperacillin, aztreonam, cefepime, ciprofloxacin, imipenem, and meropenem); 
\textit{Escherichia coli} ATCC 11775; 
\textit{E. coli} AIC221 [MG1655 phnE\textunderscore2::FRT, polymyxin-sensitive control]; 
\textit{E. coli} AIC222 [MG1655 pmrA53 phnE\textunderscore2::FRT, polymyxin-resistant]; 
\textit{E. coli} ATCC BAA-3170 (resistant to colistin and polymyxin B); 
\textit{E. coli} K-12 BW25113; 
\textit{Enterobacter cloacae} ATCC 13047; 
\textit{Klebsiella pneumoniae} ATCC 13883; 
\textit{K. pneumoniae} ATCC BAA-2342 (resistant to ertapenem and imipenem); \textit{Pseudomonas aeruginosa} PAO1; 
\textit{P. aeruginosa} PA14; 
\textit{P. aeruginosa} ATCC BAA-3197 (resistant to fluoroquinolones, $\beta$-lactams, and carbapenems); 
\textit{Salmonella enterica} ATCC 9150; 
\textit{S. enterica} subsp. \textit{enterica} Typhimurium ATCC 700720; 
\textit{Bacillus subtilis} ATCC 23857; \textit{Staphylococcus aureus} ATCC 12600; 
\textit{S. aureus} ATCC BAA-1556 (methicillin-resistant); 
\textit{Enterococcus faecalis} ATCC 700802 (vancomycin-resistant); and \textit{Enterococcus faecium} ATCC 700221 (vancomycin-resistant). 
\textit{P. aeruginosa} strains were propagated on Pseudomonas Isolation Agar, whereas all other species were maintained on Luria-Bertani (LB) agar and broth. For each assay, cultures were initiated from single colonies, incubated overnight at $\SI{37}{\degreeCelsius}$, and subsequently diluted 1:100 into fresh medium to obtain cells in mid-logarithmic phase.

\paragraph{Minimal inhibitory concentration (MIC) determination.}
MIC values were established using the standard broth microdilution method in untreated 96-well plates. 
Test peptides were dissolved in sterile water and prepared as twofold serial dilutions ranging from 1 to 64 $\si{\micro\mole\per\liter}$. Each dilution was combined at a 1:1 ratio with LB broth containing $4 \times 106$ CFU $\si{\per\milli\liter}$ of the target bacterial strain. Plates were incubated at $\SI{37}{\degreeCelsius}$ for 24 hours, and the MIC was the lowest peptide concentration that inhibited visible bacterial growth. All experiments were conducted independently in triplicate.

\newpage
\section{Example global context ($C_{\text{global}}$)}
\label{app:c_global_examples}

The global context $C_{\text{global}}$ provided to the Explorer Agent and Planner Agent consists of scored candidates drawn from the optimization history using the coverage sampling procedure described in \cref{sec:coverage_sampling}. The exact contents of $C_{\text{global}}$ change throughout optimization as new data are collected. Below we show one representative snapshot of $C_{\text{global}}$ from each domain.

\textbf{$C_{\text{global}}$ Snapshot for Molecules (score, SMILES):}

\begin{tcolorbox}[promptbox]
0.4833: CC(C)=CCn1c2cc(=O)ccc-2nc2c(C(N)=O)cccc21

0.4695: CC1(C)Oc2ccc(C\#N)cc2C(NC(=O)c2ccccn2)C1O

0.4447: CCOC(=O)C1C(=O)NC(c2ccccc2)=NC1c1cccc(Br)c1

0.4427: CN1CC(N(Cc2ccc(F)cc2F)c2ccc(C\#N)c(Cl)c2)CC1=O

0.4426: CN1C(=O)C(c2ccc(C\#Cc3ccnc(Cl)c3)cn2)CC1(C)C

0.4389: CC(O)(COc1ccccc1)C(=O)N1CCc2c(C\#N)cccc21

0.4294: Cc1ccc(-n2ncc3c(=O)n(-c4cccc(C)c4)c(=O)[nH]c32)cc1

0.4171: CC(=O)N1CCCC12c1ccccc1-c1nc(O)c3nccn3c12

0.2909: CC1Cn2c(nnc2-c2ccccn2)CN1C(=O)c1cccc(Cl)c1Cl

0.1400: O=c1c(C2=Nc3ccccc3S(=O(=O)N2)c(O)c2cc(F)ccc2n1CCC1CC1

0.07337: O=C1OC(CO)(COC(=O)c2ccccc2)CC1=Cc1ccc([N+](=O)[O-])cc1

0.03613: COc1ccc(S(=O)(=O)NC(C)C(=O)Nc2ccc3c(c2)OCO3)cc1OC

0.01600: O=c1ccncn1CC1(O)CC2NCCCC2O1

0.005560: CC1CC2CCC3OC(=O)C(C1SCCS(=O)(=O)O)C23O

0.001304: Nc1ncnc2c1c(-c1cc3ccccc3c3ccccc13)cn2C1OC(CO)C(O)C1O

1.957e-04: CC(=CC(=O)O)C(O)P(=O)(O)CCC(N)C(=O)O

1.044e-05: COc1ccc2c(c1)N=C(N)NC2

1.205e-07: COc1cccc2c1CCC1CN(CCn3c(=O)[nH]c4c(OC)c(OC)ccc4c3=O)CC21

1.657e-11: NC(=O)c1cc(CI)[nH]n1

3.136e-24: CSCCC(NC(=O)OC(C)(C)C)C(=O)N1CCC(C(=O)NC(C)C(=O)NC(C)c2ccccc2)CC1
\end{tcolorbox}

\textbf{$C_{\text{global}}$ Snapshot for Peptides (MIC, sequence):}

\begin{tcolorbox}[promptbox]
85.12:  KLLKIRRLWF

90.12:  WRKRGLKLATWLSLLNKF

95.11:  KWLKIIRLLF

99.25:  KHLKIMRLLW

106.4:  LGLKIIRLLF

109.4:  MGTKPMIKVRRKRLKQFVAK

111.3:  LGIRVRWMKTYTYCKKIKMFK

114.5:  KRVRINRHKRVLKKPVNDFPYLQF

270.0:  KNLKIIRLLA

313.2:  KHYKKLSTPPFTLNIRTLPKVKFPK

345.6:  KRKRGLKLGTQLSLNNKF

373.6:  RACLHARSIWRLHLRWRPVHQGLKLK

400.3:  RQKNHGIHFRVKANALR

432.5:  RQKNHGIRDRVLAKALR

466.2:  YAFWTFVPHPVVRFINRIP

486.6:  ANHLFTFAFPPCKILQNGRNQH

497.7:  QNQQKGATPGEFYKQFIQHC

503.5:  NSLYALGEYQLTKRY

507.3:  DNCQQVYYAWSHHHPMSSDLPA

510.3:  ESGNCQFDNEFEEHN

\end{tcolorbox}

\section{Example Explorer Agent prompts}
\label{app:explorer_prompt_ex}

The Explorer Agent is prompted using (i) a snapshot of the current global context $C_{\text{global}}$ (inserted into the prompt as a score-sorted list of previously evaluated candidates) and (ii) the current best observed score, which is used to set an explicit improvement target. Below we provide representative prompts used in each domain. See \cref{app:c_global_examples} for examples of the $C_{\text{global}}$ that is included in each full example prompt below.

\textbf{Explorer Example Prompt for Molecules:}

\begin{tcolorbox}[promptbox]
You are an expert molecule designer.
Generate molecules that you reason are most likely to score HIGHER than any in the provided MOLECULE-SCORE DATA.

\#\# MOLECULE-SCORE DATA (sorted high to low)

\{C\textunderscore global\}

\#\# TASK

Think step-by-step:

1. **Analyze the MOLECULE-SCORE DATA:** What molecular features correlate with high scores? Form 2-3 hypotheses about what the scoring function rewards.

2. **Generate:** Propose 10-20 NEW molecules that:

   - Push your hypotheses to their LOGICAL EXTREME for maximum scores
   
   - Combine best features from multiple top scorers
   
   - Explore creative new structural ideas

\#\# OUTPUT FORMAT

Return ONLY a JSON object with a list of VALID SMILES strings called 'candidates'.

Example:

\{
    "candidates": ["SMILES\textunderscore1", "SMILES\textunderscore2", "SMILES\textunderscore3", ...]
\}

\#\# GOAL

Propose new SMILES strings that could BEAT the current best score (\{best\textunderscore score\})
\end{tcolorbox}

\textbf{Explorer Example Prompt for Peptides:}

\begin{tcolorbox}[promptbox]
You are an expert antimicrobial peptide (AMP) designer. Generate peptides that you reason are most likely to achieve LOWER MIC than any in the provided data. 

The goal is to MINIMIZE the Minimum Inhibitory Concentration (MIC) averaged across 11 bacterial pathogens (including E. coli, P. aeruginosa, S. aureus/MRSA, K. pneumoniae, A. baumannii, and vancomycin-resistant Enterococci).

LOWER MIC = BETTER (less peptide needed to inhibit bacterial growth).

\#\# PEPTIDE-MIC DATA (sorted best to worst, LOWEST MIC first)

\{C\textunderscore global\}

\#\# TASK

Think step-by-step:

1. **Analyze the PEPTIDE-MIC DATA:** What sequence features correlate with low MIC? Form 2-3 hypotheses about what makes an effective AMP (e.g., cationic charge, amphipathicity, hydrophobic content, length, specific motifs).

2. **Generate:** Propose 10-20 NEW antimicrobial peptides that:

   - Push your hypotheses to their LOGICAL EXTREME for minimum MIC
   
   - Combine best features from multiple top performers
   
   - Explore creative new sequence ideas using what you know about AMPs

\#\# OUTPUT FORMAT

Return ONLY a JSON object with a list of VALID peptide sequences called 'candidates'.

Example:

\{
    "candidates": ["RKKLWLLRK", "FLPLIGKLLK", "KWKLFKKIGAVLKVL", ...]
\}

\#\# GOAL

Propose new antimicrobial peptide sequences that could BEAT the current best MIC (\{best\textunderscore score\})

\end{tcolorbox}

\section{Example Planner Agent prompts}
\label{app:planner_prompt_ex}

The Planner Agent is responsible for generating and selecting local-search task prompts to populate the Task Registry. Each Planner Agent call is provided with: (i) a snapshot of the current global context $C_{\text{global}}$, (ii) summary performance statistics for each task in the registry, and (iii) a compact summary of the currently-available tasks.

\subsection{Task Registry initialization}
\label{app:task_registry_init_ex}

We initialize the Task Registry with three domain-specific ``default'' tasks. These tasks serve two purposes:  
(1) they represent reasonable, simple modification strategies for the given domain, and  
(2) they provide examples that help the Planner Agent learn the structure and style of effective task prompts.

Below we list the three default tasks used for each domain.

\textbf{Molecules - Default Task 1 (SIMILAR):}

\begin{tcolorbox}[promptbox]
TASK: Generate SMILES that are structurally similar to the input molecule.
HINTS:

1. Modify side chains, linkers, or substituents.

2. Keep the core scaffold mostly intact.
\end{tcolorbox}

\textbf{Molecules - Default Task 2 (EXPLORE):}

\begin{tcolorbox}[promptbox]
TASK: Generate SMILES with different meaningful structural changes to the
input.

HINTS:

1. Each output should be a distinct modification type (ring size, linker swap, substituent move).

2. Make significant moves, not minor tweaks.

3. Explore broadly around the input.
\end{tcolorbox}

\textbf{Molecules - Default Task 3: (SCAFFOLD\_HOP):}

\begin{tcolorbox}[promptbox]
TASK: Generate scaffold hopping variations of the input molecule.

HINTS:

1. Make large topology-level changes 
(new ring systems, fusion patterns).

2. Avoid small local edits; make substantial core changes.

3. Try: fused<->bridged<->spiro,
cyclic<->polycyclic, aromatic<->non-aromatic cores.
\end{tcolorbox}

\textbf{Peptides - Default Task 1 (SIMILAR):}

\begin{tcolorbox}[promptbox]
TASK: Generate peptides that are conservative variants of the input.

HINTS:

1. Use similar amino acid substitutions 
(L<->I, D<->E, K<->R, F<->Y).

2. Preserve overall charge and hydrophobicity patterns.

3. Keep modifications minimal (1–2 changes).
\end{tcolorbox}

\textbf{Peptides - Default Task 2 (EXPLORE):}

\begin{tcolorbox}[promptbox]
TASK: Generate peptides with meaningfully different modifications to the input.

HINTS:
1. Try substitutions from different amino acid classes (polar<->hydrophobic, charged<->neutral).

2. Vary the length by adding or removing 1–3 residues.

3. Each output should explore a different modification strategy.
\end{tcolorbox}

\textbf{Peptides - Default Task 3: (SHUFFLE):}

\begin{tcolorbox}[promptbox]
TASK: Generate peptides by rearranging amino acids in the input.

HINTS:

1. Try swapping positions of residues.

2. Try reversing short segments (3–5 residues).

3. Try circular permutations (move N-terminal residues to C-terminus).
\end{tcolorbox}

\subsection{Task performance statistics (performance\textunderscore stats).}
\label{app:planner_stats_ex}
At initialization, the Planner Agent receives performance\textunderscore stats =
\texttt{"No performance data yet."}. Since the Task Registry is updated online, in subsequent calls performance\textunderscore stats is a table of task success rates computed from the registry, where success rate is defined as the fraction of task executions that produced at least one improving candidate. The Task Registry holds at most $20$ tasks at a time. The performance\textunderscore stats are presented to the Planner Agent using the unique task name for each task in the current Task Registry, and it's current success rate, presented in order from highest to lowest number of attempts so far. Example snapshots of the performance\textunderscore stats text included in the Planner Agent prompt for each domain are provided below:

\textbf{Example performance\textunderscore stats Text - Molecules:}

\begin{tcolorbox}[promptbox]
SIMILAR: 15/42 (36\%)

CROSSOVER: 9/28 (32\%)

POLARITY\textunderscore MOD: 7/16 (44\%)

EXPLORE: 6/14 (43\%)

SCAFFOLD\textunderscore HOP: 5/12 (42\%)

ELECTROSTATIC\textunderscore OPTIMIZATION: 4/10 (40\%)

SWAP\textunderscore FUNCTIONAL\textunderscore GROUPS: 3/9 (33\%)

SIMPLIFY: 2/8 (25\%)

RING\textunderscore EXPANSION: 0/6 (0\%)

\end{tcolorbox}

\textbf{Example performance\textunderscore stats Text - Peptides:}

\begin{tcolorbox}[promptbox]
EXPLORE: 24/51 (47\%)

SIMILAR: 10/45 (22\%)

LENGTH\textunderscore VARIATION: 8/20 (40\%)

SHUFFLE: 2/13 (15\%)

SWAP\textunderscore ANALOG: 0/12 (0\%)

SIMPLIFY: 4/8 (50\%)

SWAP\textunderscore ANALOG\textunderscore V2 : 1/3 (33\%)

\end{tcolorbox}

\subsection{Existing task summary (existing\textunderscore tasks\textunderscore summary).}
\label{app:planner_task_summary_ex}
To avoid spending prompt budget on full task descriptions, we provide an \emph{existing\textunderscore tasks\textunderscore summary} that lists each unique task name along with a truncated preview of the task description (the first 100 characters in the first line). For example, in the molecule domain this summary will contain entries such as \\
\texttt{EXPLORE: TASK: Generate SMILES with different meaningful structural changes...}, 
\texttt{SCAFFOLD\_HOP: TASK: Generate scaffold hopping variations of the input molecule...}, and 
\texttt{SIMILAR: TASK: Generate SMILES that are structurally similar...}. \\
As optimization progresses, this list will also include newly created tasks proposed by the Planner Agent (e.g., \texttt{NEW\_TASK1: TASK: ...}), in addition to the original default tasks. A similar structure is used for peptides (e.g., \texttt{EXPLORE}, \texttt{SHUFFLE}, \texttt{SIMILAR}, along with all Planner Agent–generated tasks in the Task Registry).

\subsection{Full Planner Agent prompt examples.}
\label{app:planner_full_prompt_ex}
Here we provide examples of full Planner agent prompts for each domain. See \cref{app:c_global_examples} for examples of the $C_{\text{global}}$ that is included in each full example prompt below.

\textbf{Molecules - Full Planner Agent Prompt Example:}
\begin{tcolorbox}[promptbox]
You are a prompt generator for a molecular optimization system that is trying to find the highest-scoring molecules.

\#\# MOLECULE-SCORE DATA (sorted high to low)

\{C\textunderscore global\}

\#\# TASK PERFORMANCE (success rate = score improvements / attempts)

\{performance\textunderscore stats\}

\#\# EXISTING TASKS (you can reuse by name or create new ones)

\{existing\textunderscore tasks\textunderscore summary\}

---

\#\# YOUR ANALYSIS PROCESS

1. **Study the Score Gradient:** Compare molecules with SIMILAR scores. What small structural change caused one to score slightly higher than another? These small differences are highly informative.

2. **High vs Low Contrast:** What features appear in top scorers but not low scorers? (ring types, chain lengths, functional groups, heteroatoms, flexibility)

3. **Identify Gaps:** What types of modifications have NOT been tried yet? What regions of chemical space remain unexplored?

\#\# YOUR GOAL

Generate task prompts that help a smaller LLM:

- **EXPLOIT:** Make targeted modifications based on patterns you observe in the score gradient

- **EXPLORE:** Try diverse, creative modifications to discover new promising regions

We are often stuck at local optima. To escape, we need BOTH:

- Smart exploitation of what seems to work

- Broad exploration of untried modification types

\#\# YOUR OUTPUT FORMAT

Return a JSON object with task names as keys and task descriptions as values.

- To REUSE an existing task: {"TASK\_NAME": "USE\_EXISTING"}

- To CREATE a new task: {"NEW\_NAME": "TASK: ... HINTS: ..."}

Example:
\{

    "SIMILAR": "USE\_EXISTING",
    
    "EXPLORE": "USE\_EXISTING",
    
    "FUSER": "TASK: Generate variants of the input molecule by fusing adjacent rings into polycyclic cores.
    
    HINTS:
    
    1. Merge rings connected by short chains (1-2 atoms).
    
    2. Create bicyclic or tricyclic fused systems.
    
    3. Preserve peripheral substituents."
    
\}

\#\# GUIDELINES

1. Output **8-10 tasks total** - a mix of existing and new.

2. Include 2-3 EXPLOITATION tasks (targeted at patterns you observed).

3. Include 2-3 EXPLORATION tasks (creative, untried modification types).

4. Include 2-4 reliable existing tasks that have (>0\%) success rates.

5. If a task 0 successes, avoid it or create an improved version (e.g., TASK\_NAME\_V2).

6. Keep new task descriptions concise (3-5 hints max).

7. New task names: SHORT, DESCRIPTIVE, ALL\_CAPS (e.g., ATOM\_SWAP, STABILIZE, RIGIDIFY).

\#\# CREATIVE EXPLORATION IDEAS

Consider tasks involving:

- Specific functional groups

- Specific atoms

- Specific ring modifications (aromatic<->aliphatic, 5-ring<->6-ring, fusion, spiro)

- Chain modifications (extend, shorten, branch, cyclize)

- Polarity changes (add polar groups, remove polar groups)

- Symmetry changes (break symmetry, add symmetry)

- Conformational changes (rigidify, flexibilize, add rotatable bonds)

Think creatively! What modification might lead to a breakthrough?
\end{tcolorbox}

\textbf{Peptides - Full Planner Agent Prompt Example:}
\begin{tcolorbox}[promptbox]
You are a prompt generator for an antimicrobial peptide (AMP) optimization system that is trying to find peptides with the LOWEST MIC (Minimum Inhibitory Concentration).

LOWER MIC = BETTER (less peptide needed to inhibit bacterial growth).

\#\# PEPTIDE-MIC DATA (sorted best to worst, LOWEST MIC first)

\{C\textunderscore global\}

\#\# TASK PERFORMANCE (success rate = MIC improvements / attempts)

\{performance\textunderscore stats\}

\#\# EXISTING TASKS (you can reuse by name or create new ones)

\{existing\textunderscore tasks\textunderscore summary\}

---

\#\# YOUR ANALYSIS PROCESS

1. **Study the Score Gradient:** Compare peptides with SIMILAR MICs. What small sequence change caused one to have slightly lower MIC than another? These small differences are highly informative.

2. **High vs Low Contrast:** What features appear in top performers but not poor performers? (charge distribution, hydrophobic patches, length, specific motifs)

3. **Identify Gaps:** What types of modifications have NOT been tried yet? What regions of sequence space remain unexplored?

\#\# YOUR GOAL

Generate task prompts that help a smaller LLM:

- **EXPLOIT:** Make targeted modifications based on patterns you observe in the score gradient

- **EXPLORE:** Try diverse, creative modifications to discover new promising regions

We are often stuck at local optima. To escape, we need BOTH:

- Smart exploitation of what seems to work

- Broad exploration of untried modification types

\#\# YOUR OUTPUT FORMAT

Return a JSON object with task names as keys and task descriptions as values.

- To REUSE an existing task: \{"TASK\_NAME": "USE\_EXISTING"\}

- To CREATE a new task: \{"NEW\_NAME": "TASK: ... HINTS: ..."\}

Example:

\{

    "SIMILAR": "USE\_EXISTING",
    
    "EXPLORE": "USE\_EXISTING",
    
    "CHARGE\_BOOST": "TASK: Increase the net positive charge 
    of the input peptide.
    
    HINTS:
    
    1. Replace neutral residues with K or R.
    
    2. Replace acidic residues (D, E) with neutral or basic ones.
    
    3. Add K or R at termini."
    
\}

\#\# GUIDELINES

1. Output **8-10 tasks total** - a mix of existing and new.

2. Include 2-3 EXPLOITATION tasks (targeted at patterns you observed).

3. Include 2-3 EXPLORATION tasks (creative, untried modification types).

4. Include 2-4 reliable existing tasks that have (>0\%) success rates.

5. If a task 0 successes, avoid it or create an improved version (e.g., TASK\_NAME\_V2).

6. Keep new task descriptions concise (3-5 hints max).

7. New task names: SHORT, DESCRIPTIVE, ALL\_CAPS (e.g., CHARGE\_BOOST, HELIX\_FORM, TRUNCATE).

\#\# CREATIVE EXPLORATION IDEAS FOR AMPs

Consider tasks involving:

- Charge modifications (increase/decrease cationic character)

- Hydrophobicity changes (add/remove hydrophobic residues)

- Secondary structure (promote helix, add proline kinks)

- Length modifications (truncate, extend, repeat motifs)

- Amphipathicity (arrange polar/nonpolar faces)

- Specific motif insertions (WRW, KWK, etc.)

Think creatively! What modification might lead to a breakthrough?
\end{tcolorbox}

\section{Example Worker Agent prompts}
\label{app:worker_prompt_ex}

Unlike the Explorer and Planner Agents, which use a single monolithic prompt that is dynamically updated with the current global context, Worker Agents use a two-part prompting structure consisting of a \textbf{system prompt} and a \textbf{generation-time prompt}.

\subsection{Worker system prompt.}
\label{app:worker_sys_prompt_ex}

For a given task prompt $p \in \mathcal{P}_{\text{work}}$ generated by the Planner Agent, the Worker system prompt is constructed as:
\[
\texttt{SystemPrompt} = \texttt{prefix}_{\text{domain}} + p + \texttt{suffix}_{\text{domain}}.
\]
The prefix and suffix are fixed (unchanging) for a given domain, while the task description $p$ varies. Since examples of $p$ output by the Planner Agent are provided in \cref{app:task_registry_init_ex} and \cref{app:planner_generated_tasks_ex}, here we provide the domain-specific prefix and suffix needed to complete the Worker Agent system prompt for each domain.

\textbf{Molecule domain prefix:}
\begin{tcolorbox}[promptbox]
You are an expert molecule generator operating in SMILES space.

INPUT: You will be given a single input molecule in the prompt.

\end{tcolorbox}

\textbf{Molecule domain suffix (fixed):}
\begin{tcolorbox}[promptbox]
OUTPUT FORMAT (REQUIRED): Return ONLY a JSON object with a list of 5-10 SMILES strings called 'candidates'.
\end{tcolorbox}

\textbf{Peptide domain prefix (fixed):}
\begin{tcolorbox}[promptbox]
You are an expert peptide generator operating in amino acid sequence space.

INPUT: You will be given a single input peptide in the prompt.
\end{tcolorbox}

\textbf{Peptide domain suffix (fixed):}
\begin{tcolorbox}[promptbox]
OUTPUT FORMAT (REQUIRED): Return ONLY a JSON object with a list of 5-10 peptide sequences called 'candidates'.
\end{tcolorbox}

Note that the prefix and suffix text for different domains is nearly identical, with only small changes needed to specify what type of biological objects we want the Worker Agent to generate (e.g., SMILES strings vs. peptide sequences).

\paragraph{Example full Worker system prompts.}
We additionally provide an example of a full worker system prompt for each domain, using one example default task prompt $p$ from each domain to construct the full system prompts as:
\[
\texttt{SystemPrompt} = \texttt{prefix}_{\text{domain}} + p + \texttt{suffix}_{\text{domain}}.
\]. 
\\
\textbf{Full Worker System Prompt Example - Molecules:}
\begin{tcolorbox}[promptbox]
You are an expert molecule generator operating in SMILES space.

INPUT: You will be given a single input molecule in the prompt.

TASK: Generate SMILES that are structurally similar to the input molecule.

HINTS:

1. Modify side chains, linkers, or substituents.

2. Keep the core scaffold mostly intact.

OUTPUT FORMAT (REQUIRED): Return ONLY a JSON object with a list of 5-10 SMILES strings called 'candidates'.
\end{tcolorbox}

\textbf{Full Worker System Prompt Example - Peptides:}
\begin{tcolorbox}[promptbox]
You are an expert peptide generator operating in amino acid sequence space.

INPUT: You will be given a single input peptide in the prompt.

TASK: Generate peptides by rearranging amino acids in the input.

HINTS:

1. Try swapping positions of residues.

2. Try reversing short segments (3–5 residues).

3. Try circular permutations (move N-terminal residues to C-terminus).

OUTPUT FORMAT (REQUIRED): Return ONLY a JSON object with a list of 5-10 peptide sequences called 'candidates'.
\end{tcolorbox}

% ---------------------------------------------
% ---------------------------------------------
\subsection{Worker generation-time prompt.}
\label{app:worker_gen_prompt_ex}

While the system prompt encodes the task description, the specific seed sequence or molecule is provided separately at generation time. For a current seed $x_{\text{curr}}$ (see Algorithm~\ref{alg:agent_opt}), the generation-time prompt is:

\textbf{Worker Agent Generate-time Prompt - Molecules:}
\begin{tcolorbox}[promptbox]
Input Molecule: \{x\_curr\}
Modify it to generate 5-10 VALID SMILES strings.
\end{tcolorbox}

\textbf{Worker Agent Generate-time Prompt - Peptides:}
\begin{tcolorbox}[promptbox]
Input Peptide: \{x\_curr\}
Modify it to generate 5-10 new peptides.
\end{tcolorbox}

This separation allows the same Worker Agent (same system prompt) to be reused across different seeds during the local search persistence loop. A new Worker Agent (with a new system prompt) is instantiated only when switching to a different task prompt $p \in \mathcal{P}_{\text{work}}$, where $\mathcal{P}_{\text{work}}$ is the latest set of task prompts generated by the Planner Agent.

\section{Example tasks generated by the Planner Agent}
\label{app:planner_generated_tasks_ex}
Throughout optimization, the Planner Agent generates new tasks $p$ for the Worker Agents to execute, and adds them to the existing Task Registry. In this section, we provide examples of some of the task text $p$  generated by the Planner Agent for each domain (molecules and peptides) during runs of \ourmethod{}. For each domain, we provide examples of $20$ Planner-agent-generated tasks $p$ that led to at at least one score improvement (meaning at least one molecule generated by a Worker Agent executing the task achieved a higher score than the current best during optimization).  For the molecule domain, we provide $2$ examples from \ourmethod{} runs on each of the $10$ GuacaMol objectives from \cref{sec:experiments}, for a total of $20$ examples.

\textbf{Molecules Example 1 - Objective: adip - Task name: CHAIN\_FLEXIBILITY, Task text:} 
\begin{tcolorbox}[promptbox]
Systematically modify linker chain lengths and flexibility between core structural elements.

HINTS:

1. Insert/remove CH\_2 units in aliphatic chains.

2. Add/remove rotatable bonds near functional groups.

3. Test both rigid (cycloalkyl) and flexible (alkoxy) linkers.
\end{tcolorbox}

\textbf{Molecules Example 2 - Objective: adip - Task name: RING\_SIZE\_MOD, Task text:} 
\begin{tcolorbox}[promptbox]
TASK: Adjust ring sizes in high-scoring scaffolds to explore conformational effects. 

HINTS: 

1. Convert 5-membered rings to 6-membered (or vice versa). 

2. Maintain aromaticity where possible. 

3. Ensure substituents are appropriately positioned.
\end{tcolorbox}

\textbf{Molecules Example 3 - Objective: fexo - Task name: BRANCHING, Task text:} 
\begin{tcolorbox}[promptbox]
TASK: Increase molecular branching in hydrocarbon chains.

HINTS:

1. Add methyl branches to aliphatic chains.

2. Create gem-dimethyl groups.

3. Introduce cyclopropyl rings for rigidity.
\end{tcolorbox}

\textbf{Molecules Example 4 - Objective: fexo - Task name: ATOM\_SWAP, Task text:} 
\begin{tcolorbox}[promptbox]
TASK: Replace key carbon atoms with heteroatoms (N, O, S) in aliphatic rings and linkers.

HINTS:

1. Prioritize substitutions that maintain ring size but alter electronic properties.

2. Test bioisosteric replacements (e.g., -CH2- --> -O- in linkers).
\end{tcolorbox}

\textbf{Molecules Example 5 - Objective: med1 - Task name: SPIRO\_FUSE, Task text:} 
\begin{tcolorbox}[promptbox]
TASK: Generate spiro-fused ring systems to explore novel conformational constraints.

HINTS:

1. Identify adjacent rings separated by 1-2 atoms.

2. Merge into spiro junctions (shared single atom).

3. Preserve peripheral substituents like isopropyl groups.
\end{tcolorbox}

\textbf{Molecules Example 6 - Objective: med1 - Task name: RING\_EXPANSION\_V2, Task text:} 
\begin{tcolorbox}[promptbox]
TASK: Expand non-aromatic rings from 5 to 6 members.

HINTS:

1. Target rings adjacent to ketones.

2. Use methylene insertion.

3. Maintain bicyclic rigidity.
\end{tcolorbox}

\textbf{Molecules Example 7 - Objective: med2 - Task name: QUINAZOLINONE\_KETONE\_SWAP\_V3, Task text:} 
\begin{tcolorbox}[promptbox]
TASK: Replace the quinazolinone ketone with thiazole or oxazole rings to alter electronic distribution and hydrogen bonding.

HINTS:

1. Maintain planarity at the core interaction site.

2. Ensure retention of key hydrogen bond acceptors.

3. Test both 5-membered and 6-membered heterocycle replacements.",
\end{tcolorbox}

\textbf{Molecules Example 8 - Objective: med2 - Task name: RING\_FUSION\_ENHANCE, Task text:} 
\begin{tcolorbox}[promptbox]
TASK: Generate fused polycyclic variants by merging indole with adjacent aromatic rings through strategic bond formation.

HINTS:

1. Create 6-5-6 tricyclic systems.

2. Preserve indole's NH while forming new ring junctions.

3. Explore both angular and linear fusion patterns.
\end{tcolorbox}

\textbf{Molecules Example 9 - Objective: osmb - Task name: CORE\_SWAP\_BIOISOSTERE, Task text:} 
\begin{tcolorbox}[promptbox]
TASK: Replace pyrimidine cores with bioisosteric heterocycles (e.g., triazine, pyridone, thiazine) while preserving substituent patterns.

HINTS:

1. Match nitrogen positioning in new cores.

2. Maintain planar aromaticity.

3. Evaluate both 5- and 6-membered alternative cores.
\end{tcolorbox}

\textbf{Molecules Example 10 - Objective: osmb - Task name: HYDROXYL\_POSITION, Task text:} 
\begin{tcolorbox}[promptbox]
TASK: Systematically relocate hydroxyl groups between chain positions and ring substituents.

HINTS:

1. Compare terminal vs internal hydroxyl placement

2. Test hydroxyl migration to adjacent carbons

3. Consider diol formation in chains
\end{tcolorbox}

\textbf{Molecules Example 11 - Objective: pdop - Task name: INDOLE\_BRANCHING, Task text:} 
\begin{tcolorbox}[promptbox]
TASK: Add alkyl or functionalized branches to indole rings in the input molecule.

HINTS:

1. Introduce methyl or hydroxyl groups at indole C4-C7 positions.

2. Attach small polar groups (e.g., -CH2OH) to indole nitrogen.

3. Preserve core indole hydrogen bonding capability.
\end{tcolorbox}

\textbf{Molecules Example 12 - Objective: pdop - Task name: CHAIN\_MOD, Task text:} 
\begin{tcolorbox}[promptbox]
TASK: Modify alkyl chain lengths and branching in linker regions (e.g., +1/-1 CH2, add methyl branches).

HINTS:

1. Focus on chains between amide bonds.

2. Test both elongation and shortening.

3. Introduce branching near aromatic systems.
\end{tcolorbox}

\textbf{Molecules Example 13 - Objective: rano - Task name: FLUORINE\_CHAIN\_OPT, Task text:} 
\begin{tcolorbox}[promptbox]
TASK: Optimize fluorinated chain geometry by adjusting double bond positions and terminal fluorine placement.

HINTS:

1. Shift F from terminal to penultimate position.

2. Alternate E/Z configurations in conjugated system.

3. Introduce cyclopropane into the chain for rigidity.
\end{tcolorbox}

\textbf{Molecules Example 14 - Objective: rano - Task name: DOUBLE\_BOND\_MOD, Task text:} 
\begin{tcolorbox}[promptbox]
TASK: Alter conjugated double bond systems. 

HINTS:

1. Shift /C=C/ positions closer to aromatic rings.

2. Introduce second conjugated bond in chains.

3. Test both E/Z configurations in new positions.
\end{tcolorbox}

\textbf{Molecules Example 15 - Objective: siga - Task name: FLUORINE\_CHAIN\_RIGIDIFY\_V2, Task text:} 
\begin{tcolorbox}[promptbox]
TASK: Convert flexible fluorinated chains into constrained cyclopropane or cyclobutane rings.

HINTS:

1. Focus on chains with 3-4 carbons

2. Preserve terminal fluorine positioning

3. Test both geminal and vicinal fluorine patterns
\end{tcolorbox}

\textbf{Molecules Example 16 - Objective: siga - Task name: FLUORINE\_SYMMETRY\_BREAK, Task text:} 
\begin{tcolorbox}[promptbox]
TASK: Introduce asymmetry in fluorinated cyclopropane rings while preserving total fluorine count.

HINTS:

1. Convert symmetric CF2 groups to CF-CF2 patterns

2. Create chiral centers through differential fluorination

3. Test combinations of mono/di/tri-fluorinated cyclopropane rings
\end{tcolorbox}

\textbf{Molecules Example 17 - Objective: valt - Task name: HYDROXYL\_TUNE, Task text:} 
\begin{tcolorbox}[promptbox]
TASK: Modify aromatic rings by adding/removing hydroxyl groups at positions adjacent to existing oxygen substituents.

HINTS:

1. Prioritize positions ortho/para to ketone groups

2. Maintain hydrogen bonding patterns

3. Avoid over-oxidizing aliphatic regions
\end{tcolorbox}

\textbf{Molecules Example 18 - Objective: valt - Task name: LACTAM\_STEREOCHEMISTRY\_V2, Task text:} 
\begin{tcolorbox}[promptbox]
TASK: Systematically invert stereochemistry at lactam alpha/beta carbons and N-methylation sites.

HINTS:

1. Generate all stereoisomers of top-scoring scaffolds.

2. Test combinations of (R/S) configurations at branching points.

3. Add/removal methyl groups at stereocenters to modulate rigidity.
\end{tcolorbox}

\textbf{Molecules Example 19 - Objective: zale - Task name: SUBSTITUENT\_SWAP, Task text:} 
\begin{tcolorbox}[promptbox]
TASK: Alter key substituents on the aromatic core while preserving overall scaffold.

HINTS:

1. Replace cyano groups with nitro or trifluoromethyl.

2. Shift carboxylic acid position to meta/para locations.

3. Introduce fluorine at unexplored positions for electronic effects.
\end{tcolorbox}

\textbf{Molecules Example 20 - Objective: zale - Task name: RING\_AROMATIZE, Task text:} 
\begin{tcolorbox}[promptbox]
TASK: Convert aliphatic rings to aromatic systems to improve rigidity and pi-stacking potential.

HINTS:

1. Dehydrogenate cyclohexane rings to benzene.

2. Introduce nitrogen atoms to form pyridine-like aromatic systems.

3. Remove saturated linkers between aromatic rings.
\end{tcolorbox}

% ------------------------

\textbf{Peptides Example 1 - Task name: LENGTH\_VARIATION, Task text:} 
\begin{tcolorbox}[promptbox]
TASK: Optimize peptide length through truncation/extension.

HINTS:

1. Remove 1-3 residues from N/C termini.

2. Extend termini with K/R or hydrophobic residues.

3. Test lengths between 15-25 amino acids.

4. Prioritize core active region preservation.

\end{tcolorbox}

\textbf{Peptides Example 2 - Task name: TERMINI\_OPTIMIZE, Task text:} 
\begin{tcolorbox}[promptbox]
TASK: Optimize N or C terminus modifications for improved stability and activity.

HINTS:

1. Add acetylation or amidation for protease resistance.

2. Extend termini with K/R for charge enhancement.

3. Add hydrophobic residues at termini for membrane anchoring.

4. Test truncations to find minimal effective length.
\end{tcolorbox}

\textbf{Peptides Example 3 - Task name: MOTIF\_SWAPPING, Task text:} 
\begin{tcolorbox}[promptbox]
TASK: Exchange motifs between high- and low-MIC peptides to identify critical sequences.

HINTS:

1. Replace segments in high-MIC peptides with low-MIC motifs (e.g., KLWK).

2. Retain overall charge balance.

3. Test shuffled motif variants.
\end{tcolorbox}

\textbf{Peptides Example 4 - Task name: TRUNCATE, Task text:} 
\begin{tcolorbox}[promptbox]
TASK: Identify minimal functional core sequences.

HINTS:

1. Systematically remove 1-2 residues from N/C termini.

2. Preserve central K/R-rich motifs and hydrophobic anchors.

3. Test fragments for MIC retention vs. full-length peptides.
\end{tcolorbox}

\textbf{Peptides Example 5 - Task name: D\_AMINO\_ACID\_STABILIZATION, Task text:} 
\begin{tcolorbox}[promptbox]
TASK: Introduce D-amino acids to enhance proteolytic resistance without compromising activity.

HINTS:

1. Replace 1-3 central residues with D-amino acids.

2. Prioritize hydrophobic residues (F/W/L) for D-substitution.

3. Maintain N/C-terminal L-residues for binding.

4. Test substitutions at positions 6, 9, 12.
\end{tcolorbox}

\textbf{Peptides Example 6 - Task name: AROMATIC\_RESIDUE\_SWAP, Task text:} 
\begin{tcolorbox}[promptbox]
TASK: Systematically substitute aliphatic hydrophobic residues (L/I/V) with aromatic (F/W) in hydrophobic regions.

HINTS:

1. Prioritize substitutions at positions 3, 7, and 11 in helical sequences.

2. Maintain aromatic content between 15-25\% of total sequence.

3. Combine with charge redistribution for balanced amphipathicity.

\end{tcolorbox}

\textbf{Peptides Example 7 - Task name: HYDROPHOBIC\_FACE\_BALANCE, Task text:} 
\begin{tcolorbox}[promptbox]
TASK: Optimize hydrophobic face composition to prevent aggregation while maintaining membrane affinity.

HINTS:

1. Replace 1-2 aromatic residues with aliphatic hydrophobes (L/I/V).

2. Create a hydrophobic face with 4-5 residues including 1-2 F/W.

3. Ensure hydrophobic moment >0.5 for amphipathicity.",
\end{tcolorbox}

\textbf{Peptides Example 8 - Task name: CHARGE\_DISTRIBUTION, Task text:} 
\begin{tcolorbox}[promptbox]
TASK: Optimize spatial distribution of cationic charges.

HINTS:

1. Space K/R residues every 3-4 positions to enhance membrane interaction.

2. Avoid charge clustering at termini; distribute along the peptide.

3. Pair cationic residues with hydrophobic ones for synergistic effects.

4. Test alternating charge/hydrophobic patterns.
\end{tcolorbox}

\textbf{Peptides Example 9 - Task name: BETA\_AMINO\_ACID\_INSERTION, Task text:} 
\begin{tcolorbox}[promptbox]
TASK: Insert beta-amino acids to alter peptide backbone conformation and enhance activity.

HINTS:

1. Identify positions where beta-amino acids can improve helical structure.

2. Insert beta-amino acids at non-core motif positions.

3. Balance hydrophobicity and cationic charge in modified sequences.

4. Evaluate protease resistance improvements.
\end{tcolorbox}

\textbf{Peptides Example 10 - Task name: W\_ENRICHMENT, Task text:} 
\begin{tcolorbox}[promptbox]
TASK: Increase tryptophan (W) content in antimicrobial peptides to enhance membrane binding.

HINTS:

1. Substitute hydrophobic residues (L, I, V) with W in key positions.

2. Add W at N/C termini to improve membrane interaction.

3. Target 3-4 W residues total; avoid excessive W to prevent aggregation.

4. Prioritize W enrichment in hydrophobic regions of amphipathic structures.
\end{tcolorbox}

\textbf{Peptides Example 11 - Task name: AMPHIPATHIC\_BALANCE, Task text:} 
\begin{tcolorbox}[promptbox]
TASK: Improve amphipathic segregation to enhance bacterial membrane disruption.

HINTS:

1. Rearrange hydrophobic residues (L, I, V, F, W) to one helical face.

2. Position cationic residues (K, R) on the opposite face.

3. Use helical wheel projections to validate segregation.
\end{tcolorbox}

\textbf{Peptides Example 12 - Task name: W\_SPACING\_OPTIMIZATION, Task text:} 
\begin{tcolorbox}[promptbox]
TASK: Optimize spacing between tryptophan residues to enhance membrane interaction.

HINTS:

1. Space W residues at i, i+3, or i+4 intervals for helical alignment.

2. Maintain 2-3 W residues in hydrophobic regions.

3. Avoid adjacent W clusters to prevent aggregation.
\end{tcolorbox}

\textbf{Peptides Example 13 - Task name: CATIONIC\_REINFORCEMENT, Task text:} 
\begin{tcolorbox}[promptbox]
TASK: Increase cationic charge in specific positions to enhance bacterial membrane interaction.

HINTS:

1. Replace neutral residues with K or R in hydrophilic regions.

2. Add K/R at positions adjacent to hydrophobic clusters.

3. Target a net positive charge of +4 to +6.

4. Avoid clustering cationic residues to prevent steric hindrance.",
\end{tcolorbox}

\textbf{Peptides Example 13 - Task name: INSERT\_MOTIF, Task text:} 
\begin{tcolorbox}[promptbox]
TASK: Insert known antimicrobial motifs into the peptide sequence.

HINTS:

1. Try WRW, RKRK, or FLPLIG motifs.

2. Insert at N-terminus, C-terminus, or central region.

3. Maintain overall charge and hydrophobicity balance.

4. Test motif orientation (forward/reverse).",
\end{tcolorbox}

\textbf{Peptides Example 14 - Task name: TRUNCATE\_MIDDLE, Task text:} 
\begin{tcolorbox}[promptbox]
TASK: Shorten the peptide by removing residues from the middle.

HINTS:

1. Remove 2-4 consecutive central residues.

2. Preserve N/C-terminal motifs and charge.

3. Verify helical structure remains intact.

4. Test MIC after each truncation.
\end{tcolorbox}

\textbf{Peptides Example 15 - Task name: AROMATIC\_PATCH\_ENHANCE, Task text:} 
\begin{tcolorbox}[promptbox]
TASK: Increase aromatic hydrophobic residues (W/F) in clusters.

HINTS:

1. Add W/F in positions that form aromatic patches.

2. Maintain or enhance cationic residue distribution.

3. Prioritize clusters in central regions of the peptide.
\end{tcolorbox}

\textbf{Peptides Example 16 - Task name: WRW\_MOTIF\_OPTIMIZATION, Task text:} 
\begin{tcolorbox}[promptbox]
TASK: Optimize WRW motif placement and surrounding residues.

HINTS:

1. Insert WRW in hydrophobic regions.

2. Flank with cationic residues (K/R).

3. Maintain helical structure if applicable.

4. Test variations in motif spacing.
\end{tcolorbox}

\textbf{Peptides Example 17 - Task name: CATIONIC\_CLUSTER\_BOOST, Task text:} 
\begin{tcolorbox}[promptbox]
TASK: Cluster cationic (K/R) and aromatic (W/F) residues to form disruptive motifs.

HINTS:

1. Create adjacent K/R-W/F pairs.

2. Position clusters near hydrophobic faces.

3. Maintain 4-6 net positive charges.
\end{tcolorbox}

\textbf{Peptides Example 18 - Task name: PROLINE\_INSERTION, Task text:} 
\begin{tcolorbox}[promptbox]
TASK: Introduce proline residues to modulate peptide flexibility and structure.

HINTS:

1. Replace non-essential residues with proline at strategic positions.

2. Insert proline at 1/3 and 2/3 sequence positions.

3. Limit proline count to 1-2 per peptide.

4. Test MIC impact of structural flexibility changes.
\end{tcolorbox}
 
\textbf{Peptides Example 19 - Task name: HYDROPHOBIC\_PATCH\_ENHANCEMENT, Task text:} 
\begin{tcolorbox}[promptbox]
TASK: Create localized hydrophobic patches while preserving cationic residues.

HINTS:

1. Substitute neutral residues (A, S, T) with I/L/V in adjacent positions.

2. Maintain 3-4 aromatic residues in the sequence.

3. Avoid disrupting existing charge clusters.
\end{tcolorbox}

\textbf{Peptides Example 20 - Task name: KLRWFKK\_HYDROPHOBIC\_OPT, Task text:} 
\begin{tcolorbox}[promptbox]
TASK: Optimize hydrophobic residues within the KLRWFKK motif to enhance membrane interaction.

HINTS:

1. Substitute residues at positions 3, 6, and 9 with hydrophobic amino acids (I, V, F, W).

2. Maintain helical periodicity by preserving i, i+3, i+4 spacing.

3. Prioritize substitutions that increase overall hydrophobic moment.
\end{tcolorbox}

\section{Hand-designed static Worker Agent prompts (ablation)}
\label{sec:app_static_worker_prompts}
In \cref{fig:agents_ablation}, we ablate the usefulness of the Planner Agent in \ourmethod{} by removing the Planner entirely  and instead using a set of $10$ static (unchanging) domain-specific task prompts $p_1, \ldots, p_{10}$. This ablation directly demonstrates that the dynamic modification tactics created by the Planner Agent during optimization lead to improved performance compared to just using a fixed/statics set of modification tactics throughout. In this section, we provide the exact text defining the static, domain-specific task prompts used for this ablation. 
The first three static task prompts, $p_1, p_2,$ and $p_3$, are always exactly the three domain-specific ``default" tasks used to initialize the Task Registry, which are provided in \cref{app:task_registry_init_ex}. 
Below, we provide the remaining static task prompts $p_4, \ldots, p_{10}$ that we used to run this ablation on the three molecule design tasks in \cref{fig:agents_ablation}:

\textbf{Ablation Static Prompt 4 ($p_4$) - Task name: ATOM\_SWAP, Task text:}  
\begin{tcolorbox}[promptbox]
TASK: Perform single atom swaps.

HINTS:

1. Swap atoms in rings and linkers (e.g., C<->N, O<->S, H<->F).

2. Keep ring count and chain length the same.
\end{tcolorbox}

\textbf{Ablation Static Prompt 5 ($p_5$) - Task name: LINEARIZE, Task text:}  
\begin{tcolorbox}[promptbox]
TASK: Make the molecule more flexible/linear.

HINTS:

1. Break bulky ring systems into chains.

2. Replace fused rings with rotatable linkers.

3. Increase rotatable bond count.
\end{tcolorbox}

\textbf{Ablation Static Prompt 6 ($p_6$) - Task name: RIGIDIFY, Task text:}  
\begin{tcolorbox}[promptbox]
TASK: Make the molecule more rigid.

HINTS:

1. Lock rotatable linkers into rings (e.g., propyl --> piperidine).

2. Fuse aromatic rings (e.g., benzene --> naphthalene).

3. Reduce rotatable bonds.
\end{tcolorbox}

\textbf{Ablation Static Prompt 7 ($p_7$) - Task name: FUSER, Task text:}  
\begin{tcolorbox}[promptbox]
TASK: Fuse adjacent rings into polycyclic cores.

HINTS:

1. Merge rings connected by short chains (1-2 atoms).

2. Create bicyclic or tricyclic fused systems.

3. Preserve peripheral substituents.
\end{tcolorbox}

\textbf{Ablation Static Prompt 8 ($p_8$) - Task name: REASSEMBLE, Task text:}  
\begin{tcolorbox}[promptbox]
TASK: Recombine molecular fragments in new arrangements.

HINTS:

1. Reattach functional groups at different positions.

2. Swap positions of tail/linker/ring units.

3. Keep approximate size, vary connectivity.
\end{tcolorbox}

\textbf{Ablation Static Prompt 9 ($p_9$) - Task name: BREAK\_SYMMETRY, Task text:}  
\begin{tcolorbox}[promptbox]
TASK: Make the input molecule less symmetric.

HINTS:

1. Add substituents to only one side of symmetric scaffolds.

2. Replace one of two identical groups with something different.

3. Vary chain lengths on symmetric branches.
\end{tcolorbox}

\textbf{Ablation Static Prompt 10 ($p_{10}$) - Task name: STABILIZER, Task text:}  
\begin{tcolorbox}[promptbox]
TASK: Modify the input molecule to improve general chemical stability and reduce hazardous reactivity.

HINTS:

1. Replace chemically unstable or hazardous motifs (e.g., peroxides, azides/diazo, acyl halides, anhydrides, highly strained rings).

2. Reduce overly reactive functional groups by swapping for more inert alternatives.

3. If already stable, propose small stabilizing variations (e.g., saturation of labile bonds).
\end{tcolorbox}

%%%%%%%%%%%%%%%%%%%%%%%%%%%%%%%%%%%%%%%%%%%%%%%%%%%%%%%%%%%%

\newpage
\section*{NeurIPS Paper Checklist}

\begin{enumerate}

\item {\bf Claims}
    \item[] Question: Do the main claims made in the abstract and introduction accurately reflect the paper's contributions and scope?
    \item[] Answer: \answerYes{} % Replace by \answerYes{}, \answerNo{}, or \answerNA{}.
    \item[] Justification: All claims stated in this paper are supported by empirical results provided in \cref{sec:experiments} and \cref{sec:appendix-results}. The stated scope of the work accurately reflects what is discussed in the paper. 

\item {\bf Limitations}
    \item[] Question: Does the paper discuss the limitations of the work performed by the authors?
    \item[] Answer: \answerYes{} % Replace by \answerYes{}, \answerNo{}, or \answerNA{}.
    \item[] Justification: See discussion of limitations in \cref{app:limits}. 

\item {\bf Theory assumptions and proofs}
    \item[] Question: For each theoretical result, does the paper provide the full set of assumptions and a complete (and correct) proof?
    \item[] Answer: \answerNA{} % Replace by \answerYes{}, \answerNo{}, or \answerNA{}.
    \item[] Justification: This work does not include theoretical results. 

    \item {\bf Experimental result reproducibility}
    \item[] Question: Does the paper fully disclose all the information needed to reproduce the main experimental results of the paper to the extent that it affects the main claims and/or conclusions of the paper (regardless of whether the code and data are provided or not)?
    \item[] Answer: \answerYes{} % Replace by \answerYes{}, \answerNo{}, or \answerNA{}.
    \item[] Justification: All of the hyperparameters we used and other implementation details are provided in \cref{sec:experiments} and \cref{sec:app_imp_details}. Additionally, code to reproduce results is provided via an anonymous GitHub link in \cref{sec:experiments}. 

\item {\bf Open access to data and code}
    \item[] Question: Does the paper provide open access to the data and code, with sufficient instructions to faithfully reproduce the main experimental results, as described in supplemental material?
    \item[] Answer: \answerYes{} % Replace by \answerYes{}, \answerNo{}, or \answerNA{}.
    \item[] Justification: All of the hyperparameters we used and other implementation details are provided in \cref{sec:experiments} and \cref{sec:app_imp_details}. Additionally, code to reproduce results is provided via an anonymous GitHub link in \cref{sec:experiments}.

\item {\bf Experimental setting/details}
    \item[] Question: Does the paper specify all the training and test details (e.g., data splits, hyperparameters, how they were chosen, type of optimizer) necessary to understand the results?
    \item[] Answer:  \answerYes{} % Replace by \answerYes{}, \answerNo{}, or \answerNA{}.
    \item[] Justification: All of the hyperparameters we used and other implementation details are provided in \cref{sec:experiments} and \cref{sec:app_imp_details}. Additionally, code to reproduce results is provided via an anonymous GitHub link in \cref{sec:experiments}.

\item {\bf Experiment statistical significance}
    \item[] Question: Does the paper report error bars suitably and correctly defined or other appropriate information about the statistical significance of the experiments?
    \item[] Answer: \answerYes{} % Replace by \answerYes{}, \answerNo{}, or \answerNA{}.
    \item[] Justification: All empirical results in all plots and tables provide both the mean and standard error taken over multiple random runs (see \cref{sec:experiments}). 

\item {\bf Experiments compute resources}
    \item[] Question: For each experiment, does the paper provide sufficient information on the computer resources (type of compute workers, memory, time of execution) needed to reproduce the experiments?
    \item[] Answer: \answerYes{} % Replace by \answerYes{}, \answerNo{}, or \answerNA{}.
    \item[] Justification: All compute details are provided in \cref{sec:compute}.
    
\item {\bf Code of ethics}
    \item[] Question: Does the research conducted in the paper conform, in every respect, with the NeurIPS Code of Ethics \url{https://neurips.cc/public/EthicsGuidelines}?
    \item[] Answer: \answerYes{} % Replace by \answerYes{}, \answerNo{}, or \answerNA{}.
    \item[] Justification: We have made sure to adhere to the NeurIPS Code of Ethics in all aspects of our work.

\item {\bf Broader impacts}
    \item[] Question: Does the paper discuss both potential positive societal impacts and negative societal impacts of the work performed?
    \item[] Answer: \answerYes{} % Replace by \answerYes{}, \answerNo{}, or \answerNA{}.
    \item[] Justification: See discussion of broader impacts in \cref{sec:impacts}.
    
\item {\bf Safeguards}
    \item[] Question: Does the paper describe safeguards that have been put in place for responsible release of data or models that have a high risk for misuse (e.g., pre-trained language models, image generators, or scraped datasets)?
    \item[] Answer: \answerNA{} % Replace by \answerYes{}, \answerNo{}, or \answerNA{}.
    \item[] Justification: The paper does not release new data or models with potential societal concerns. 

\item {\bf Licenses for existing assets}
    \item[] Question: Are the creators or original owners of assets (e.g., code, data, models), used in the paper, properly credited and are the license and terms of use explicitly mentioned and properly respected?
    \item[] Answer: \answerYes{} % Replace by \answerYes{}, \answerNo{}, or \answerNA{}.
    \item[] Justification: The creators of all assets used to produce all results in this paper are cited in \cref{sec:experiments}. All assets used are open source software or models. 

\item {\bf New assets}
    \item[] Question: Are new assets introduced in the paper well documented and is the documentation provided alongside the assets?
    \item[] Answer: \answerNA{} % Replace by \answerYes{}, \answerNo{}, or \answerNA{}.
    \item[] Justification: This work does not release any new assets.

\item {\bf Crowdsourcing and research with human subjects}
    \item[] Question: For crowdsourcing experiments and research with human subjects, does the paper include the full text of instructions given to participants and screenshots, if applicable, as well as details about compensation (if any)? 
    \item[] Answer: \answerNA{} % Replace by \answerYes{}, \answerNo{}, or \answerNA{}.
    \item[] Justification: This work does not involve crowdsourcing nor research with human subjects.

\item {\bf Institutional review board (IRB) approvals or equivalent for research with human subjects}
    \item[] Question: Does the paper describe potential risks incurred by study participants, whether such risks were disclosed to the subjects, and whether Institutional Review Board (IRB) approvals (or an equivalent approval/review based on the requirements of your country or institution) were obtained?
    \item[] Answer: \answerNA{} % Replace by \answerYes{}, \answerNo{}, or \answerNA{}.
    \item[] Justification: No human subjects were involved in this work.

\item {\bf Declaration of LLM usage}
    \item[] Question: Does the paper describe the usage of LLMs if it is an important, original, or non-standard component of the core methods in this research? Note that if the LLM is used only for writing, editing, or formatting purposes and does \emph{not} impact the core methodology, scientific rigor, or originality of the research, declaration is not required.
    %this research? 
    \item[] Answer: \answerYes{} % Replace by \answerYes{}, \answerNo{}, or \answerNA{}.
    \item[] Justification: LLMs are indeed an essential component of our proposed method (see \cref{sec:method}). 

\end{enumerate}

\end{document}